\pdfoutput=1
\documentclass{article}
\usepackage{iclr2027_conference,times}
\iclrfinalcopy
\usepackage[utf8]{inputenc}
\usepackage[T1]{fontenc}
\usepackage{amsmath,amsfonts,amssymb,booktabs,graphicx,microtype,tabularx,longtable}
\usepackage{algorithm}
\usepackage[hidelinks]{hyperref}
\usepackage{url}
\title{Timestep Weighting: A Hidden Key to Effective ELBO-Based Flow-Matching RL}
\author{
 Qinwei Ma$^{1*}$ \quad Jingzhe Shi$^{2*}$ \quad Simin Fan$^{3}$ \quad Ling Li$^{4}$ \quad Alex Lamb$^{1}$\\[3pt]
\normalfont\small $^{1}$College of AI, Tsinghua University\\
\normalfont\small $^{2}$Department of Electrical and Computer Engineering, Princeton University\\
\normalfont\small $^{3}$EPFL\\
\normalfont\small $^{4}$Computer Science Department, Tsinghua University\\[3pt]
\normalfont\small $^*$Equal contribution
}

\newcommand{\E}{\mathbb E}
\newcommand{\KL}{D_{\mathrm{KL}}}
\begin{document}
\maketitle
\begingroup
\renewcommand{\thefootnote}{}
\footnotetext{\raggedright Email addresses: \texttt{qinweimartin@gmail.com}, \texttt{sjzworking@gmail.com}, \texttt{lambalex@tsinghua.edu.cn}.}
\endgroup
\lhead{Preprint; Under Review}
\begin{abstract}

ELBO-based reinforcement learning offers a sampler-agnostic approach to fine-tuning flow matching models with reward feedback. Timestep weighting in ELBO-based RL has large impact on performance, and it also provides a unified view (as we show in this work) to understand prediction losses heuristically chosen in prior work, yet it remains under-researched and is often chosen to inherit pretrain configs. We investigate impacts and dynamics of timestep weighting in ELBO-based RL. We show that effective weighting depends on both the reward landscape and stage of learning. (1) Through experiments on controlled CIFAR image generation, complemented by robotics, we investigate how weighting impacts reward-driven updates across noise levels. (2) Through gradient analysis, we reveal distinct patterns of cross-noise coordination across tasks and their evolution during training. These findings motivate the hypothesis that useful weighting depends on the gap between the policy's current behavior and the behavior favored by the reward. (3) Guided by this analysis, we study simple static weighting, budgeted profile selection, and dynamic schedules that improve performance beyond conventional target choices. Our results establish timestep weighting as an important design choice for flow-matching RL and motivate further research into methods that choose and adapt it throughout learning.
\end{abstract}
\section{Introduction}
\label{sec:intro}
Flow matching supports generative policies for images and robot actions \citep{lipman2023flow}. ELBO-based reinforcement learning (RL) optimizes these policies using prediction-loss surrogates, avoiding expensive likelihood evaluation. Existing approaches differ in their prediction targets: Flow Policy Optimization (FPO) uses noise, FPO++ uses velocity, and Variational GRPO (V-GRPO) uses the clean sample \citep{mcallister2025flow,yi2026flow,tang2026vgrpo}. In a common coordinate, these targets assign different weights to errors across noise levels. Target choice thus implicitly specifies a \emph{timestep weighting}---the relative training emphasis on each noise level, implemented through loss weights or timestep sampling.\footnote{Related work is discussed in Appendix~\ref{app:related}.}

This implicit choice has substantial consequences. For example, the original FPO paper reports unstable image fine-tuning and deteriorating sample quality 
\citep{mcallister2025flow};
in our controlled image experiments, simply replacing velocity weighting with clean-sample ($x_0$) weighting substantially alleviates this phenomenon across several rewards. These observations motivate treating timestep weighting as an important explicit design choice for RL. Whereas pretraining fits a data distribution, RL reinforces reward-preferred changes in an evolving policy. The relevant allocation therefore depends on both the reward and the policy's current behavior. Our central question is: how to distribute training effort across noise levels to support these changes, beyond the few profiles prescribed by conventional targets?

Our experiments on controlled CIFAR image-generation, complemented by robotics, establish that effective weighting is reward- and policy-dependent.
Different rewards and learning stages exhibit substantially different preferences across timestep weightings, both in RL post-training for image generation and from-scratch RL in simulated robotics scenarios. For the reward landscape, we interpret different weightings as emphasis on different noise-resolved policy corrections. To study the learning process, we conduct gradient measurements to reveal distinct coordination patterns: successful image RL post-training adaptation exhibits weak cross-noise agreement, whereas acquiring new behavior for far-away rewards on image or RL from scratch on robotics can involve broader coordination of gradients that changes during training. Together, these findings motivate a \emph{behavioral-gap hypothesis}: useful allocation depends on the remaining discrepancy between current and reward-preferred behavior, and evolves as that discrepancy changes.

Our analysis motivates our design of three simple yet effective strategies: static softmax constructs weights from the pretrained model's initial reward-conditioned gradient profile; naive successive halving selects fixed profiles through short training stages; dynamic schedules adapt allocation during learning. These strategies are shown to improve performance beyond conventional target choices. In particular, in image RL post-training, a high-to-low-noise annealing schedule outperforms the best tested fixed-$\alpha$ weightings on three rewards, demonstrating the value of stage-dependent allocation.

Overall, we (1) identify timestep weighting as a consequential RL design choice, (2) analyze its reward and stage dynamics, and (3) demonstrate gains from simple selection and adaptation strategies, motivating methods that track useful allocations throughout learning. We hope our work can inspire understanding of timestep weighting in ELBO-based RL under different scenarios, and can motivate further study in timestep weighting choices or scheduling.

\section{Preliminaries: Timestep-Weighted Policy Optimization}
\label{sec:theory}

\begin{figure}[t]
\centering
\setlength{\abovecaptionskip}{5pt}
\includegraphics[width=\linewidth]{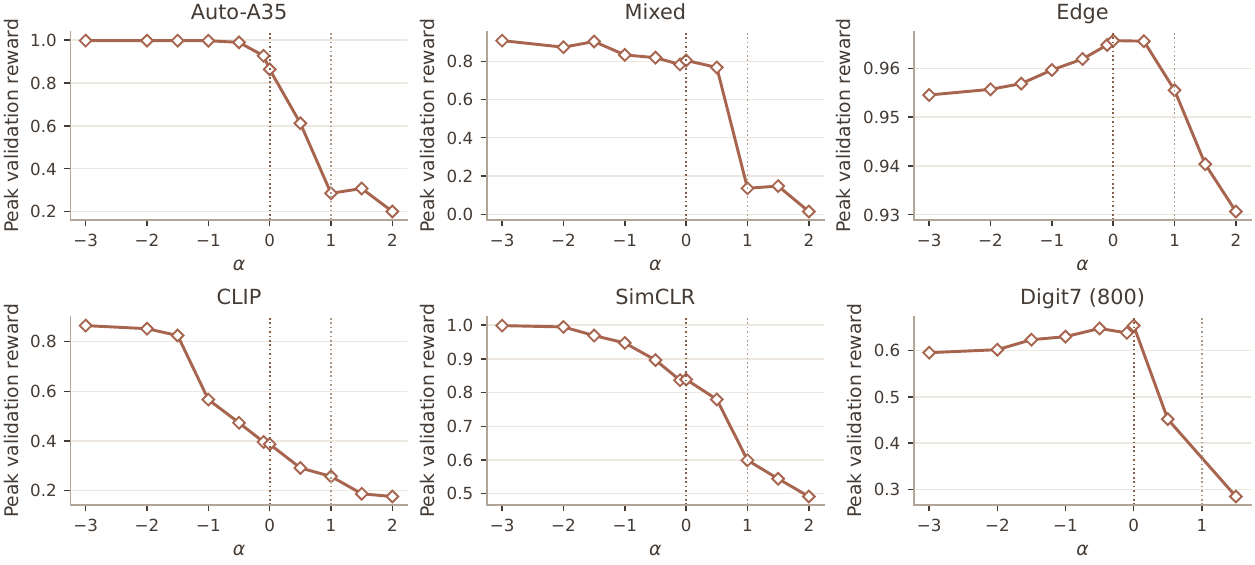}
\caption{CIFAR-10 post-training: mean per-run validation peak under different timestep weighting scheduling (i.e., $\alpha$: smaller $\alpha$ places more emphasis on larger noise levels) for six image rewards. Intuitively, global rewards prefer timestep weighting that leans toward larger noise levels, but our deeper analysis shows that the preference is also strongly related to the behavioral gap between the model and the reward-favored behavior. More details can be found in Sections~\ref{sec:cifar} and~\ref{sec:mechanisms}, Appendices~\ref{app:cifar} and~\ref{app:cifar-results}.}
\label{fig:image_landscape}
\label{fig:landscape}
\end{figure}

\subsection{Timestep weighting and prediction targets}
We use a linear noise-to-data path from Gaussian noise $\epsilon$ to a clean endpoint $x$:
\begin{equation}
 x_t=(1-t)\epsilon+t x,\qquad u=x-\epsilon,\qquad
 \epsilon\sim\mathcal N(0,I),\quad t\in[0,1].
 \label{eq:path}
\end{equation}
Here $t=0$ is noise, $t=1$ is data, and conditioning is implicit. The same velocity output $v_\theta(x_t,t)$ induces $\widehat x=x_t+(1-t)v_\theta$ and $\widehat\epsilon=x_t-tv_\theta$. With per-dimension losses $\ell_v=\|v_\theta-u\|^2/d$, $\ell_{x_0}=\|\widehat x-x\|^2/d$, and $\ell_\epsilon=\|\widehat\epsilon-\epsilon\|^2/d$, these predictions satisfy
\begin{equation}
 \ell_{\rm target}=w_{\rm target}(t)\ell_v,\qquad
 w_v(t)=1,\quad w_{x_0}(t)=(1-t)^2,\quad w_\epsilon(t)=t^2.
 \label{eq:targets}
\end{equation}
where $x_0$ names the clean target. These pointwise identities hold for predictions induced by one velocity output; they do not equate separately parameterized networks.

The three target factors are special cases of a nonnegative velocity-loss coefficient $\boldsymbol{w}(t)$, which we call \textbf{timestep weighting}. A weight on a native target loss is multiplied by $w_{\rm target}(t)$ once to obtain this common-coordinate coefficient. For base time density $q_0$, fixed $f(t)=\E_\epsilon\ell_v$, and $0<Z<\infty$,
\begin{equation}
 \underbrace{\E_{q_0}[\boldsymbol{w}(t)f(t)]}_{\text{explicit weighting}}
 =Z\underbrace{\E_{q_w}[f(t)]}_{\text{profile sampling}},\qquad
 q_w(t)=\frac{q_0(t)\boldsymbol{w}(t)}{Z},\quad Z=\E_{q_0}\boldsymbol{w}(t).
 \label{eq:weight-sampling}
\end{equation}
Thus a profile can be applied explicitly or through sampling. The distribution $q_w$ governs auxiliary training times, not the generation ODE grid, generated samples, or robot environment steps; Appendix~\ref{app:measure-equivalence} gives the measure formulation and finite-sample distinction.

\subsection{Separating profile shape from overall scale}
Normalized bin masses specify profile shape, but equal mass need not imply equal loss magnitude. Most controlled CIFAR studies therefore match initial whole-loss RMS on a frozen coupled bank of per-example, per-bin velocity losses $L_{ib}$:
\begin{equation}
 F_i(q)=\sum_bq_bL_{ib},\quad S(q)=\sqrt{N^{-1}\sum_i F_i(q)^2},\quad
 a(q,g)=g\frac{S(q_{x_0})}{S(q)}.
 \label{eq:calibration}
\end{equation}
Here $q_b$ is bin probability, $N$ is bank size, and $g=1$ unless specified. The bank remains fixed while model losses and gradients evolve. The estimator samples from $q$ and applies only $a(q,g)$, with no second $\boldsymbol{w}$ or $Z$ factor. On the same coupled bank, proportional effective profiles therefore have identical calibrated losses across target coordinates. Appendix~\ref{app:calibration} proves this invariance, and Appendix~\ref{sec:scale} varies shape and scale. Four-candidate $q+Z$ screening and native robotics use their own stated conventions.

\subsection{ELBO-surrogate policy optimization}
PPO/GRPO require policy log probabilities to form an advantage-weighted clipped objective \citep{schulman2017ppo,shao2024deepseekmath}. FPO replaces the costly flow log likelihood with an ELBO-derived negative prediction loss \citep{mcallister2025flow}: using $\mathcal B_\theta\le\log p_\theta$ as a surrogate gives
\begin{equation}
 \underbrace{e^{\log p_\theta-\log p_{\rm old}}}_{r_\theta(x\mid\mathsf c)}
 \;\rightsquigarrow\;
 \underbrace{e^{\mathcal B_\theta-\mathcal B_{\rm old}}}_{\widehat r_\theta(x\mid\mathsf c)}
 \;\approx\; e^{\mathcal L_{\rm old}-\mathcal L_\theta}.
 \label{eq:ratio-bridge}
\end{equation}
The advantage estimator and PPO clipping are retained. Algorithm~\ref{alg:cifar} shows our CIFAR implementation with GRPO advantages and timestep weighting through $q_k$ sampling.

\begin{algorithm}[H]
\caption{Calibrated timestep-weighted CIFAR ELBO-surrogate RL}
\label{alg:cifar}
Adapted from FPO \citep{mcallister2025flow}; $\operatorname{sg}=\operatorname{stopgrad}$.\par
\begin{tabularx}{\linewidth}{@{}rX@{}}
1 & \textbf{for} $k=0,1,\ldots$ \textbf{do} \\
2 & \quad $\theta_{\rm old}\leftarrow\operatorname{sg}(\theta)$. \\
3 & \quad \textbf{Weighting:} $Z_k\leftarrow\E_{q_0}[\boldsymbol w_k(t)]$, $q_k(t)\leftarrow q_0(t)\boldsymbol w_k(t)/Z_k$, $a_k\leftarrow a(q_k,g)$. \\
4 & \quad Collect $x_i\sim p_{\theta_{\rm old}}$, score $R_i$, and compute stopped GRPO advantages $A_i$. \\
5 & \quad For each $x_i$, store $M$ pairs $(t_{im},\epsilon_{im})\sim q_k\times\mathcal N(0,I)$ and old losses $\ell_{v,{\rm old},im}$. \\
6 & \quad Recompute current losses $\ell_{v,\theta,im}$ on the same endpoints, times, and noises. \\
7 & \quad Compute $D_i$ and $r_i$ by Equation~\eqref{eq:rl-mc}, then $\mathcal L_{\rm RL}$ by Equation~\eqref{eq:rl-ppo}. \\
8 & \quad Update $\theta$ once using $\nabla_\theta\mathcal L_{\rm RL}$. \\
9 & \textbf{end for}
\end{tabularx}
\end{algorithm}

Using the per-dimension velocity losses from Section~\ref{sec:theory}, the paired Monte Carlo ratio and clipped objective are
\begin{align}
 D_i&=\frac{a_k}{M}\sum_{m=1}^M
 \left(\ell_{v,{\rm old},im}-\ell_{v,\theta,im}\right),\qquad
 r_i=\exp\!\bigl(\operatorname{clip}(D_i,-c,c)\bigr),
 \label{eq:rl-mc}\\
 \mathcal L_{\rm RL}
 &=-\frac1n\sum_i\min\!\left\{r_iA_i,\operatorname{clip}(r_i,1-\rho,1+\rho)A_i\right\}.
 \label{eq:rl-ppo}
\end{align}
Here $a_k=a(q_k,g)$, $c$ bounds the log surrogate, and $\rho$ is the PPO clip width. Native objectives and implementation details are in Appendices~\ref{app:native-objectives} and~\ref{app:cifar}.

\section{Reward- and Task-Dependent Weighting Preferences}
\label{sec:performance}
Timestep weighting changes how policy updates emphasize reward-relevant corrections across noise levels. This raises two empirical questions: how much does this choice affect performance, and do different rewards and tasks favor different allocations? We investigate these questions using a one-dimensional weighting family in image generation and robot control.

\subsection{Experimental setup and weighting profiles}
We fine-tune a shared unconditional U-Net pretrained on CIFAR-10 (\href{https://huggingface.co/FrankCCCCC/cfm-cifar10-32}{cfm-cifar10-32}), using group-normalized rewards and $q$-sampling. Robot policies learn from scratch through online interaction: FPO on MuJoCo Playground and FPO++ on Isaac Lab \citep{mcallister2025flow,yi2026flow,zakka2025playground,mittal2025isaaclab}. Details are in Appendices~\ref{app:cifar}--\ref{app:robotics}.

Our image rewards target different aspects of generation. Auto-A35 combines automobile semantics with a specified-region red attribute; Mixed is automobile semantics blended with local red high-frequency structure. Edge matches reference-image multiscale edge statistics. CLIP/SimCLR encourage automobile semantics through frozen representations. Digit7 combines handwritten-seven semantics with layout structure beyond CIFAR's pretraining categories. Formulas and constants are in Appendix~\ref{app:rewards}.

We use a scalar $\alpha$ to vary timestep weighting along a predefined high-to-low-noise axis: larger $\alpha$ places relatively more emphasis on lower-noise regions. We anchor this axis using reference profiles appropriate to each setting. For CIFAR and FPO++, $\alpha=0$ and $1$ correspond to $x_0$- and velocity-derived weight profiles, respectively. For native FPO, $\alpha_\epsilon=1$ is the noise-prediction reference, while $\alpha_\epsilon=0$ adds the factor $(1-t)^2$ to its native loss. With $t=0$ denoting noise and $t=1$ data, we use a common power profile applied to each native objective:
\begin{equation}
 w_\alpha(t)\propto(1-t)^{2(1-\alpha)}.
 \label{eq:alpha}
\end{equation}

From experiments, we can see that preferred shapes are largely stable across tested scales (Appendix~\ref{sec:scale}). Supplementary text-to-image experiments following the V-GRPO framework of \citet{tang2026vgrpo} are in Appendix~\ref{app:vgrpo}. Metric definitions are given in Appendix~\ref{app:settings}.

\subsection{Different rewards and tasks favor different profiles}
\label{sec:cifar}
\paragraph{CIFAR post-training favors reward-specific noise emphasis.} In Figure~\ref{fig:image_landscape}, automobile-related and semantic rewards generally favor stronger high-noise emphasis, whereas Edge and Digit7 favor profiles closer to clean-target weighting. Rewards also differ in how sharply a profile should be chosen: Auto-A35 has a broad high-noise plateau, while Edge has a more concentrated peak near the clean-target reference. Thus reward-dependent allocation can improve over simply adopting a standard prediction target.

\begin{figure}[t]
\includegraphics[width=0.9\linewidth]{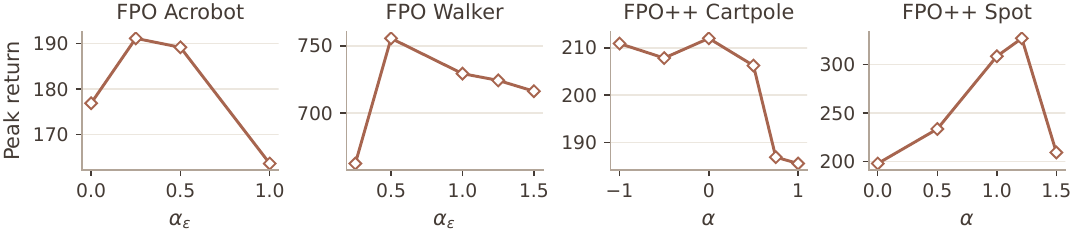}
\caption{Robot control from scratch: mean per-run peak return under different fixed $\alpha$ for FPO and FPO++, each relative to its native target. Acrobot, Walker and Cartpole prefer more emphasis on larger noise levels than native, while Spot prefers slightly smaller noise. More details can be found in Section~\ref{sec:robotics} and Appendices~\ref{app:robotics} and~\ref{app:robotics-results}.}
\label{fig:robotics_landscape}
\label{fig:robotics}
\end{figure}
\paragraph{Robot control from scratch favors task-specific profiles.}
\label{sec:robotics}

Figure~\ref{fig:robotics_landscape} shows distinct task preferences within each native weighting family. Acrobot and Walker benefit from shifting emphasis toward higher noise relative to native FPO. Within FPO++, Cartpole also favors higher-noise emphasis relative to its native reference, whereas Spot favors the native allocation or a modest shift toward lower noise. The useful direction of reweighting therefore depends on the task and its native objective.

\section{Behavioral Gaps and Evolving Cross-Noise Coordination}
\label{sec:mechanisms}
Section~\ref{sec:performance} shows that useful timestep weighting depends on the task. We investigate this dependence through two questions: what corrections does a reward call for at the current policy, and how do those corrections interact across noise levels as training proceeds?

\subsection{Behavioral gaps shape cross-noise learning across tasks and regimes}
\label{sec:field}
\label{sec:gram}

The relevant learning signal depends jointly on the reward and the policy's current capabilities. We study this behavioral gap through a reward-relative field and cross-batch gradient alignment, distinguishing refinement of existing behavior from acquisition of missing behavior.

\paragraph{The behavioral gap between reward and current policy determines the correction.}
Let the current policy define a reference endpoint density $p_{\rm ref}$ and its reward-favored tilt $p_R(x)\propto p_{\rm ref}(x)e^{\beta R(x)}$, for $\beta>0$ and a finite normalizer. Along Equation~\eqref{eq:path}, Gaussian corruption gives $y=x+\lambda\epsilon$, with $\lambda=(1-t)/t$. The reward-relative field
\begin{equation}
 \Delta_\lambda(y)=\log\frac{(p_R*\varphi_\lambda)(y)}{(p_{\rm ref}*\varphi_\lambda)(y)}
 \label{eq:field}
\end{equation}
compares the reward-favored and current distributions at noise level $\lambda$, where $\varphi_\lambda$ has covariance $\lambda^2I$. Its gradient determines the population flow-target correction up to a path-dependent factor (Appendix~\ref{app:fieldproof}). Weighting emphasizes different smoothed views of this correction while preserving $p_R$ for fixed reward, reference, and temperature. Thus the relevant question is not only what a reward measures, but what behavior the current policy must acquire or refine to improve it.

\paragraph{Measuring reproducible learning directions.}
We measure how these reward-driven updates relate across noise bins using parameter gradients on independent on-policy batches. For a batch gradient $G_b$ in bin $b$, write $u_b=G_b/\|G_b\|$. The population cross-batch cosine satisfies
\begin{equation}
 C_{bc}=\mathbb E[u_b^\top u'_c]
       =\mathbb E[u_b]^\top\mathbb E[u_c],
 \label{eq:cross-direction-signal}
\end{equation}
where primes denote an independent batch. The diagonal measures whether a bin's direction persists across samples; off-diagonal entries combine this directional reliability with agreement between bins. Positive blocks indicate reproducible, mutually aligned directions, while negative blocks indicate opposition. Weak entries can arise from variable directions or weak alignment between stable directions. The heatmap therefore reveals the strength and organization of reproducible directional signal, rather than raw gradient magnitude. Independent batches separate this structure from agreement induced by shared examples; measurement details and the shared-fluctuation analysis are in Appendix~\ref{app:cifar-gradients}.

\begin{figure}[t]
\centering
\includegraphics[width=.82\linewidth]{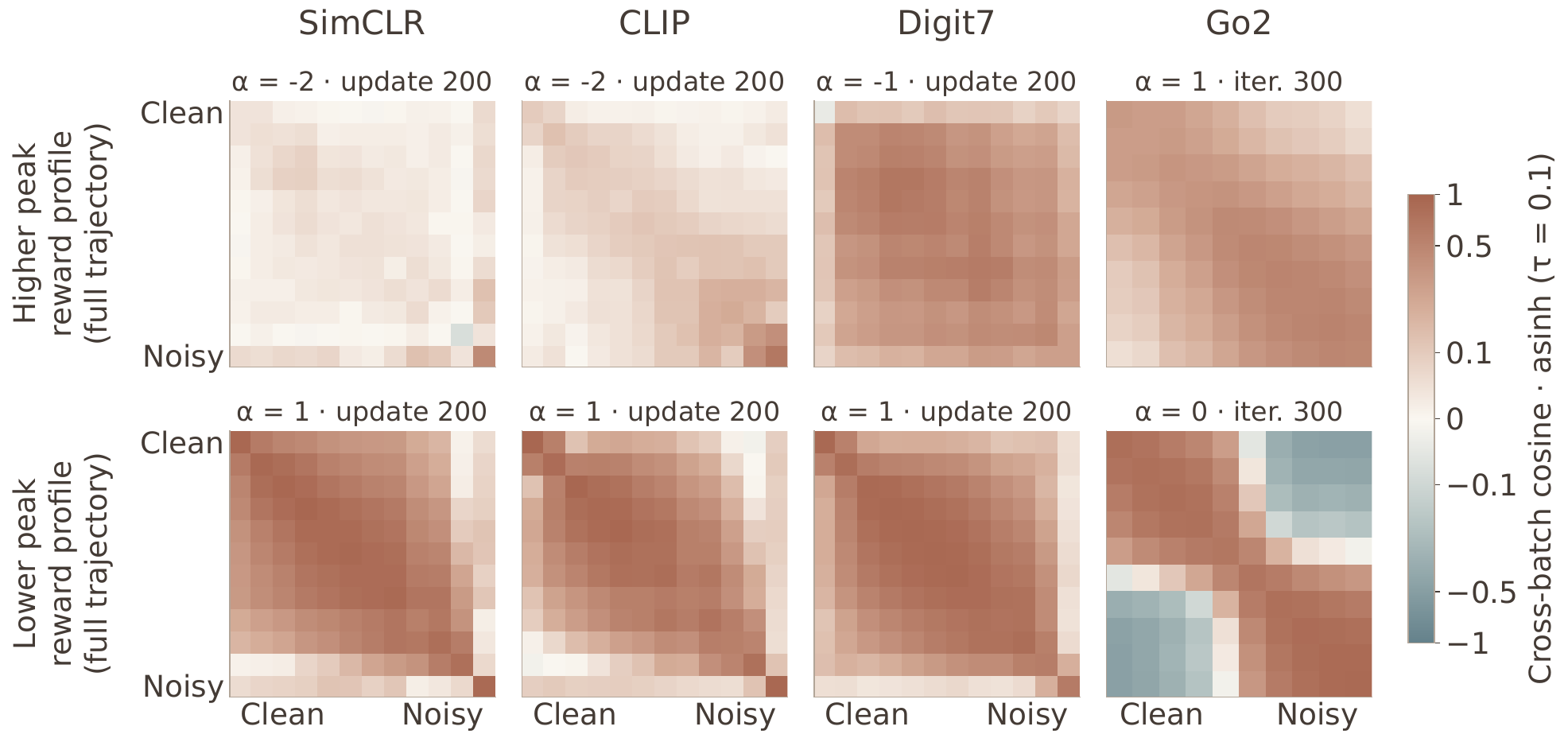}
\caption{Cross-batch gradient cosine between noise bins at an early checkpoint; top/bottom rows are the better/worse $\alpha$ of each task. Post-trained CIFAR rewards show selective agreement among a few bins, while from-scratch Go2 benefits from broad agreement across noise levels. More details can be found in Section~\ref{sec:field} and Appendix~\ref{app:approved-heatmaps}.}
\label{fig:geometry}
\end{figure}

\paragraph{Refining existing behavior differs from acquiring missing behavior.}
Figure~\ref{fig:geometry} reveals a regime-dependent contrast: successful CIFAR post-training exhibits selective cross-noise agreement, whereas acquiring new behavior can benefit from broader coordination. For example, the better Go2 profile preserves positive coordination where its poorer counterpart has opposing noise groups; Digit7 occupies an intermediate regime, starting from a pretrained policy but requiring new structure.

This motivates our \emph{behavioral-gap hypothesis}: weighting should reflect how much reward-favored behavior the policy must acquire, rather than the reward alone.

\subsection{Stage-dependent coordination and learning allocation}
\label{sec:evolving}
\begin{figure}[t]
\centering
\includegraphics[width=.80\linewidth]{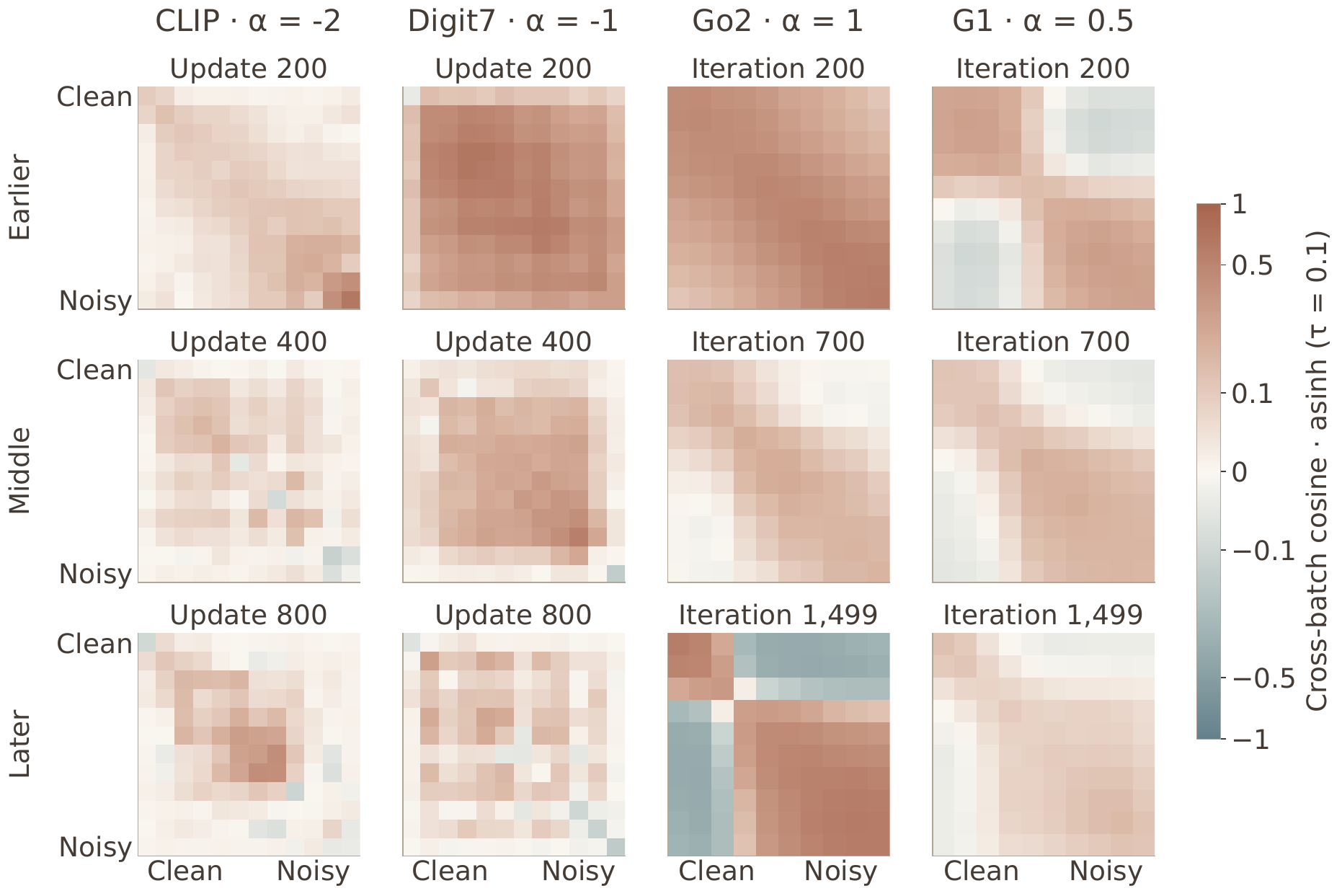}
\caption{Cross-batch gradient cosine between noise bins over training for fixed-$\alpha$ runs. Post-training (CLIP) gradually shifts learning toward middle and low noise, while from-scratch training (Go2, G1) starts with broad agreement that later weakens. More details can be found in Section~\ref{sec:evolving} and Appendix~\ref{app:approved-heatmaps}.}
\label{fig:training-evolution}
\end{figure}

Learning requirements also change across training stages, and their evolution differs between post-training and training from scratch. We track these changes to understand why a useful weighting may need to change during training. To connect their evolution to weighting, consider a local additive surrogate with bin losses $L_b$, gradients $g_b=\nabla_\theta L_b$, and update $\delta\theta=-\eta\sum_b\omega_b g_b$. To first order,
\begin{equation}
 \Delta L_c=-\eta\sum_b\omega_b g_c^\top g_b+O(\|\delta\theta\|^2).
 \label{eq:local-crossbin}
\end{equation}
Aligned gradients permit an update from one bin to reduce another bin's loss; opposing gradients create a local tradeoff. Weighting controls each contribution, while the changing policy changes the gradients themselves. This motivates tracking their normalized directional relationships along fixed-profile trajectories (Figure~\ref{fig:training-evolution}); Appendix~\ref{app:local-gram} gives the preconditioned analysis.

\paragraph{Post-training shifts toward middle- and low-noise learning.}
Even a fixed high-noise-emphasizing profile can support this shift, as CLIP illustrates, suggesting that greater middle- and low-noise allocation becomes useful once early high-noise gains saturate.

\paragraph{Training from scratch begins with broader coordination.}
Early behavior acquisition exhibits broad cross-bin agreement that later weakens or reorganizes, suggesting a transition toward more selective refinement; Digit7, as an out-of-distribution reward for a CIFAR-trained model, also shows a related pattern.

These regime-dependent changes motivate the opposite weighting schedules tested in Section~\ref{sec:path}.

\section{Strategies for Choosing and Adapting Timestep Weights}
\label{sec:methods}

Following the analysis in Section~\ref{sec:mechanisms}, we construct three simple yet effective methods for timestep weighting: static weights from initial reward-conditioned signals, profile selection through short-run performance, and allocation adapted to the stage of learning.

\subsection{Static softmax complements target and fixed-profile choice}
Static softmax provides a reward-conditioned alternative to target-derived profiles and fixed exponents. For endpoint $i$ and time bin $b$, let $z_{i,b}=-2A_i(v_\theta-u_i)/d$ be the reward-weighted output-space descent signal for the per-dimension velocity loss. To assess initial reward-conditioned signal reliability, we measure within-bin output-signal consistency, distinct from the cross-bin parameter-gradient alignment in Section~\ref{sec:mechanisms}:
\begin{equation}
 c_b=\frac{\|\E z_{i,b}\|^2}{\E\|z_{i,b}\|^2+\delta}.
 \label{eq:coherence}
\end{equation}
Computed at the initial checkpoint, $c_b$ is larger when examples within bin $b$ suggest consistent output changes rather than canceling; this is similar to the signal-to-noise ratio of this bin.

\begin{figure}[t]
\centering
\includegraphics[width=\linewidth]{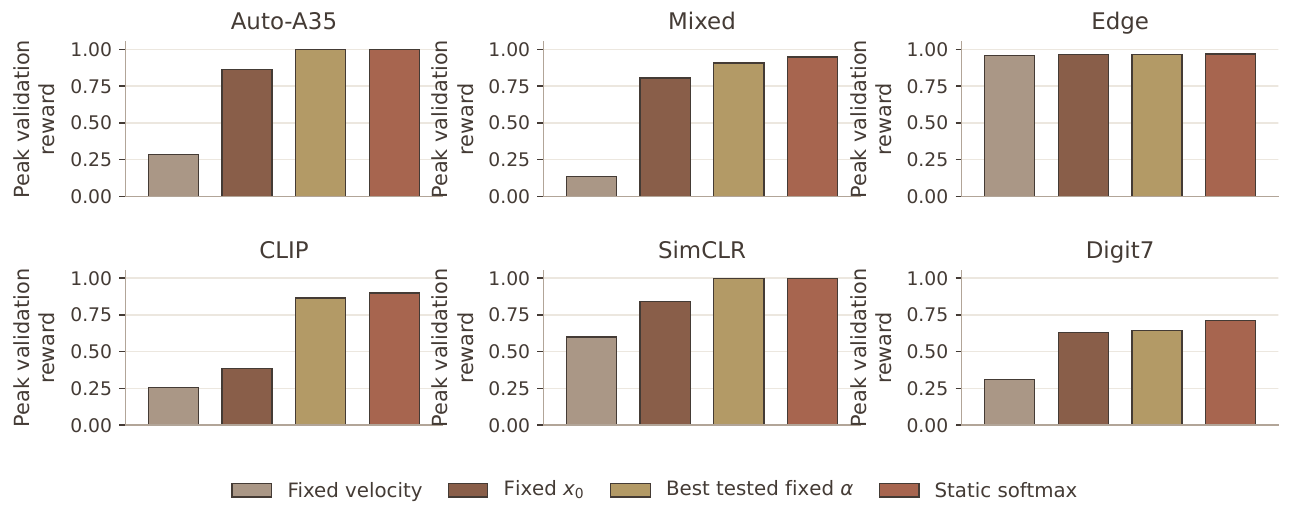}
\caption{Six CIFAR rewards: mean per-run validation peak of static softmax, compared with target-derived references and fixed-$\alpha$ profiles. Static softmax improves over the best fixed $\alpha$ on CLIP, Mixed and Digit7 and matches it on the others. More details can be found in Appendix~\ref{app:methods}.}
\label{fig:static}
\end{figure}

We standardize the estimated coherence across bins with its population standard deviation and deploy
\begin{equation}
 d_b=\frac{c_b-\overline c}{\operatorname{std}_{\rm pop}(c)},\qquad
 w_b=B\operatorname{softmax}(\eta_{\rm norm}d)_b,\qquad q_b=w_b/B.
 \label{eq:static}
\end{equation}
The scalar follows Equation~\eqref{eq:calibration}; concentration is selected by mean per-run validation peak. Static softmax improves beyond the best tested fixed exponent on CLIP, Mixed, and Digit7, while other rewards show plateaus or little change (Figure~\ref{fig:static}). Greater concentration is not uniformly better. Appendix~\ref{app:methods} analyzes concentration and online refreshes.

\subsection{Successive halving screens favorable regions efficiently}
\label{sec:halving}

\paragraph{Task-specific behavioral-gap regularity motivates successive halving.} Task-dependent preferences analyzed in Section~\ref{sec:gram} motivate selecting profiles through direct performance feedback. Successive halving screens their observed learning outcomes \citep{jamieson2016nonstochastic}. We apply the standard successive halving method, eliminating the worse-performing half of the candidate $\alpha$s after $2^iN$ steps in round $i$.

\begin{figure}[t]
\centering
\includegraphics[width=0.9\linewidth]{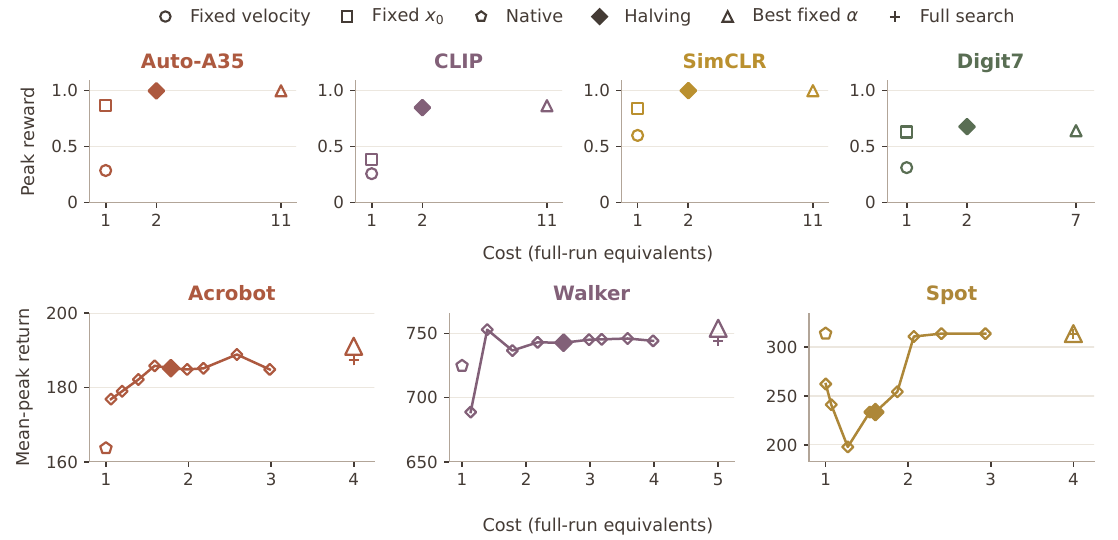}
\caption{Successive halving: peak performance versus training cost for CIFAR (top) and robotics (bottom). At about half the cost of trying all candidates, halving matches or beats the native target in most tasks, but misses Spot's best profile. More details can be found in Section~\ref{sec:halving} and Appendices~\ref{app:halving}--\ref{app:halving-robotics}.}
\label{fig:halving}
\end{figure}

We plot reward versus training budget in Figure~\ref{fig:halving}. As shown, selected winners can perform as well as or better than native-target references across image and robotic rewards at $2$--$3$ full-run budgets.

The effectiveness of successive halving further suggests that \textbf{reward task preference for timestep weighting is relatively stable across the training stage}.

\subsection{Stage-aware weighting improves the learning trajectory}
\label{sec:path}
The regime-dependent dynamics in Section~\ref{sec:evolving} motivate opposite schedules: CIFAR starts with high-noise emphasis and increases $\alpha$ as learning shifts toward middle and lower noise; robotics starts with broader, relatively lower-noise coverage and decreases $\alpha$ as behavior acquisition gives way to refinement. The latter direction tests our hypothesis that late from-scratch learning becomes closer to post-training.

\begin{figure}[htbp]
\centering
\includegraphics[width=0.9\linewidth]{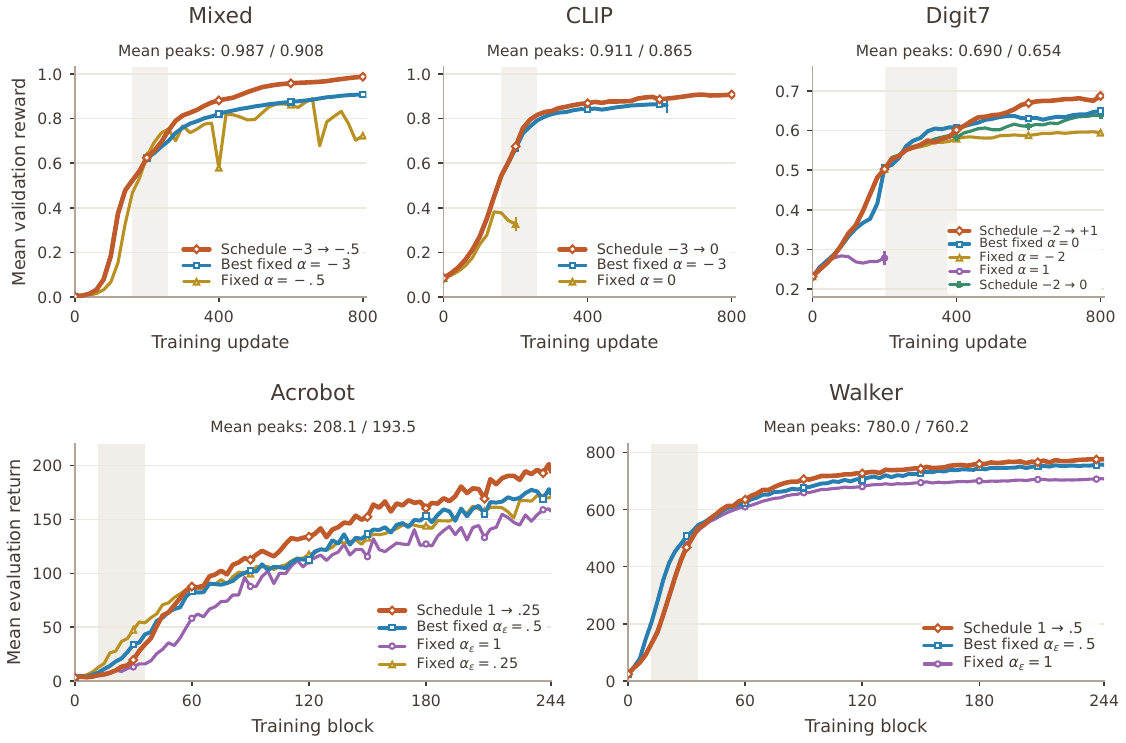}
\caption{Dynamic versus fixed timestep weighting on CIFAR (top) and robotics (bottom). CIFAR schedules shift emphasis from larger to smaller noise, and robotics schedules the opposite way; in each task the schedule reaches a higher mean peak than the best fixed $\alpha$. More details can be found in Section~\ref{sec:path} and Appendices~\ref{app:schedules} and~\ref{app:robotics-dynamic}.}
\label{fig:dynamic-paths}
\label{fig:mixed-path}
\label{fig:robotics-dynamic}
\end{figure}

On Mixed, CLIP, and Digit7, the schedules each exceed the mean per-run validation peak of their corresponding best tested fixed-$\alpha$ weighting (Figure~\ref{fig:dynamic-paths}, top). Digit7 also shows that a profile poor from initialization can become useful later: velocity weighting benefits training after the high-noise-first phase. On Acrobot and Walker, the schedules exceed the mean per-run peaks of both fixed endpoints and the best tested fixed-sweep references (bottom; Appendix~\ref{app:robotics-dynamic}).

Together, these schedules illustrate how timestep allocation can adapt to the evolving gap between current and reward-preferred behavior.

\section{Conclusions and Discussions}
\label{sec:conclusion}
\subsection{Conclusions}
Timestep weighting is an important design choice in ELBO-based flow-matching RL, extending beyond conventional prediction targets.

In Section~\ref{sec:performance}, we show that different rewards (e.g., global vs.\ fine-grained) and different tasks (e.g., RL post-training vs.\ RL from scratch) have different timestep weighting preferences.

In Section~\ref{sec:mechanisms}, we analyze details of such preference from \textbf{behavioral gap regularity} and \textbf{stage-dependent coordination}. From gradient alignment analysis, we find that, for RL from scratch tasks or when the reward is far from the pretrained model, RL learning involves a co-learning phase where gradient alignment is higher across noise timestep bins (behavioral gap regularity), and vice versa for close rewards. Our stage-dependent analysis further shows that timestep scheduling matters and should be task- and regime-specific.

Our analysis can motivate effective reweighting methods. As shown in Section~\ref{sec:methods}, we propose (1) static softmax based on gradient SNR; (2) successive halving for task-specific properties; and (3) alpha scheduling for different training stages. These methods represent different categories of timestep weighting mechanisms, and we believe there is much room for further improvement. 

Overall, we have shown that timestep weighting is important in ELBO-based RL training, and has its specific properties and mechanisms that are worth further analyzing.

\subsection{Limitations and Future Work}

Because of the complexity of ELBO-based RL, we mainly ablate restricted families of timestep reweighting. We believe timestep reweighting is very important, and there is plenty of room for timestep scheduling in RL training in future work. For example, on some robotics environments, these methods are on par with or better than fixed-$\alpha$ configurations by only a small margin. Future work inspired by our analysis may further investigate better timestep scheduling in RL training.

Moreover, our explanation is primarily gradient-based and local.  Our analysis leans toward how the model learns different noise bins during the optimization process, while other types of analysis are possible, for example, scaling-law-style studies that directly analyze the final performance of each bin after training.
\clearpage
\subsection*{AI use statement}
We used large language models to assist with manuscript drafting, editing and coding. The authors take responsibility for the final manuscript, claims, code, and reported results.

\subsection*{Reproducibility statement}
We will publicly release all code used for training, evaluation, analysis, and visualization. The appendices provide theoretical derivations, reward definitions, experimental settings, and detailed results to support reproduction. As the contribution centers on analysis of training objectives and dynamics, the planned release focuses on code and configurations and does not include model checkpoint weights.

\subsection*{Ethics statement}
This work studies optimization in image generation and simulated robot control using existing datasets and evaluation tasks. We identify no additional ethical concerns specific to this study beyond the established considerations associated with generative modeling and reinforcement learning. 

\clearpage
\bibliographystyle{iclr2027_conference}
\bibliography{references}
\clearpage
\appendix
The appendices begin with related work, followed by theoretical and empirical analyses. Model and evaluation settings are in Appendix~\ref{app:settings}, supplementary result summaries in Appendix~\ref{app:full-results}, and the complete robotics heatmap atlas appears last.

\section{Related Work}
\label{sec:related}
\label{app:related}
\paragraph{Weighting as a training design choice.}
The dependence of a squared-error objective on prediction parameterization is part of the broader diffusion design space \citep{ho2020ddpm,kingma2021vdm,kingma2023elbo,karras2022edm}. P2 and Min-SNR show how emphasizing particular noise levels changes practical learning \citep{choi2022p2,hang2023minsnr}. Our target conversion places the evaluated flow-RL implementations in common velocity coordinates, and our calibration makes one definition of scale explicit. The reward-relative field then describes the distributional change relevant to RL, whose endpoint law and useful update geometry differ across tasks and policy states.

\paragraph{The role of the surrogate.}
Gaussian-transition methods such as DDPO and DPOK exploit likelihoods along a stochastic denoising trajectory \citep{black2024ddpo,fan2023dpok}. Flow-GRPO constructs stochastic transitions from a flow model \citep{flowgrpo2025}. FPO uses flow-matching loss differences, while V-GRPO constructs an atomic-action ELBO ratio surrogate \citep{mcallister2025flow,tang2026vgrpo}. These are prominent constructions rather than an exhaustive taxonomy. Our results concern the evaluated ELBO/matching-loss route, with FPO++ providing the robot-control implementation \citep{yi2026flow}. Weight placement relative to exponentiation and clipping is part of the surrogate, which is why the common measure view is accompanied by implementation-specific gradient analysis.

\paragraph{Distributional targets and local optimization.}
KL-regularized reward tilts also underlie preference-learning formulations such as DPO and its diffusion adaptation \citep{rafailov2023dpo,wallace2023diffusiondpo}. Our conditional smoothed-density identity gives a noise-dependent representation of such a tilt. Its realization through an advantage-weighted online surrogate is a separate optimization question. The empirical contribution establishes reward-dependent performance and simple useful choices within the allocation space. Initial responses, cross-scale gradients, and path interventions connect this allocation space to reward- and stage-dependent learning.

\paragraph{Timestep tasks and specialization.}
Flow Matching learns vector fields along probability paths \citep{lipman2023flow}. ANT, DeMe, and TimeStep Master address negative transfer or timestep specialization \citep{go2023negative,ma2025deme,zhuang2025timestepmaster}. These training-time allocation questions become reward- and stage-dependent during policy optimization, as both the desired correction and the policy distribution evolve.

\paragraph{Timestep sampling schedules and curricula.}
Timestep sampling and weighting are also tuned for flow and diffusion pretraining, for example through logit-normal sampling in rectified-flow transformers \citep{esser2024sd3}, learned per-noise-level loss weighting \citep{karras2024edm2}, and adaptive or acceleration-oriented non-uniform sampling \citep{kim2024adaptivetimestep,wang2024speed}. Curriculum approaches further change this emphasis during training, ordering denoising tasks by difficulty \citep{kim2024curriculum} or switching from middle-biased to uniform sampling \citep{sun2026curriculumsamplingtwophasecurriculum}. These methods target a fixed data distribution. In RL, the useful allocation additionally depends on the reward and on the evolving gap between current and reward-preferred behavior, which motivates our reward- and stage-dependent schedules.

\paragraph{Complementary adaptation and selection routes.}
Human-feedback fine-tuning, differentiable reward optimization, and Advantage Weighted Matching provide complementary connections between rewards and generative-model training \citep{lee2023humanfeedback,clark2024draft,xue2025awm}. Existing V-GRPO self-normalization is an example of implicit timestep weighting. Our frozen reward-conditioned profiles and directed schedules make allocation explicit, while standard successive halving \citep{jamieson2016nonstochastic} supplies budgeted selection from observed training outcomes.

\clearpage
\section{Theory: Coordinates, Reward-Relative Fields, and Shared Updates}
\label{app:theory}
\subsection{Prediction-target conversion}
For the path in Equation~\eqref{eq:path}, the endpoint and noise associated with a velocity $u$ satisfy $x=x_t+(1-t)u$ and $\epsilon=x_t-tu$. Substituting the model output gives
\begin{equation}
 \widehat x-x=(1-t)(v_\theta-u),\qquad
 \widehat\epsilon-\epsilon=-t(v_\theta-u).
\end{equation}
Squaring proves Equation~\eqref{eq:targets}; division by a common dimension factor preserves the identities. These are pointwise relations for predictions induced by the same velocity output. Architectural parameterization and preconditioning may introduce additional factors, which belong to the implemented objective.

The robotics noise coefficient is $s=1-t$, whereas the Gaussian convolution standard deviation in our field is $\lambda=(1-t)/t$. In particular, native-$\epsilon$ FPO with weights proportional to $s^{2(1-\alpha_\epsilon)}$ has effective velocity-coordinate profile
\begin{equation}
 \frac{s^{2(1-\alpha_\epsilon)}}{\E[s^{2(1-\alpha_\epsilon)}]}(1-s)^2.
 \label{eq:fpo-effective}
\end{equation}
The factor $(1-s)^2=t^2$ is intrinsic to its noise target. No constant shift in $\alpha$ absorbs this factor into the velocity-based power family. FPO and FPO++ therefore share an allocation question, with distinct native baselines.

\subsection{Native objectives and weight placement}
\label{app:native-objectives}

For the calibrated CIFAR profile at update $k$, let $Z_k=\mathbb E_{q_0}[\boldsymbol w_k(t)]$, $q_k=q_0\boldsymbol w_k/Z_k$, and $a_k=a(q_k,g)$. The population loss is
\begin{equation}
 \mathcal L^{\boldsymbol w_k}_\vartheta(x_i)
 :=\frac{a_k}{Z_k}\E_{t\sim q_0,\epsilon}
 \!\left[\boldsymbol w_k(t)\ell_{v,\vartheta}(x_i,t,\epsilon)\right]
 =a_k\E_{t\sim q_k,\epsilon}
 \!\left[\ell_{v,\vartheta}(x_i,t,\epsilon)\right],
 \label{eq:rl-weighted-population}
\end{equation}
Here $\ell_{v,\vartheta}$ is the per-dimension velocity prediction loss. Our finite-sample estimator uses the right-hand sampling form; Equation~\eqref{eq:weight-sampling} establishes the population identity.

For the FPO power family, write $\sigma=1-t$. At $\alpha_\epsilon=0$, the native squared-error factor obeys
\begin{equation}
 \sigma^2\ell_\epsilon=(1-\sigma)^2\ell_{x_0}
 =\sigma^2(1-\sigma)^2\ell_v,
\end{equation}
for predictions induced by the same velocity field and a common per-dimension reduction. At $\alpha_\epsilon=1$ the reference is $\ell_\epsilon$. FPO++ instead anchors $\alpha=0,1$ to the $x_0$- and velocity-derived weight profiles. These are loss-factor identities and reference-profile correspondences; they do not replace the native nonlinear objective or its weight placement.

\paragraph{Objectives and weight placement.}
Write $\Delta\ell=\ell_{\rm old}-\ell_\theta$ on paired endpoints, times, and noise draws, and define the PPO maximization surrogate
\begin{equation}
 \mathcal P_\rho(r,A)=\min\{rA,\operatorname{clip}(r,1-\rho,1+\rho)A\}.
\end{equation}
The calibrated CIFAR objective is exactly Equations~\eqref{eq:rl-mc}--\eqref{eq:rl-ppo}: $t_{im}\sim q_w$, $D_i=a(q_w,g)M^{-1}\sum_m\Delta\ell_{v,im}$, and $\mathcal L_{\rm RL}=-n^{-1}\sum_i\mathcal P_\rho(\exp[\operatorname{clip}(D_i,-c,c)],A_i)$. The profile is carried by sampling; the frozen-bank whole-loss scalar lies inside the exponent. Appendix~\ref{app:cifar} gives the training constants.

FPO~\citep{mcallister2025flow} instead uses its native noise-target residual and explicit profile on the executed support. Suppressing native scalar constants and log-ratio guards, its actor objective has the form
\begin{equation}
 D_i^\epsilon=\frac1M\sum_m w_{\alpha_\epsilon}(t_{im})\Delta\ell_{\epsilon,im},\qquad
 \mathcal L_{\rm FPO}=-\frac1n\sum_i\mathcal P_\rho(e^{D_i^\epsilon},A_i).
 \label{eq:fpo-objective}
\end{equation}
Thus CIFAR and FPO share aggregation before exponentiation and PPO clipping, while retaining their respective residual, sampling, and scale conventions. The noise-target profile $w_{\alpha_\epsilon}\propto(1-t)^{2(1-\alpha_\epsilon)}$ acquires the further $t^2$ velocity-coordinate factor in Equation~\eqref{eq:fpo-effective}; equal numerical exponents need not imply equal allocation.

For FPO++~\citep{yi2026flow}, let $\ell^{++}=\|v_\theta-u\|^2/\sqrt d$ and let $\widetilde\ell^{++}$ denote the loss after its native loss clamps (including the negative-advantage clamp on the current loss). With clamped advantage $\widetilde A$, its placement is
\begin{align}
 r_{im}^{++}&=\exp\!\left[C_{\rm STE}(\widetilde\ell^{++}_{{\rm old},im}-\widetilde\ell^{++}_{\theta,im})\right],\nonumber\\
 \mathcal L_{\rm FPO++}&=\frac1{nM}\sum_{i,m}w_\alpha(t_{im})\,
 \mathcal P_{\rm TR}(r_{im}^{++},\widetilde A_i;\rho,\mathrm{mode}).
 \label{eq:fpopp-objective}
\end{align}
Here $C_{\rm STE}$ is the native upper difference clamp with a straight-through gradient, and $\mathcal P_{\rm TR}$ denotes FPO++'s native per-sample actor-loss proxy, parameterized by its clip width and trust-region mode. The trust-region proxy retains the native FPO++ loss, distinct from the PPO minimum above. Weighting multiplies the proxy after per-draw exponentiation and trust-region processing; it is not inserted into the exponent.

\begin{table}[htbp]
\centering\small
\caption{Weight placement in the evaluated objectives. Equations~\eqref{eq:rl-mc}--\eqref{eq:rl-ppo}, \eqref{eq:fpo-objective}, and~\eqref{eq:fpopp-objective} define the operations.}
\label{tab:placement}
\begin{tabularx}{\linewidth}{@{}lXX@{}}
\toprule
Objective & Profile placement & Residual and reduction \\
\midrule
Calibrated CIFAR & Sample $t\sim q_w$; scalar $a$ inside exponent & Velocity error / $d$; aggregate, exponentiate, PPO clip \\
FPO & Explicit $w_{\alpha_\epsilon}$ inside exponent & Noise error; aggregate, exponentiate, PPO clip \\
FPO++ & Explicit $w_\alpha$ outside per-draw proxy & Velocity error / $\sqrt d$; process each draw, then average \\
\bottomrule
\end{tabularx}
\end{table}
Moving weights across exponentiation changes the objective. Likewise, Equation~\eqref{eq:weight-sampling} equates fixed-integrand linear expectations, not finite-Monte-Carlo exponentiated or clipped objectives: changing the sampling law changes the distribution of the aggregated loss difference. These distinctions can affect optimization and are retained in each implementation.

\subsection{Reference-frame invariance of whole-loss calibration}
\label{app:calibration}
Coefficient mass is a convenient way to represent a profile, but our experiments aim to control the magnitude of the weighted loss itself. Different noise levels have different residual magnitudes, so equal coefficient mass does not imply equal loss magnitude. Moreover, independently normalizing weights in different target coordinates can assign different scales to the same effective shape. We therefore define scale through a positive homogeneous functional $S$ of the whole effective loss on a common reference bank. If $h$ denotes the effective coefficient and $F(h)$ its bank-loss vector, the calibrated shape representative is
\begin{equation}
 \widehat h=\frac{h}{S(F(h))},\qquad
 F(h)=S(F(h))F(\widehat h),\qquad S(F(\widehat h))=1.
 \label{eq:shape-scale-definition}
\end{equation}
This construction applies when $S(F(h))>0$ and uses the linearity of $F$ in $h$. Proportional effective profiles produce the same $\widehat h$. We can then vary shape at a fixed calibrated scale, or vary scale while keeping the profile fixed.

For coupled bank representations, the calibrated equality is
\begin{equation}
 gC_{x_0}\frac{F_i^a}{S(F^a)}=gC_{x_0}\frac{F_i^b}{S(F^b)}.
 \label{eq:coordinate-invariance}
\end{equation}
The scale unit belongs to the effective weighted loss. Let $\nu$ be a common time measure and $q_a=dQ_a/d\nu$ the sampling density in target representation $a$. For corresponding predictions satisfying $\ell_a=k_a\ell_{\rm ref}$, the common-coordinate coefficient is
\begin{equation}
 h_a=q_aw_ak_a,\qquad F_i^a=\int h_a(t)\ell_{{\rm ref},i}(t)\,d\nu(t).
\end{equation}
To implement a prescribed effective coefficient $h$, set $w_a^*=h/(q_ak_a)$ on its supported domain. If this native weight is independently normalized by $Z_a=\E_{Q_a}w_a^*$, its effective coefficient becomes $h/Z_a$. Thus native mean-one normalization preserves shape but introduces a coordinate-dependent overall mass. Normalizing $w_a$ alone cannot remove that difference.

Suppose $h_a=c h_b$ with a constant $c>0$ on the common support, and retain the same examples, corruptions, loss reduction, and corresponding model predictions. Then $F_i^a=cF_i^b$. Every positive homogeneous scale functional with $S(cF)=cS(F)$, positive on the bank vectors being calibrated, gives
\begin{equation}
 \widetilde F_i^a=gC\frac{F_i^a}{S(F^a)}
 =gC\frac{cF_i^b}{cS(F^b)}=\widetilde F_i^b.
\end{equation}
The calibration bank is frozen at the initial model; its normalizer is a fixed scalar during training for each deployed profile. If correctly coupled pointwise losses preserve $F^a(\theta_a)=cF^b(\theta_b)$ with the same $c$ at subsequent corresponding model states, the frozen scalars preserve that functional identity after calibration. Differently parameterized networks need not follow the same optimization trajectory.

The image calibration uses the aggregate loss vector
\begin{equation}
 F_i(q)=\sum_{b=1}^{12}q_bL_{ib},\quad
 S(F)=\sqrt{\frac1{512}\sum_{i=1}^{512}F_i^2},\quad
 C=C_{x_0}=S(F(q_{x_0})).
\end{equation}
Here $S(q)=S(F(q))$, as defined in Equation~\eqref{eq:calibration}. Consequently $S(gC F/S(F))=gC$ for every nonzero initial bank vector. Aggregation over bins precedes squaring over examples, using the common 512-example bank and $x_0$ anchor. Across different effective shapes, this matches only the initial aggregate objective RMS: it neither makes their loss functions equal nor fixes their relative scales throughout training.

The invariance carries the same bank, anchor, and normalization statistic across target coordinates. With the same finite draws and pointwise target conversion, proportional sampled losses remain identical after calibration. This equality carries through the same subsequent exponentiation, clipping, and loss placement. Changing the sampling law or random pairs can preserve a fixed-integrand population integral without preserving its finite Monte Carlo realization or the full nonlinear RL objective. Gradient norms, clipping activity, and Adam displacements remain outcomes of training.

\paragraph{A rest-mass analogy for the common calibration.}
Target coordinates provide different reference frames for the same effective allocation. A common frozen bank, $x_0$ anchor, and homogeneous whole-loss normalizer remove proportional coordinate factors, like a rest-mass-style calibration to a shared reference. Matching this initial RMS leaves allocation, subsequent gradients, and optimizer steps free to differ.

\subsection{KL tilt and the conditional reward field}
\label{app:fieldproof}
Fix a reference endpoint distribution $p_{\rm ref}$, reward $R$, and inverse temperature $\beta>0$. If $0<Z_R=\E_{p_{\rm ref}}e^{\beta R}<\infty$, the KL-regularized problem has target
\begin{equation}
 \max_p\{\E_pR-\beta^{-1}\KL(p\|p_{\rm ref})\},\qquad
 p_R(x)=p_{\rm ref}(x)e^{\beta R(x)}/Z_R.
 \label{eq:tilt}
\end{equation}
For distributions where the following expectations and divergences are well-defined, the identity below proves the optimum. The field depends on both reward and reference: behavior already represented by the generator presents a different adaptation problem from rarely generated structure.

\begin{equation}
 \Delta_\lambda(y)=\log\E_{\rm ref}[e^{\beta R(X)}\mid X+\lambda\epsilon=y]-\log Z_R.
 \label{eq:conditional}
\end{equation}

For any $p$ absolutely continuous with respect to $p_{\rm ref}$,
\begin{equation}
 \KL(p\|p_R)=\KL(p\|p_{\rm ref})-\beta\E_p R+\log Z_R.
\end{equation}
Rearranging shows that the objective in Equation~\eqref{eq:tilt} equals $\beta^{-1}\log Z_R-\beta^{-1}\KL(p\|p_R)$, which is maximized at $p_R$. The assumptions $\beta>0$ and $0<Z_R<\infty$ ensure a valid target distribution.

Write $p_\lambda=p*\varphi_\lambda$. Directly substituting the tilt into the convolution yields
\begin{equation}
 \frac{p_{R,\lambda}(y)}{p_{{\rm ref},\lambda}(y)}
 =\frac{\int p_{\rm ref}(x)e^{\beta R(x)}\varphi_\lambda(y-x)\,dx}
 {Z_R\int p_{\rm ref}(x)\varphi_\lambda(y-x)\,dx}
 =\frac{\E_{\rm ref}[e^{\beta R(X)}\mid Y=y]}{Z_R},
\end{equation}
which proves Equation~\eqref{eq:conditional}. For small $\beta$, when the required moments and expansion exist,
\begin{equation}
 \Delta_\lambda(y)=\beta\bigl(\E_{\rm ref}[R(X)\mid Y=y]-\E_{\rm ref}R\bigr)+O(\beta^2).
\end{equation}
The conditional reward in this expression is averaged under the reference posterior of the clean endpoint. Gaussian smoothing of a bare reward function would omit that dependence.

Differentiating the Gaussian kernel gives the smoothed-score identity
\begin{equation}
 \nabla_y\log p_\lambda(y)=\frac{\E_p[X\mid Y=y]-y}{\lambda^2}.
 \label{eq:tweedie}
\end{equation}
Meanwhile, $\epsilon=(x_t-tX)/(1-t)$ implies
\begin{equation}
 u_p^*(x_t,t)=\E_p[X-\epsilon\mid x_t]
 =\frac{\E_p[X\mid Y=x_t/t]-x_t}{1-t}.
\end{equation}
For differentiable smoothed densities and finite conditional first moments, the population conditional flow targets obey, for $0<t<1$,
\begin{equation}
 u_R^*(x_t,t)-u_{\rm ref}^*(x_t,t)
 =\frac{1-t}{t^2}\left.\nabla_y\Delta_{\lambda(t)}(y)\right|_{y=x_t/t}.
 \label{eq:velocityfield}
\end{equation}
Subtract the expressions for $p_R$ and $p_{\rm ref}$, apply Equation~\eqref{eq:tweedie}, and use $\lambda^2/(1-t)=(1-t)/t^2$ to obtain Equation~\eqref{eq:velocityfield}. This proof uses conditional first moments and differentiation under the convolution at $0<t<1$.

Conditioning on a prompt or observation $c$ gives the same identities with $p(x\mid c)$ and $R(x,c)$. For a robot policy, an observation-conditioned value or advantage tilt describes a local policy-improvement target; the observation visitation distribution determines how these conditional problems are sampled. This is distinct from the training state, which includes model parameters, optimizer moments, and the current rollout law.

\subsection{Gaussian augmentation and the noise-level ELBO view}
\label{app:augmentation}
For a normalized latent-variable model $p_\theta(x,z\mid\mathsf c)$ and a normalized variational law $Q(z\mid x,\mathsf c)$ with appropriate support and finite expectations, Jensen's inequality gives $\mathcal B_\theta=\E_Q[\log p_\theta(x,z\mid\mathsf c)-\log Q(z\mid x,\mathsf c)]\le\log p_\theta(x\mid\mathsf c)$; if $\mathcal L_\theta=-\mathcal B_\theta$ is called the negative-ELBO loss, it is $-\mathcal L_\theta$ that is the lower bound. Shared parameter-independent terms cancel in the old/current difference; arbitrary loss reweighting or rescaling need not preserve this bound, and a difference of two ELBOs does not bound the log-likelihood ratio.

The purpose of this connection is to explain what distributions are emphasized by a weighting profile. For the linear path in Equation~\eqref{eq:path}, define $Y=x_t/t$ at $0<t<1$. Its marginal density for endpoint law $p$ is
\begin{equation}
 Y=X+\lambda(t)\epsilon,\qquad
 p_{\lambda(t)}=p*\varphi_{\lambda(t)},\qquad
 \lambda(t)=\frac{1-t}{t}.
\end{equation}
Thus noise-level losses examine a family of Gaussian-smoothed endpoint distributions. Changing allocation changes which members of that family receive emphasis.

\paragraph{Coordinate conversion and monotonicity.}
Use $\gamma=\log\mathrm{SNR}=2\log[t/(1-t)]$, distinct from the noise standard deviation $\lambda$. Thus $d\gamma/dt=2/[t(1-t)]>0$. On $0<t_-<t_+<1$, write $\gamma_\pm=\gamma(t_\pm)$. The loss in Equation~\eqref{eq:weight-sampling} can be written in the diffusion ELBO coordinate as
\begin{equation}
 \mathcal L_{\boldsymbol w}(x)
 =\frac12\int_{\gamma_-}^{\gamma_+}W(\gamma)\E_\epsilon\|\widehat\epsilon-\epsilon\|^2\,d\gamma,
 \qquad W(\gamma(t))=\frac{q_0(t)\boldsymbol w(t)(1-t)}{d\,t}.
 \label{eq:augmentation-coordinate}
\end{equation}
This follows directly from $\|\widehat\epsilon-\epsilon\|^2=d\,t^2\ell_v$ and the Jacobian. Here $W$ is only the converted coefficient, not an additional weighting choice. The condition of \citet[Section 4]{kingma2023elbo} is that $W$ be nonincreasing in log-SNR, hence nonincreasing in our noise-to-data time $t$ (their forward diffusion time runs in the opposite direction). Monotonicity of $\boldsymbol w(t)$ alone is insufficient.

\paragraph{Endpoint terms and a normalized augmentation law.}
For the corresponding variational diffusion model, let $K_\theta(\gamma;x)$ be the KL divergence between the forward path conditional on $x$ and the reverse-model path, restricted from noise level $\gamma$ to the noisy endpoint $\gamma_-$. Its derivative is $\partial_\gamma K_\theta=\tfrac12\E_\epsilon\|\widehat\epsilon-\epsilon\|^2$ \citep[Equation 9]{kingma2023elbo}. For absolutely continuous nonnegative $W$, integration by parts gives
\begin{align}
 \mathcal L_{\boldsymbol w}
 &=W(\gamma_+)K_\theta(\gamma_+)-W(\gamma_-)K_\theta(\gamma_-)
   +\int_{\gamma_-}^{\gamma_+}[-W'(\gamma)]K_\theta(\gamma)\,d\gamma,
 \label{eq:augmentation-parts}\\
 \pi(d\gamma)&=\frac{W(\gamma_+)\delta_{\gamma_+}(d\gamma)+[-W'(\gamma)]\,d\gamma}{W(\gamma_-)},\qquad
 \frac{\mathcal L_{\boldsymbol w}}{W(\gamma_-)}+K_\theta(\gamma_-)=\E_\pi K_\theta(\gamma).
 \label{eq:augmentation-mixture}
\end{align}
If $W'\le0$ and $0<W(\gamma_-)<\infty$, $\pi$ is a probability measure; its clean-end atom retains the boundary term even when $W(\gamma_+)>0$. The noisy-end KL is parameter-independent for a fixed terminal prior, not necessarily zero. Each $K_\theta$ is an expected negative ELBO on corrupted data plus a parameter-independent entropy term. Thus, after the displayed normalization and constants, minimizing the loss maximizes an average noise-conditioned augmented-data ELBO. In the rescaled coordinate $Y=X+\lambda\epsilon$, the corresponding expected log density is $\E_{\gamma\sim\pi}\E_{Y\mid x,\gamma}\log p_{\theta,\gamma}(Y)$, not the log density of a mixture over $\gamma$.

Full-endpoint limits require finite limiting mass, boundary terms, and expectations; they are not automatic for weights singular at $t=0$ or $1$. Monotone piecewise-smooth profiles use the same formula with the positive Stieltjes measure $-dW$, including jump masses. General empirical profiles and calibration remain valid prediction-loss surrogates without necessarily satisfying these augmentation-bound conditions.

In RL, the same Gaussian corruption can be applied to both the reference endpoint law and the reward-tilted law. Their log-density ratio is precisely the field in Equation~\eqref{eq:field}; its gradient gives the required velocity correction through Equation~\eqref{eq:velocityfield}. This yields the interpretation used in the main text: weighting allocates learning across differently smoothed views of reward-induced policy change. The KL-tilted endpoint target remains fixed for fixed reward, reference, and temperature, while the surrogate's emphasis across scales changes. The subsequent policy-ratio and clipping operations enter through the specific RL objective in Table~\ref{tab:placement}.

\subsection{Sampling and schedule equivalence}
\label{app:measure-equivalence}
Let $w\ge0$, $0<Z=\E_{q_0}w<\infty$, and $f$ be integrable under the weighted measure. Then
\begin{equation}
 \E_{q_0}[w(t)f(t)]
 =\int q_0(t)w(t)f(t)\,dt
 =Z\E_{\widetilde q}[f(t)],\qquad \widetilde q=q_0w/Z.
\end{equation}
For a discrete mean-one profile on $B$ equal bins, $Z=1$ and $\widetilde q_b=w_b/B$. This is the allocation identity used for image training. For a monotone time reparameterization $t=\phi(\tau)$, the same integral has coefficient $q_0(\phi(\tau))w(\phi(\tau))|\phi'(\tau)|$ and integrand $f(\phi(\tau))$. This is a change of integration coordinate with the Jacobian retained.

The associated Monte Carlo estimators need not have the same variance. For $M$ independent time draws and scalar $f$,
\begin{align}
 \operatorname{Var}\!\left[\frac1M\sum_jw(t_j)f(t_j)\right]
 &=\frac1M\left(\E_{q_0}[w^2f^2]-(\E_{q_0}[wf])^2\right),\\
 \operatorname{Var}\!\left[\frac ZM\sum_j f(\widetilde t_j)\right]
 &=\frac1M\left(Z\E_{q_0}[wf^2]-(\E_{q_0}[wf])^2\right).
\end{align}
Applying a nonlinear exponential or clipping function to these estimators changes their expected objectives and gradient distributions in an implementation-dependent way. A change in the physical probability path, SDE, or scheduler also changes the integrand and is a separate operation.

On a continuous unstabilized support $s\in(0,1)$, $s^{2(1-\alpha)}$ is integrable at zero for $\alpha<1.5$. Discrete support, a truncated interval, or a stabilizer defines other valid implementations, including the larger exponents evaluated here.

\subsection{Shared minimizers and finite shared-parameter updates}
\label{app:local-gram}
Suppose fixed population losses $L_t$ admit a common minimizer $\theta^*$, with $L_t(\theta)-L_t(\theta^*)\ge0$ almost everywhere. If $w(t)>0$ almost everywhere and the integrals exist, then
\begin{equation}
 \int q_0(t)w(t)\bigl[L_t(\theta)-L_t(\theta^*)\bigr]dt\ge0.
\end{equation}
Equality holds only when the integrand's nonnegative loss gap is zero almost everywhere under the weighted measure. Thus all positive weights preserve the jointly realizable population minimizers. Finite model capacity, finite optimization, and moving on-policy losses determine the practical deviations from this idealization.

For additive bin losses with locally Lipschitz gradients and a locally frozen positive-definite preconditioner $P$, the update in Equation~\eqref{eq:amplification} gives
\begin{equation}
 L_c(\theta+\delta\theta)-L_c(\theta)
 =-\eta\sum_b\omega_b g_c^\top P g_b+O(\|\delta\theta\|^2).
 \label{eq:gram}
\end{equation}
Positive inner products support simultaneous local loss reduction; their effect on reward also depends on the reward gradient. This is a local additive model, not a reconstruction of the realized Adam update. At $P=I$, writing an output descent signal as $z_b$ gives $g_b=\E[J_\theta^\top z_b]$, with $J_\theta$ the model-output Jacobian. Consequently, output-space coherence and parameter-space compatibility differ through the Jacobian and through cross-bin expectations. In native FPO, differentiating the aggregate exponential and clipping gate gives weight-dependent factors shared across bin pieces. Gradient geometry must therefore be evaluated at the specified weight profile and policy state.

\clearpage
\section{Reward Definitions, Shape--Scale Findings, and Sampling}
\label{app:shape-scale}
\subsection{Reward definitions}
\label{app:rewards}
The rewards keep their native output scales. Let $x^{01}=\operatorname{clip}((x+1)/2,0,1)$, $B(x^{01})$ be Gaussian blur with $\sigma=1$, and $P$ be the patch at rows $20{:}28$, columns $10{:}22$. Define a soft red map
\begin{equation}
 r(z)=\operatorname{sigmoid}\!\left(\frac{z_R-\max(z_G,z_B)-.10}{.08}\right).
\end{equation}
\paragraph{Automobile conjunctions.}
Let $p_{\rm auto}(x)$ be class-1 confidence from the frozen CIFAR classifier. Auto-A35 is
\begin{equation}
 R_{\rm Auto}(x)=p_{\rm auto}(x)\operatorname{sigmoid}\!\left(6\bigl[\langle r(B(x^{01}))\rangle_P-\tfrac12\langle r(B(x^{01}))\rangle_{P^c}\bigr]\right).
\end{equation}
For Mixed, form residual $e=x^{01}-B(x^{01})$ and its channel-mean finite-difference magnitude $d=\operatorname{mean}_{\rm ch}\sqrt{(\nabla_1 e)^2+(\nabla_2 e)^2+10^{-12}}$, using forward differences and zero padding at the last row/column. Then
\begin{equation}
 R_{\rm Mixed}(x)=p_{\rm auto}(2B(x^{01})-1)\operatorname{sigmoid}\!\left(12[\langle d\,r(x^{01})\rangle_P-.03]\right).
\end{equation}
These multiplicative rewards are used for training and evaluation. Separate differentiable proxies use attribute coefficients $.35$ and $.20$; the measured black-box rewards remain the products above.

\paragraph{Edge and semantic rewards.}
Edge averages $[1+((f_j-\mu_j)/a_j)^2]^{-1}$ over eight Sobel/Laplacian mean/SD features at blur scales zero and one. Its fixed calibration uses the first 4,096 CIFAR training images, width multiplier $2.5$, relative floor $.05$, and absolute floor $.001$. CLIP reports automobile probability from $\operatorname{softmax}(20\,\mathrm{cosine})$ over ten class prototypes formed from five text templates per class. SimCLR uses ten normalized class prototypes from a frozen encoder. With target cosine $s_+$, other cosines $s_j$, and $\tau=.08$, it reports
\begin{equation}
 R_{\rm SimCLR}=\operatorname{sigmoid}\!\left(\frac{s_+-\tau\log(\tfrac19\sum_j e^{s_j/\tau})}{\tau}\right).
\end{equation}
\paragraph{Digit7.}
The semantic score is $C=\operatorname{sigmoid}(12(s_+-\overline s_-))$, using a centered handwritten-seven prompt and five negative prompts (blank white, solid black, digits three/eight, and abstract texture). The structural score uses grayscale ink $I=1-\operatorname{mean}_{\rm ch}x^{01}$, occupancy $m=\langle I\rangle$, its ink-weighted centroid $(c_x,c_y)$ on $[-1,1]^2$, and border density ratio $b$ on the outer three-pixel border:
\begin{align}
 T&=e^{-((m-.18)/.12)^2}
 e^{-(c_x^2+c_y^2)/.55^2}e^{-b/.10}
 \operatorname{clip}\!\left(\frac{\operatorname{std}_{\rm pop}(1-I)}{.5},0,1\right),\\
 R_{\rm Digit7}&=.75C+.25T.
\end{align}
Centroid and border-density denominators have stabilizer $10^{-6}$. The semantic backend is frozen OpenCLIP; this reward is not a digit-classifier accuracy.

\subsection{Joint shape--scale grid}

\label{sec:scale}
\begin{figure}[t]
\centering
\includegraphics[width=\linewidth]{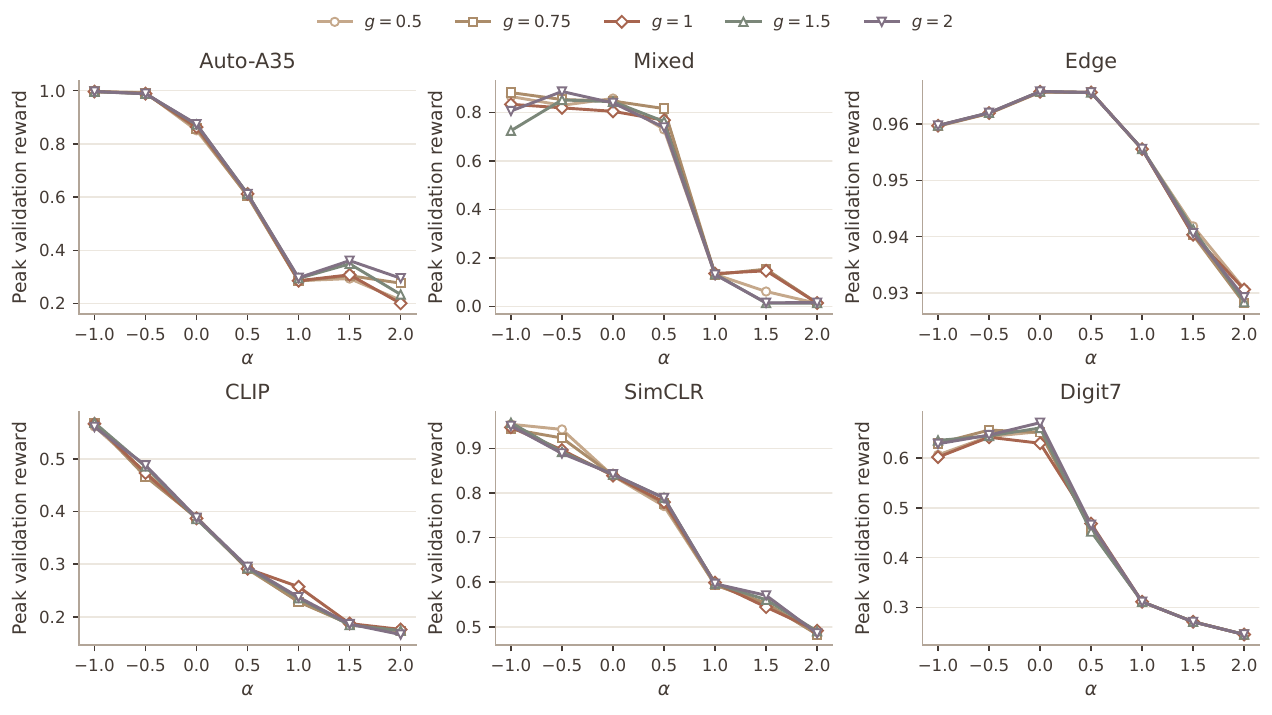}
\caption{Six CIFAR rewards on a calibrated $7$-exponent by $5$-scale grid ($n=3$ per cell): mean per-run validation peak versus $\alpha$, with one colored curve per whole-loss scale $g$. Auto-A35, Edge, CLIP, and SimCLR keep their preferred shape across these scales; Mixed's crossing curves and Digit7's central switch show where scale can change the winner. Reward axes use native units; the ranks are in Appendix~\ref{app:cifar-results}.}
\label{fig:shape-scale}
\end{figure}

Preferred shape is largely stable across the tested calibrated scales, supporting a practical order: choose shape first, then tune scale. The effect of scale on preferred shape is relatively small for most tested rewards, even where attainable reward changes appreciably. We evaluate six rewards, seven profiles, five scales, and three seeds per cell. The common-bank calibration in Equation~\eqref{eq:calibration} controls whole-loss RMS; the scale multiplier varies over $g\in\{.5,.75,1,1.5,2\}$. Figure~\ref{fig:shape-scale} shows the entire mean surface as overlaid shape curves, with one line per scale.

For Auto-A35, Edge, CLIP, and SimCLR, the curves preserve the preferred shape and their broad ordering across the tested scale range. Their separation along the exponent axis remains visible at each scale. Scale nevertheless changes attainable reward within a profile, including the transition region between strong and weak semantic-reward profiles. Digit7 retains a broadly similar landscape but switches its leading shape within the central region.

\begin{figure}[t]
\centering
\includegraphics[width=\linewidth]{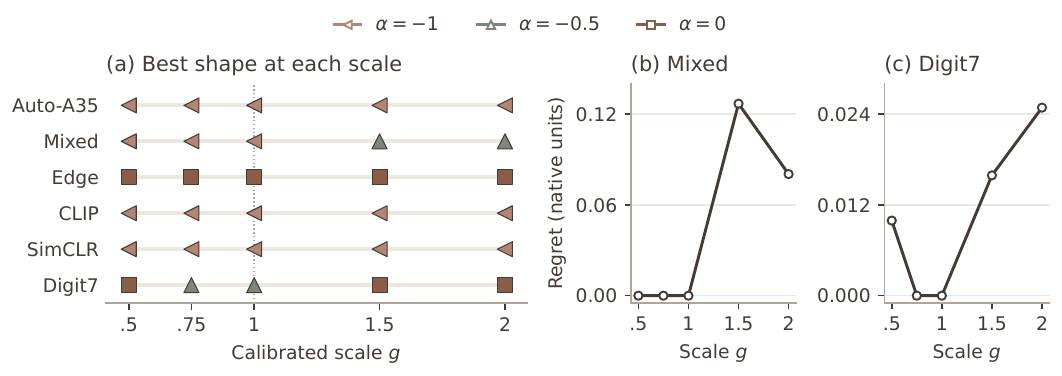}
\caption{Six CIFAR rewards on the seven-profile, three-seed grid: best mean-peak exponent at each calibrated scale $g$ (left) and mean-peak regret from retaining the $g=1$ winner (right). Four rewards keep the same winning shape, whereas Mixed and Digit7 incur scale-dependent regret; shape-first tuning may need a recheck for those rewards. The right-panel reward axes are independent; exact selections appear in Table~\ref{tab:scale}.}
\label{fig:scale-transfer}
\end{figure}

Mixed shows the clearest interaction: changing scale reshapes the landscape and changes which allocation is favored. Its crossing curves make clear why a fixed-scale winner need not remain the preferred choice after retuning loss magnitude. The two questions are therefore distinct: scale can change reward, yet a shape selected at the reference scale can remain useful after that change. To measure the latter on this grid, define
\begin{equation}
 \operatorname{regret}_r(g)=\max_\alpha\overline P_r(\alpha,g)-\overline P_r(\alpha_r^*(1),g),\qquad
 \alpha_r^*(1)\in\arg\max_\alpha\overline P_r(\alpha,1).
\end{equation}
Appendix Figure~\ref{fig:scale-transfer} shows the scale-specific winners and the regret curves for the two rewards whose winners change. These group-mean comparisons support shape-first tuning within this calibrated range, with a local shape recheck when scale changes the preferred region, as for Mixed and Digit7. Appendix~\ref{app:cifar-results} gives a complementary rank map.

The rank map and selected-shape summary are in Appendix~\ref{app:cifar-results}.

\subsection{Sampling versus explicit weighting}
\label{app:sampling}
The fixed-integrand identity in Appendix~\ref{app:measure-equivalence} motivates implementing a profile through timestep sampling, which removes the fluctuating per-draw multiplier $w(t)$. The variance expressions there have no universal ordering: the outcome also depends on the loss at each time. We distinguish this motivation from the recorded performance comparison below.

An earlier matched study compared constant-$Z$ profile sampling with uniform-time explicit weighting for nine reward-specific profiles and three seeds per reward. Pairs shared reward, seed, profile, validation bank, optimizer, and stopping rule; realized stopping times could differ. This study predates the frozen-bank RMS calibration. Table~\ref{tab:sampling} summarizes its observed validation-peak differences. Edge improves consistently at the stated tolerance; Auto-A35 has a positive mean with substantial profile/seed heterogeneity. These are performance outcomes, not measurements of estimator variance. Broader historical alpha/beta-family comparisons also have mixed signs, so sampling is not uniformly superior.

\begin{table}[ht]
\centering\small
\caption{Historical matched estimator comparison. Entries are mean paired peak differences $\pm$ sample SD across profile--seed pairs; W/T/L uses a $.003$ tie tolerance. Each reward has nine profiles and three seeds, not 27 independent training seeds.}
\label{tab:sampling}
\begin{tabular}{@{}lrrr@{}}
\toprule Reward & Profile--seed pairs & Peak difference $q-$explicit & W/T/L\\
\midrule
Auto-A35 & 27 & $+0.05478\pm0.15620$ & 13/5/9 \\
Edge & 27 & $+0.00796\pm0.00538$ & 23/4/0 \\
\bottomrule\end{tabular}

\end{table}

\clearpage
\section{Local Response and Cross-Scale Gradient Analysis}
\label{app:gradients}
\subsection{Local reward response}
\label{app:mechanisms}
Initial response measurements expose reward-dependent opportunities, while their association with early learning changes across observation windows. Figure~\ref{fig:response-main} illustrates the five matched rewards; full statistics are in Appendix~\ref{app:response-results}.

\begin{figure}[htbp]
\centering
\includegraphics[width=\linewidth]{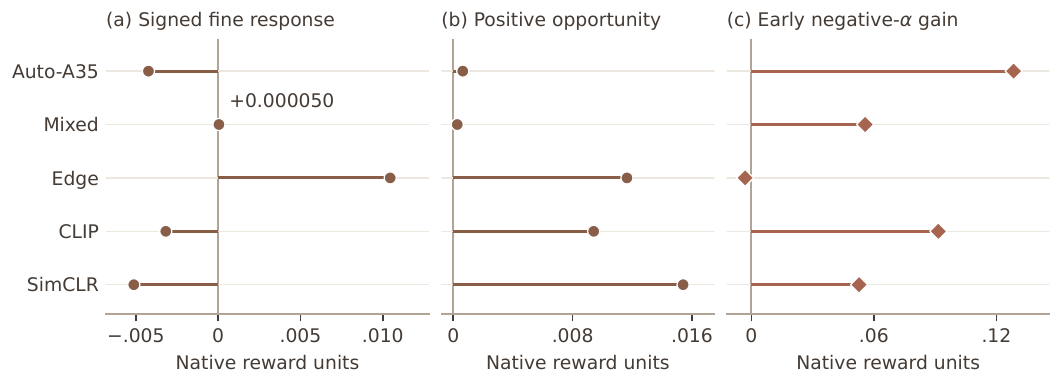}
\caption{Five matched CIFAR rewards: initial signed local response and positive opportunity under $4/255$ image perturbations (256 pretrained images), compared with the early validation-reward advantage of high-noise $\alpha\in\{-2,-1\}$ over $x_0$ at updates 20--200 ($n=3$). Local response varies by reward and does not alone determine the benefit of high-noise emphasis; Mixed's positive signed response is marked. Native axes are independent; Table~\ref{tab:response-full} defines the response summaries and reports their values.}
\label{fig:response-main}
\end{figure}

For each generated endpoint $x$, define $d_\pm=R(x\pm\delta)-R(x)$ using the recorded clipping and brightness transformations. The measurements distinguish negative-response magnitude $M=\E[(\max(-d_+,0)+\max(-d_-,0))/2]$, signed even response $C=\E[(d_++d_-)/2]$, positive opportunity $O=\E\max(d_+,d_-,0)$, and double-decrease frequency $\Pr(d_+<0,d_-<0)$. The analysis averages brightness conditions within an anchor before averaging anchors.

Coarse, middle, and fine bands refer to spatial Fourier frequencies, with boundaries $0,.08,.22$, and Nyquist. The amplitudes $2/255$, $4/255$, and $8/255$ are a separate axis. All initial responses use 256 anchors from the same pretrained model. Mixed's displayed fine-scale response is net positive (Table~\ref{tab:response-full}). The early-reward comparison uses the equal-weight mean of observed rewards at updates $20,40,\ldots,200$ for $\alpha=-2,-1$, relative to the same-cohort $x_0$ control. This is an auxiliary learning-dynamics summary, separate from the primary peak metric.

The cross-reward association is descriptive and depends on the observation window (Table~\ref{tab:response-windows}). The initial responses are fixed-model scorer measurements.

Explicit Automobile has one paired seed against $x_0$; the other displayed training comparisons use three. Digit7's separate full-800 $q$ cohort changes from a negative early advantage to a positive later-window advantage. It uses a different validation bank from the joint grid.

The explicit-cohort association is not statistically significant and changes with the outcome summary or omission of a reward; Automobile and Auto-A35 share a scorer component. The evidence therefore supports a window-dependent relation.

\subsection{Interpreting the main heatmaps}
\label{app:approved-heatmaps}
Figures~\ref{fig:geometry} and~\ref{fig:training-evolution} show recorded cross-batch parameter-gradient cosines at individual checkpoints. Each matrix axis orders generative noise from clean to noisy; optimizer updates or control-training iterations appear only in panel titles. CIFAR uses twelve bins, ordered b11 to b0, and four diagnostic batches. Go2 and G1 use ten bins, ordered by increasing noise coefficient, and eight batches. The cross diagonal retains measured same-bin reproducibility. All panels use the signed range $[-1,1]$, the common warm palette, and $z(c)=\operatorname{asinh}(c/.1)/\operatorname{asinh}(10)$ with ticks in raw cosine units. The diagnostic objectives are defined in Appendices~\ref{app:cifar-gradients} and~\ref{app:robotics-new-geometry}.

\paragraph{Profile comparison.}
Table~\ref{tab:heatmap-profile-sources} in Appendix~\ref{app:gradient-results} gives the full-trajectory peak used to order the two profiles within each column of Figure~\ref{fig:geometry}. ``Higher'' and ``lower'' compare only those two displayed profiles. CIFAR peaks are fixed-validation rewards; robotics peaks are the equal-weight mean of each seed's highest recorded evaluation return. The heatmaps use an earlier single checkpoint, not the peak checkpoint or a phase average. SimCLR and CLIP use the same seed in both rows; Digit7 uses different seeds and is a descriptive comparison of two resulting policies. All CIFAR panels belong to the same long-horizon training family; the independent early-run measurements elsewhere in this appendix are not spliced into it.

At Go2 iteration 300, the coordination contrast is present in both seeds. At G1's late checkpoint, most off-diagonal entries in each seed have absolute cosine below $.1$, so the weak average is not merely cancellation between opposite-signed seed matrices. The full panel identities and per-seed observations are in Appendix~\ref{app:gradient-results}.

\subsection{CIFAR within- and cross-batch measurements}
\label{app:cifar-gradients}
The main cohort contains 80 states: four rewards (Auto-A35, CLIP, SimCLR, Digit7), four training exponents ($-2,-1,0,1$), and five checkpoints. Each state is one selected training seed, measured with four independent on-policy batches, 12 time bins, and 64 draws per bin. Checkpoints at updates 50/100 (E50/E100) and 200/400/800 (H200/H400/H800) come from separate training prefixes; only the 200/400/800 checkpoints are connected as one trajectory in the figures. Alpha comparisons describe the resulting policy states, with their own on-policy distributions and selected seeds.

The H-prefix policies were trained with profile sampling and constant $Z$; the diagnostic below instead samples equally within each bin.

Let $G_{i,b}$ be the saved parameter gradient of the equal-bin, unweighted diagnostic loss for batch $i$ and bin $b$. The diagnostic removes the training profile from bin sampling and explicit weights. Its diagnostic objective is
\begin{equation}
 r=\exp\!\left(\operatorname{clip}\!\left((L_{\rm old}-L_{\rm new})/d,-20,20\right)\right),\qquad
 \ell_b=-\E\min\{rA,\operatorname{clip}(r,.92,1.08)A\},
\end{equation}
where $d=3072$ and the expectation is over examples in the batch. It then differentiates $\ell_b$ with respect to the model parameters. The recorded vectors are therefore gradients of this diagnostic surrogate, not a forced $r=1$ objective or realized optimizer steps. With $B=4$, define the plotted matrices by
\begin{align}
 C^{\rm within}_{bc}&=\frac1B\sum_i\frac{G_{i,b}^{\top}G_{i,c}}{\|G_{i,b}\|\|G_{i,c}\|},\\
 C^{\rm cross}_{bc}&=\frac1{B(B-1)}\sum_{i\ne j}\frac{G_{i,b}^{\top}G_{j,c}}{\|G_{i,b}\|\|G_{j,c}\|}.
 \label{eq:pairwise-cosines}
\end{align}
The cross diagonal measures same-bin reproducibility; it is not replaced with one. The 12 ordered cross-batch pairs are dependent averages of four batches, not 12 independent replicates. The axes place clean b11 at the upper left and noisy b0 at the lower right. All CIFAR heatmaps use $\operatorname{asinh}(c/.1)/\operatorname{asinh}(10)$ for color display only, with raw-cosine colorbar ticks. The supplementary gradient/reward curves use raw training reward. The CIFAR rows of main Figure~\ref{fig:geometry} are instead ordered by full-trajectory fixed-validation peaks, with the two metrics kept separate.

\begin{figure}[t]
\centering
\includegraphics[width=\linewidth]{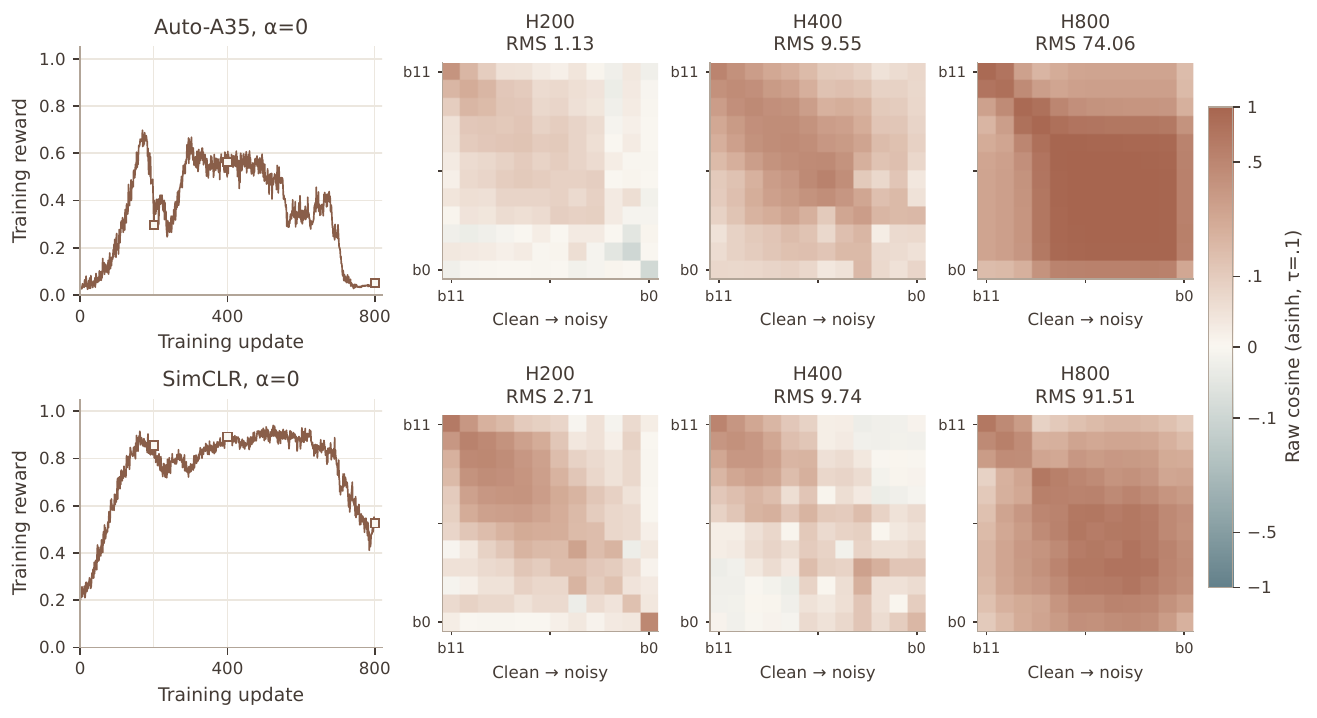}
\caption{Auto-A35 and SimCLR at fixed $\alpha=0$: raw training reward trajectories (left) and cross-batch gradient cosine across noise bins at updates 200, 400, and 800 (right). Both deteriorating branches develop stronger cross-bin agreement and larger raw-gradient RMS, so stronger coordination alone does not imply better reward. Colors and axes match Figure~\ref{fig:geometry}; Appendix~\ref{app:gradient-counterexamples} details measurements and the contrasting CLIP branch.}
\label{fig:reinforcement}
\end{figure}

For unit gradients $U_b=G_b/\|G_b\|$ and independent batches, the population cross entry is $\E[U_b]^\top\E[U_c]$. Its magnitude depends on both mean-direction alignment and the reproducibility of each bin. Consequently, small cross entries alone do not establish orthogonal population mean directions. The finite four-batch estimates, including cross diagonals, can be negative. The raw-gradient RMS in Figure~\ref{fig:reinforcement} is $\sqrt{(12B)^{-1}\sum_{i,b}\|G_{i,b}\|^2}$ and does not include a reconstruction of the optimizer's preconditioning or clipping.

\paragraph{Direct shared-fluctuation decomposition.}
Unlike Equation~\eqref{eq:pairwise-cosines}, the following decomposition operates on raw inner products before normalization:
\begin{align}
 W_{bc}&=B^{-1}\sum_i G_{i,b}^{\top}G_{i,c},\qquad
 K_{bc}=[B(B-1)]^{-1}\sum_{i\ne j}G_{i,b}^{\top}G_{j,c},\\
 N_{bc}&=W_{bc}-K_{bc}
 =\frac1{B-1}\sum_i(G_{i,b}-\overline G_b)^\top(G_{i,c}-\overline G_c),\\
 C^{\rm fluct}_{bc}&=\frac{N_{bc}}{\sqrt{N_{bb}N_{cc}}}.
 \label{eq:fluctuation}
\end{align}
Here $N$ is the directly centered batch-gradient covariance Gram, and $K$ is an unbiased cross-batch estimate of the stable-mean Gram under independent sampling. The ratio $S=\operatorname{tr}(K)/\operatorname{tr}(N)$ compares the sums of estimated stable-mean energy and fluctuation energy across bins. These traces do not measure off-diagonal correlation contributions. In particular, $S<1$ does not imply that every cross-bin correlation is fluctuation-dominated. The finite estimate $K$ need not be positive semidefinite: 124 of 960 diagonal estimates and seven state traces are negative. We preserve these as unresolved stable signal, not negative population squared norms. All 960 fluctuation diagonals are positive.

\begin{figure}[t]
\centering
\includegraphics[width=\linewidth]{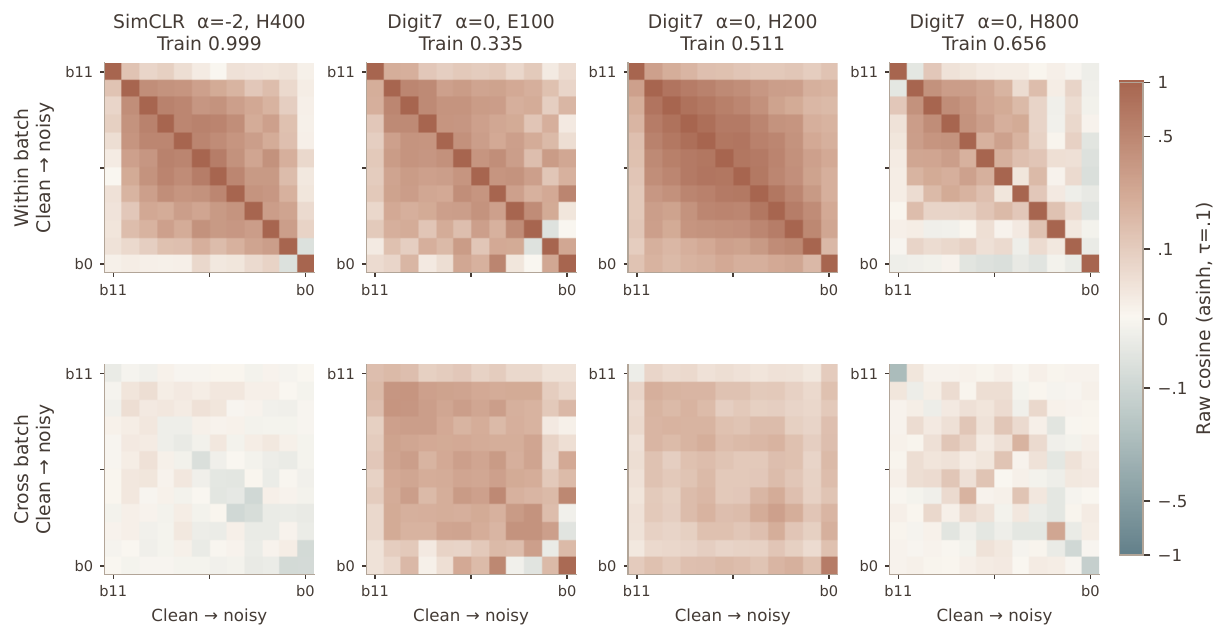}
\caption{CIFAR checkpoint diagnostics: within-batch (top) versus independent-cross-batch (bottom) parameter-gradient cosines between clean-to-noisy timestep bins. Apparent within-batch cooperation can weaken across batches, as in successful SimCLR, so shared-sample structure need not be reproducible directional signal; raw-gradient fluctuations exceed stable-mean energy in 75/80 measured states. E/H identify separate training prefixes; color ticks show raw cosines, with sampling and Gram definitions in Appendix~\ref{app:cifar-gradients}.}
\label{fig:within-cross}
\end{figure}

\paragraph{Shared samples induce structured cross-scale fluctuations.}
The CIFAR gradients distinguish between agreement within the same batch and agreement across independently drawn batches. Strong within-batch structure can become weak in the cross-batch matrix, as in the successful SimCLR branch in Figure~\ref{fig:within-cross}. The same generated examples and advantages can induce a common response at several noise levels without that response being stable across samples.

The raw-Gram decomposition attributes part of this difference to structured shared fluctuations: batch-fluctuation energy exceeds estimated stable-mean energy in 75 of 80 states, and mean off-diagonal fluctuation correlation is positive in all 80. These correlations are spatially structured, with stronger middle/low-noise neighborhoods than high-to-low coupling. A paired control that shares endpoints and advantages also increases agreement. Such fluctuations can reflect useful sample-specific corrections as well as sampling variability. The full decomposition is in Table~\ref{tab:shared-fluctuation}; robotics exhibits a related within/cross contrast in Appendix~\ref{app:robotics-new-geometry}.

The modest average fluctuation correlation and negative entries distinguish this pattern from one uniformly strong common direction. For the block summaries in Table~\ref{tab:shared-fluctuation}, high, middle, and low noise are bins 0--3, 4--7, and 8--11; within-block means omit diagonals.

A fifth paired control reuses batch 0's generated endpoints, initial noise, raw rewards, and group advantages while redrawing timestep and corruption noise. Its mean same-bin cosine exceeds the independent-batch counterpart in 75 of 80 states, with median difference $.1860$. It supports an effect of sharing this collection of variables, without separating endpoint and advantage contributions. The control does not enter $W,K,N$.

Digit7 has a larger median $S$ than the other three rewards (Table~\ref{tab:shared-fluctuation-rewards}). Within its $\alpha=0$ branch measured at updates 200, 400, and 800, $S$ falls from $.2018$ to $.0728$ and then $-.1437$, while training reward rises. The last estimate denotes unresolved stable signal with four batches. This is consistent with changing coordination needs; it does not make $S$ a universal measure of learning quality. For example, high-performing SimCLR $-2$ retains weak stable-mean estimates, while poor-performing CLIP $+1$ can have strong ones.

\paragraph{Ratio markers.}
A dagger marks the two states with any recorded batch outside $|r-1|\le .08$: CLIP $\alpha=0$ at H800 ($\max|r-1|=.130474$, two of five recorded batches), and Digit7 $\alpha=1$ at H800 ($.577620$, all five). The five records include the paired control. The actual clipped-gradient fraction is unavailable because it requires joint per-example ratios and advantages. Table~\ref{tab:shared-fluctuation} compares all states with the subset omitting these two.

\paragraph{Common reinforcement can amplify both progress and deterioration.}
\label{sec:reinforcement}
The benefit of coordination depends on what is being reinforced. Along the measured Auto-A35 and SimCLR trajectories with $\alpha=0$, late reward deterioration accompanies denser cross-bin agreement and substantial growth in raw gradient norms (Figure~\ref{fig:reinforcement}). Stronger common directions therefore need not mean more useful learning.

A local model explains why alignment can amplify learning in either a useful or an unstable direction. For locally fixed additive bin losses, let $g_b=-\nabla L_b$, with nonnegative allocation coefficients $\omega_b$. With a locally frozen positive-definite preconditioner $P$ and $\|v\|_P^2=v^\top Pv$, write
\begin{equation}
 \delta\theta\approx\eta P\sum_b\omega_b g_b,\qquad
 \left\|\sum_b\omega_b g_b\right\|_P^2
 =\sum_b\omega_b^2\|g_b\|_P^2
   +2\sum_{b<c}\omega_b\omega_c g_b^\top P g_c,
 \label{eq:amplification}
\end{equation}
Positive cross terms reinforce the combined direction. Whether that direction improves reward instead depends on $\nabla J^\top\delta\theta$, not on its norm or mutual alignment alone. Appendix~\ref{app:local-gram} gives the corresponding local loss-change relation and its application to the native surrogates.

We hypothesize that persistent common reinforcement can form a feedback loop in which policy change makes subsequent surrogate corrections more concentrated or larger, further amplifying the same drive and increasing the risk of training collapse. The observed Auto-A35 and SimCLR patterns motivate this mechanism. Additional deterioration in CLIP accompanies gradient amplification while mean alignment falls (Appendix~\ref{app:gradient-counterexamples}). The useful distinction is between coordination that supports progress and subsides as corrections are resolved, and sustained common drive that fails to improve behavior. Weighting controls their superposition; maximizing coherence alone is insufficient.

\subsection{Stage dependence and complementary geometries}
\label{app:gradient-counterexamples}
\begin{figure}[htbp]
\centering
\includegraphics[width=.88\linewidth]{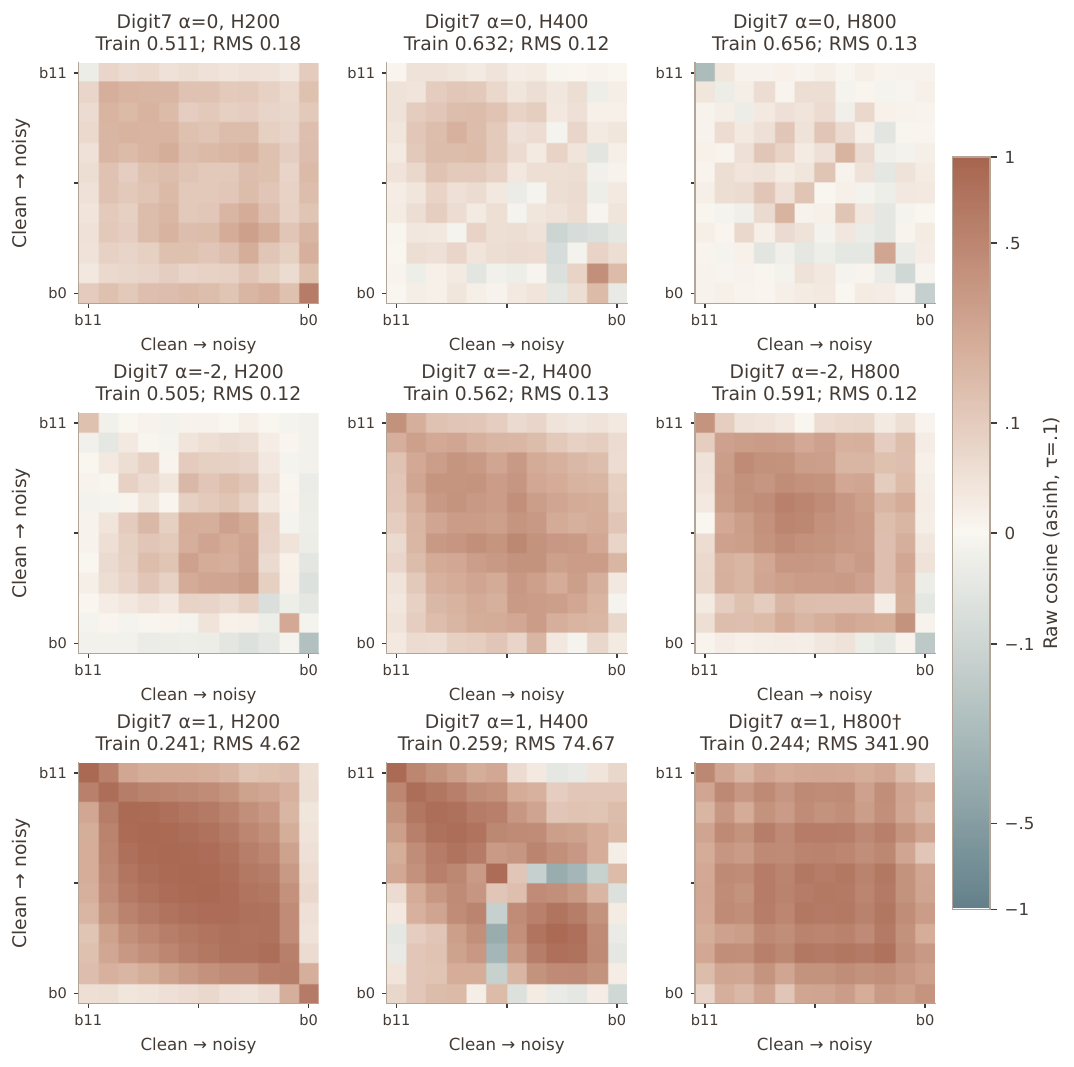}
\caption{Digit7: cross-batch gradient cosine across noise bins at updates 200/400/800 for fixed $\alpha=0,-2,1$; labels give raw reward and gradient RMS. Reward improves as agreement weakens for $0$, while successful $-2$ retains common directions: progress has no single cosine signature. Colors show raw-cosine ticks; $\dagger$ flags a ratio departure defined in Appendix~\ref{app:cifar-gradients}.}
\label{fig:digit-gradient-stages}
\end{figure}

The behavioral gap in Section~\ref{sec:gram} concerns the difference between current and desired generated behavior, rather than raw reward differences across scorers. The coordination account is a hypothesis explaining stage dependence; these measurements do not directly quantify structure formation. Figure~\ref{fig:reinforcement} follows Auto-A35 $\alpha=0$, seed 20260725, and SimCLR $\alpha=0$, seed 20260726. Its sparse gradient checkpoints establish co-occurring deterioration, alignment, and amplification, but not their temporal order or image mode collapse.
The Digit7 $\alpha=0$ branch combines improving reward with weakening cross-batch agreement (Figure~\ref{fig:digit-gradient-stages}). The $-2$ branch retains common directions with sustained or slightly improving reward, showing that success does not require agreement to vanish. The flagged $+1$ endpoint is retained as a diagnostic state rather than evidence about a common unclipped objective.

\begin{figure}[htbp]
\centering
\includegraphics[width=.96\linewidth]{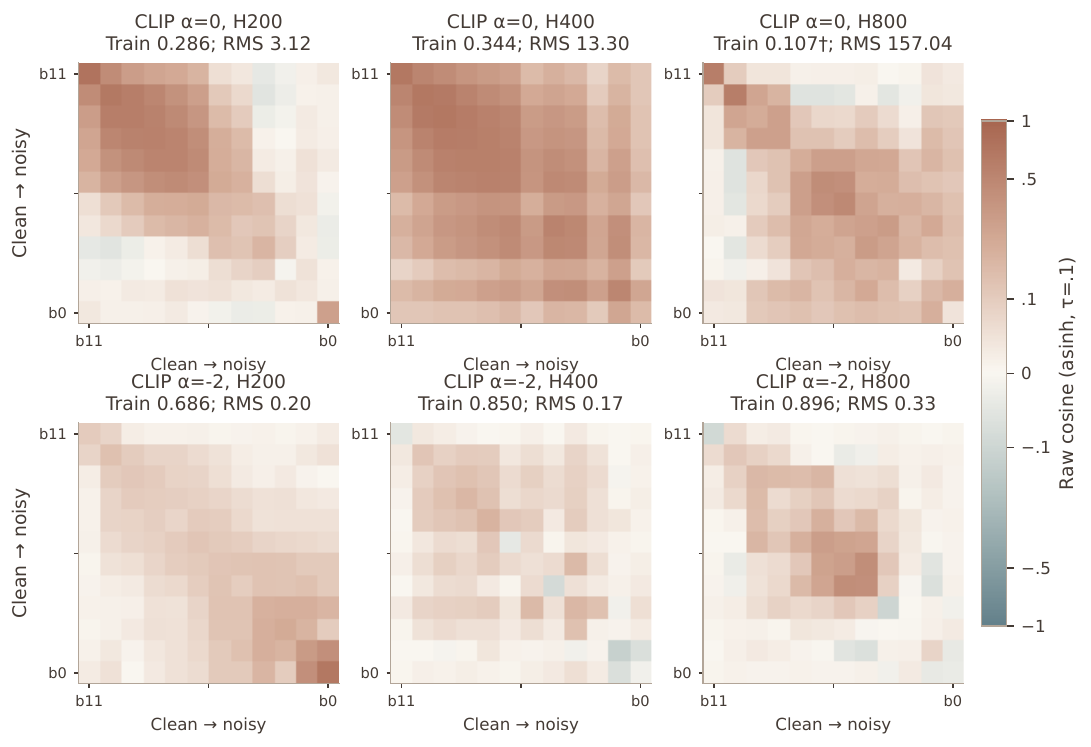}
\caption{CLIP: cross-batch gradient cosine across noise bins at updates 200/400/800 ($\alpha=0$ top, $-2$ bottom). Deteriorating $0$ loses agreement, while better $-2$ retains a middle/low-noise positive block; deterioration has no universal cosine signature. Colors, raw-reward/RMS labels, and the ratio-departure dagger follow Figure~\ref{fig:digit-gradient-stages}.}
\label{fig:clip-gradient-stages}
\end{figure}

CLIP provides two complementary observations (Figure~\ref{fig:clip-gradient-stages}). Its $\alpha=0$ reward deterioration accompanies falling average agreement, in contrast to the Auto-A35 and SimCLR reinforcement examples. At fixed $\alpha=-2$, late agreement weakens overall but retains a localized middle-to-low-noise block. The nearly white late cross matrix of SimCLR $-2$, despite strong within-batch structure, has a different geometry. Changes in these angular patterns do not quantify contributions to realized parameter updates.

\subsection{Robotics: stage-dependent cooperation and conflict}
\label{app:robotics-new-geometry}
The cross-batch measurement cohort contains 783 checkpoints: 141 FPO and 642 FPO++ records. At each checkpoint, eight fresh on-policy batches supply unweighted actor-parameter gradients. Equation~\eqref{eq:pairwise-cosines} is applied with $B=8$: normalize each batch--bin gradient first, then average within-batch or distinct-batch dot products. Zero-norm pairs are omitted, with their valid-pair counts preserved. The cross diagonal is the finite estimate of same-bin reproducibility, not one; four active diagonal estimates are negative. The 56 ordered cross-batch pairs share eight batches and are not independent replicates.

\paragraph{Gradient and sampling protocol.}
FPO freezes the checkpoint and observation statistics, draws each batch from a fresh rollout's first training minibatch (30,720 transitions), and uses 24,576 draws at each native noise coefficient. Recomputing the old loss at the frozen parameters forces $r=1$; the weight table is one. Advantages are checkpoint GAE, standardized within the minibatch. FPO++ uses a complete rollout per batch, equal-bin corruption draws, and the unweighted advantage--MSE actor gradient $-(NM)^{-1}\sum_{(n,m)\in b} A_n\nabla\ell_{nm}$ at $r=1$, without ratio clipping or clamping. Its advantages are standardized over the rollout and then clamped to $\pm100$. Rollout sizes and draw counts follow the task configuration. These protocols share the cosine statistic with CIFAR, but not its batch count, sample size, or diagnostic objective: CIFAR has four batches, 64 draws per bin, and the actual exp-ratio/PPO-min surrogate in Appendix~\ref{app:cifar-gradients}. A common color scale therefore supports reading the same statistic, without asserting matched estimator noise or directly comparable population alignment strength across domains.

\paragraph{Phase and seed aggregation.}
FPO early/middle/late panels use checkpoints at blocks 20/60/122. FPO++ uses iterations 0--40/50--90/100--149 for Cartpole, and 0--400/500--900/1000--1499 for the other tasks. Within each phase, checkpoint matrices are averaged within each seed, then seeds receive equal weight. These phase-averaged panels are supplementary. Main Figures~\ref{fig:geometry} and~\ref{fig:training-evolution} use the exact individual checkpoints specified in their captions, averaging seeds only. Every task, profile, and phase appears in the atlas in Appendix~\ref{app:robotics-atlas}; captions retain reduced-seed panels after failure. The active FPO support is $\sigma=.1,\ldots,.9$ (the dead $\sigma=1$ bin is omitted); FPO++ labels ten bins by midpoints $.05,\ldots,.95$. Since $\sigma$ is noise, ascending row and column order places low noise at the upper left, matching CIFAR's clean-to-noisy direction. All heatmaps use the same warm palette, $\tau=.1$ asinh display, and raw-cosine ticks over $[-1,1]$.

\paragraph{Spatial structure and stage dependence.}
Cartpole spans $.782$--$.908$ across profiles and stages, whereas G1 spans $.045$--$.091$. A small average need not mean all entries are weak: late Walker, Swimmer, and Go2 at native exponent 1 have positive within-block and negative between-block means (Table~\ref{tab:robotics-cross-blocks}). Go2's best tested profile $\alpha=1$ has late off-diagonal mean $.079$, below $\alpha=0$'s $.433$ despite its higher mean seed peak return ($41.57$ versus $39.39$). Spot provides another pattern: its native late branch has mean $.201$, compared with $.081$ for $\alpha=0$. Across the 44 task--profile groups, 31 have lower late than early means. These observations describe changing cooperation and conflict; they are not a monotonic criterion for policy quality.

\paragraph{Within/cross contrast.}
All 783 checkpoints have higher mean off-diagonal within-batch than cross-batch cosine (median gap $.1588$; Table~\ref{tab:robotics-within-cross}), consistent with shared-sample fluctuations. Repeated checkpoints are not independent training replications. Normalized cosines alone cannot recover the raw-gradient energy decomposition.

\clearpage
\section{Weighting Selection and Training-Path Interventions}
\label{app:methods}
\subsection{Output-gradient estimator and static deployment}
\begin{figure}[ht]
\centering
\includegraphics[width=\linewidth]{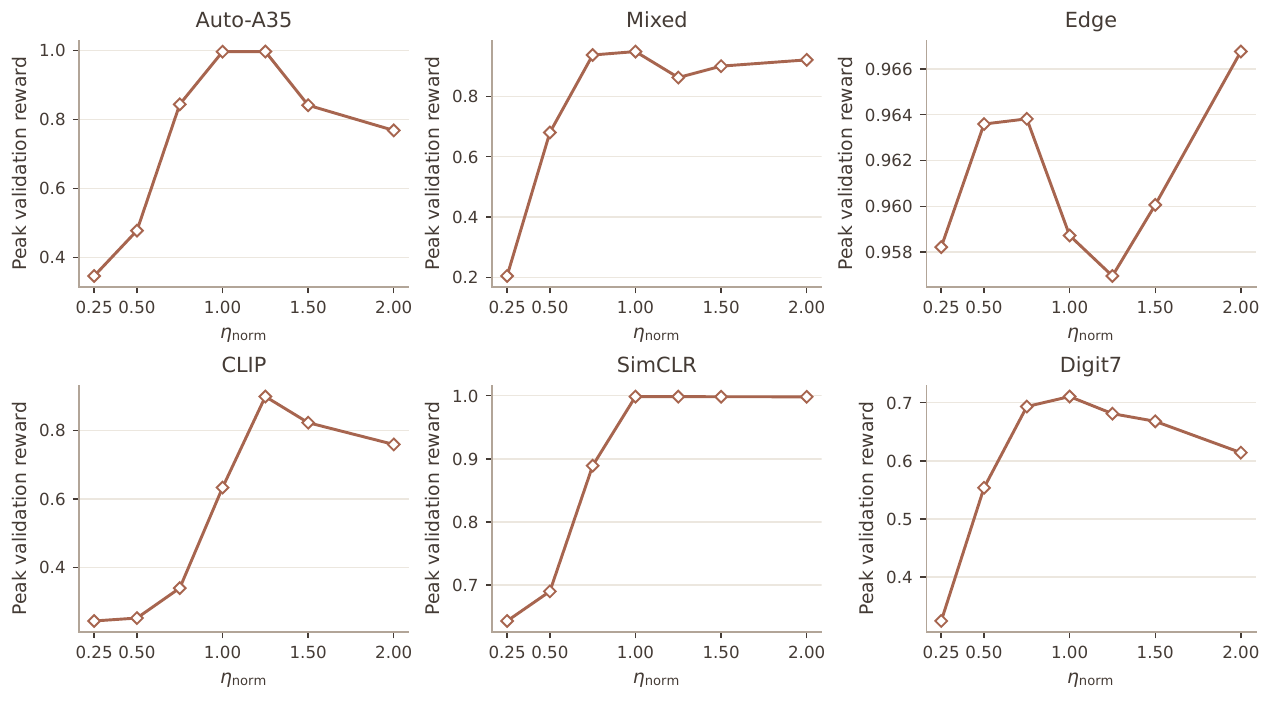}
\caption{Six CIFAR rewards: mean per-run fixed-validation peak versus static-softmax concentration $\eta_{\rm norm}$. Concentration improves some rewards but Digit7 declines above one, so stronger bin selection is not uniformly beneficial. The tested concentration settings are shown on the horizontal axis.}
\label{fig:static-eta}
\end{figure}

Reward-conditioned output coherence provides a useful initial allocation, with a reward-dependent concentration optimum. Figure~\ref{fig:static-eta} shows the concentration response, including the decline beyond Digit7's selected concentration.
For $M$ output-signal samples $z_j$ within a bin, the pairwise estimate of squared mean signal is
\begin{equation}
 U=\frac{\|\sum_jz_j\|^2-\sum_j\|z_j\|^2}{M(M-1)},\qquad
 V=\frac1M\sum_j\|z_j\|^2.
\end{equation}
The deployed coherence uses a nonnegative signal estimate with a stabilized second-moment denominator. Pairwise self-product removal targets draw-level noise; samples sharing an endpoint or group advantage remain statistically related. The finite-sample ratio therefore estimates Equation~\eqref{eq:coherence} with a random denominator and truncation. The static direct rule normalizes the nonnegative frozen coherence profile; the static softmax rule standardizes it across bins with population SD and then applies Equation~\eqref{eq:static}.

Static softmax improves selected reward peaks, but excessive concentration can reverse the gain (Figure~\ref{fig:static-eta}). The fixed-reference comparisons are in Figure~\ref{fig:static}; configuration and baseline definitions are in Appendix~\ref{app:method-settings}.

Digit7's selected static mean is $.710581$, compared with $.641535$ for its best fixed-sweep reference and $.674514$ for its matched method-cohort $x_0$ control (Appendix~\ref{app:static-results}). The fixed-sweep comparison uses separate validation banks; the method-cohort differences are paired. Its static and online means are close, with greater online seed dispersion; static-direct was not measured for Digit7.

\subsection{Online refresh rules and main outcomes}
\label{app:online-analysis}
Online coherence weighting has mixed outcomes across rewards; full profile replacement performs poorly even after concentration selection. The two refresh rules differ in their memory of previous signals.
The EMA online variant updates output-signal moments with decay $.8$ and bias correction. It applies concentration 300 to the current unstandardized coherence:
\begin{equation}
 w^{(k+1)}=B\operatorname{softmax}(300\,\widehat c^{(k)}),\qquad
 q^{(k+1)}=w^{(k+1)}/B.
 \label{eq:online}
\end{equation}
The estimated profile is deployed at the next update with recalibrated $a(q,1)$. Concentration 300 acts on raw coherence, whereas static concentration acts on the standardized statistic. Sampling and initialization settings are in Appendix~\ref{app:method-settings}.

Online improves SimCLR over its $x_0$ control, whereas CLIP, Auto-A35, and Mixed have lower online mean peaks than their controls. Digit7's online mean is higher than $x_0$, with mixed paired differences; one online run stops early and the others complete the budget. These comparisons retain successful and unfavorable allocations.

\paragraph{Full replacement of the online profile.}
A second variant starts from $x_0$ and pools the latest 20 completed fit batches. It forms a fresh pairwise coherence estimate $\widehat c^{\rm window}_b\in[0,1]$ and replaces the entire profile every 20 updates:
\begin{equation}
 q_b^{\rm next}=\frac{\max\{\operatorname{softmax}(\eta_{\rm raw}\widehat c^{\rm window})_b,10^{-6}\}}
 {\sum_j\max\{\operatorname{softmax}(\eta_{\rm raw}\widehat c^{\rm window})_j,10^{-6}\}}.
\end{equation}
The window resets after each refresh, with no EMA or persistent base. Deployment begins at the next update and uses the recomputed common-bank scalar. Here $\eta_{\rm raw}$ multiplies bounded raw coherence, without the across-bin population-SD standardization of Equation~\eqref{eq:static}.

The concentration sweep covers $.25,.5,.75,1,1.25$ for Auto-A35 and CLIP, and one for Digit7. Even the best concentration per reward remains below its matched $x_0$, selected-static, and EMA-online controls. All 33 runs stop by the degradation criterion: 15/15 Auto-A35 runs at updates 200--300, 15/15 CLIP runs at 200--740, and 3/3 Digit7 runs at 200--300. Table~\ref{tab:full-refit-summary} retains the best observed means and uncertainty. The failures concern this refresh rule and raw-coherence scale; they do not establish a general failure of online weighting.

\subsection{Training paths and stage-dependent gains}
\label{app:schedules}
Prescribed schedules improve all three rewards relative to their best tested fixed profiles, demonstrating the importance of allocation history (Figure~\ref{fig:dynamic-paths}).
The schedules vary the same calibrated family used for fixed-profile comparisons. With update $k$, define
\begin{equation}
 \alpha_k=\alpha_{\rm early}+(\alpha_{\rm late}-\alpha_{\rm early})
 \operatorname{clip}\!\left(\frac{k-k_0}{k_1-k_0},0,1\right).
\end{equation}
The Mixed/CLIP transitions span updates 160--260; Digit7 spans 200--400. Model and optimizer states continue across the transition. Complete endpoint exponents, seeds, and training constants are in Appendix~\ref{app:method-settings}; aggregate arm outcomes are in Appendix~\ref{app:dynamic-results}.

\paragraph{Best tested fixed-$\alpha$ weightings and observed stopping.}
Fixed $-3$ has the highest mean peak in the eleven-profile Mixed grid, and all three seeds reach 800. The fixed-$-.5$ comparison uses three completed full-budget trajectories. Fixed $-3$ also has the highest mean peak in the eleven-profile CLIP grid. The CLIP fixed controls naturally stopped following collapse, which is part of their observed learning behavior. Their peaks are retained, without extrapolating unobserved tails or assigning an update-800 return. Digit7's fixed $0$ is the mean-peak winner in the separate nine-profile full-800 grid; the separate fixed-$+1$ control additionally illustrates poor learning from that initialization-time allocation.  Figure~\ref{fig:dynamic-paths} plots three-seed means only on the common observed nodes, whereas peak summaries use each seed's full observed trajectory.

\paragraph{Schedule outcomes and complementary static construction.}
Mixed exceeds fixed $-3$ in all three seeds. Its mean gain over fixed $-.5$ is concentrated in the first control seed, which deteriorates substantially. The larger-span CLIP schedule exceeds fixed $-3$, fixed $-2$, and fixed $0$ in all three observed seed peaks. Digit7's $-2\rightarrow+1$ schedule exceeds fixed $0$, fixed $-2$, and the $-2\rightarrow0$ schedule in all three seeds for both peak and update-800 reward. The $-2\rightarrow0$ schedule itself remains below fixed $0$ in the mean, with mixed paired signs.

Static softmax provides another useful CLIP allocation (Table~\ref{tab:dynamic}). Its selected mean peak is close to the larger-span schedule; their paired peak differences have two positive signs and one negative sign. The schedule experiment tests the value of changing the training path, whereas static softmax tests construction of a frozen reward-conditioned profile. Neither result requires a general ranking of the two procedures.

\subsection{Robot-control schedules in native coordinates}
\label{app:robotics-dynamic}
Figure~\ref{fig:robotics-dynamic} compares Acrobot schedule A20 and Walker schedule W5 by their mean per-run observed peaks. Both use the linear interpolation in Appendix~\ref{app:schedules} with training-block index $k$, $(k_0,k_1)=(12,36)$, and horizon $H=244$ blocks. Their native-noise exponents are respectively $1\to .25$ and $1\to .5$, constant before and after the transition. This window spans approximately $5$--$15\%$ of training, following initial acquisition. Blocks measure robot training progress, not generative timesteps. Both schedules have six available runs; all peaks use Equation~\eqref{eq:peak}, rather than the peak of the mean learning curve.

Acrobot's mean peak is $208.097670$ versus fixed $1$: $163.631157$ ($n=3$), fixed $.25$: $184.474308$ ($n=7$), and best ordinary fixed $.5$: $193.492714$ ($n=9$). Walker's mean peak is $780.010782$ versus fixed $1$: $713.319045$ ($n=14$) and fixed $.5$: $760.164659$ ($n=10$), which is both its end profile and best ordinary fixed reference. Gains over these best fixed-reference means are $14.605$ and $19.846$. The fixed references pool all ordinary runs, including reruns. The comparison is descriptive; cohort-specific results and additional controls are in Appendix~\ref{app:robotics-dynamic-results}.

\paragraph{Direction in common velocity coordinates.}
With $x_t=(1-t)\epsilon+tx$ and $\sigma=1-t$, prediction errors obey $\ell_\epsilon=t^2\ell_v$. Ignoring stabilizers and normalization constants, the native FPO loss-difference coefficient has shape
\begin{equation}
 w_{\alpha_\epsilon}(\sigma)\ell_\epsilon
 \ \propto\ \sigma^{2(1-\alpha_\epsilon)}(1-\sigma)^2\ell_v.
\end{equation}
Decreasing $\alpha_\epsilon$ increases relative high-noise emphasis while retaining the native low-noise factor $(1-\sigma)^2$; $\alpha_\epsilon=1$ is uniform only as a multiplier of the noise-prediction loss. CIFAR uses velocity coordinates, so its increasing-$\alpha$ schedules move in the opposite direction. These are loss-coordinate identities, not an equivalence of nonlinear clipped RL updates. The schedules and cosine diagnostics motivate the behavioral-gap interpretation without establishing saturation or coordination as the cause of the gains.

\subsection{Fixed profiles across longer training budgets}
\label{app:robotics-budget}
\begin{figure}[ht]
\centering
\includegraphics[width=\linewidth]{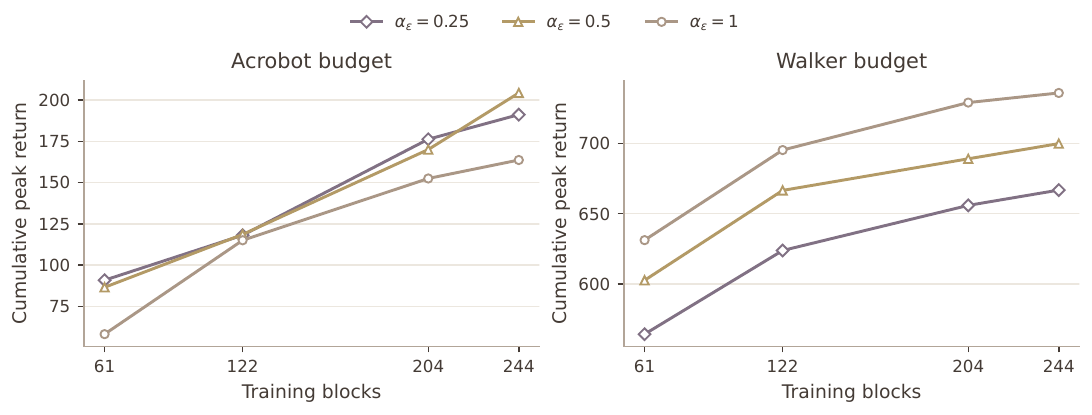}
\caption{Matched three-seed long-budget FPO cohort: mean of each run's cumulative evaluation peak through four training-block budgets for fixed $\alpha_\epsilon\in\{.25,.5,1\}$. Acrobot's leading profile changes from $.25$ at 61 blocks to $.5$ at 122, back to $.25$ at 204, and to $.5$ at 244; Walker retains $1>.5>.25$ throughout. Lines connect budgets, not changing-weight schedules. Each seed contributes a single run, so values can differ from Figure~\ref{fig:robotics}, which averages repeated runs per seed; the $\alpha_\epsilon=0$ arm is reported in the text.}
\label{fig:robotics-budget}
\end{figure}

The long-budget study evaluates Acrobot and Walker with $\alpha_\epsilon\in\{0,.25,.5,1\}$ using three training seeds over 244 blocks. Figure~\ref{fig:robotics-budget} reports cumulative observed peaks from this same cohort at four budgets. Walker $\alpha_\epsilon=0$ has two divergent runs, at blocks 151 and 239; we report this event rather than ranking the surviving run as a complete three-seed arm. The fixed-61-block and long-budget cohorts are presented separately. This single-run-per-seed long-budget cohort also differs from the repeat-averaged cohort in Figure~\ref{fig:robotics}, where Walker peaks at $\alpha_\epsilon=.5$.

Figure~\ref{fig:robotics-budget} shows how these fixed-profile curves change across training stages. Acrobot's low-exponent advantage narrows and then widens, with the leading profile moving within that region. Walker retains the ordering $\alpha_\epsilon=1>.5>.25$ among the plotted complete arms at all four budgets. These comparisons follow the same matched cohort across budgets; each run keeps its exponent fixed throughout training.

\subsection{Four-candidate selection and training cost}
\label{app:halving}
Four-candidate selection uses half the exhaustive training budget and exceeds native-target references on the four evaluated CIFAR rewards, while CLIP remains below its best fixed-sweep reference (Figure~\ref{fig:halving}). Candidates are ranked by the current rung's validation mean at updates 200 and 400, with the winner continued to 800. Complete tournament settings are in Appendix~\ref{app:search-settings}; robotics selection summaries are in Appendix~\ref{app:halving-results}.

\paragraph{Sampling and primary metric.}
This experiment uses the $q+Z$ estimator: sample bin $b$ with probability $w_b/\sum_c w_c$, sample uniformly inside that bin, and multiply the sampled objective only by $Z=\operatorname{mean}_b w_b$. There is no second explicit timestep weight or common-bank calibration multiplier $a(q,g)$ in this experiment. The primary metric is the selected winner's observed peak over its own fixed-validation nodes $0,20,\ldots,800$, averaged across the two seeds. Selection and reporting use the same bank; the result is not an independent post-selection confirmation. Appendix~\ref{app:halving} retains the seed peaks and their update indices, at six-decimal precision.

\paragraph{Training cost and reference comparisons.}
Each tournament executes $4(200)+2(400-200)+(800-400)=1600$ optimizer updates, versus $4(800)=3200$ for exhaustive training of the same four candidates: 50\%. The eight tournaments therefore total 12,800 updates, versus 25,600. This counts training updates, excluding evaluation runtime; there are no additional confirmation runs in these totals. Fixed-profile reference sweeps are excluded from the tournament search cost. Eliminated arms have no full-budget continuation, so the best-fixed points are separate-cohort reference values, not an oracle on the tournament seed pair. Digit7 uses the seven-profile shape--scale sweep at $g=1$ for velocity, $x_0$, and best-$\alpha=-.5$, consistently with Figure~\ref{fig:static}. The matched method-cohort and nine-profile full-800 controls belong to separate comparisons.

For Figure~\ref{fig:halving}, one full-run equivalent is 800 optimizer updates per replication. Thus fixed velocity and $x_0$ cost $800/800=1$, SH4 costs
\begin{equation}
 C_{\rm SH4}=\frac{4\cdot200+2(400-200)+(800-400)}{800}=2,
\end{equation}
and the historical best-fixed sweep costs $K\cdot800/800=K$, with $K=11$ for Auto-A35, CLIP, and SimCLR and $K=7$ for Digit7. These are nominal caps; actual historical validation early-stopping savings are unknown. The matched four-candidate exhaustive budget is four equivalents, but no corresponding matched performance is available, so no point is plotted there. Historical fixed references use $n=3$ and the calibrated $q+a$ estimator, whereas SH4 uses $n=2$ and $q+Z$; their cohorts and calibration differ. The figure retains reported mean per-seed peak validation rewards and uses a logarithmic CIFAR cost axis.

\subsection{Retrospective robotics successive-halving replay}
\label{app:halving-robotics}
Replay illustrates both useful early selection and misselection: Acrobot and Walker improve over their native references, while a 20\% first cut misses Spot's full-budget winner. Profiles are selected by successively halving a common resampled seed bank; candidate sets, cuts, and seed sampling are specified in Appendix~\ref{app:search-settings}.

The FPO reference peaks in Figure~\ref{fig:halving} differ from Figure~\ref{fig:robotics} by up to five return units; we retain the replay-specific references and within-seed aggregation defined below.

Let $P_i(\alpha)$ be the full recorded evaluation peak for training seed $i$. When a seed has repeated FPO runs, their returns are first averaged at each evaluation block and $P_i$ is the peak of that averaged trajectory. The fixed empirical reference is $\overline P(\alpha)=\tfrac13\sum_{i=1}^3P_i(\alpha)$. If $\pi_{n,X}(\alpha)$ is the replay selection probability, the reported selected score is
\begin{equation}
 \E[\overline P(\widehat\alpha)]=\sum_\alpha\pi_{n,X}(\alpha)\overline P(\alpha).
\end{equation}
This is the expected score of the selected profile on the empirical bank. Independent empirical scoring refers to draws from that bank, rather than fresh held-out training seeds. The best fixed-reference score $\max_\alpha\overline P(\alpha)$ is a separate oracle summary.

At the highlighted 20\% first-cut budget, Acrobot selects $\alpha_\epsilon=0,.25,.5$ with probabilities $11/27,15/27,1/27$. Walker selects $.5,1,1.25,1.5$ with probabilities $17/27,3/27,6/27,1/27$. Spot selects $\alpha=.5$ with probability one, below its full-budget native winner. Table~\ref{tab:halving-choice} separates the selected expectation from the fixed oracle. Alternative cut budgets are retained in Table~\ref{tab:halving-budgets}; a first cut at $40\%$ selects Spot's native $\alpha=1$ with probability one.

The training cost is $C=\sum_{\alpha,i}T_{\alpha,i}/H$, where $T_{\alpha,i}$ is each candidate--draw replicate's actual stopping budget, including eliminated candidates, and $H$ is one full candidate--seed budget. FPO costs use environment steps and Spot uses completed updates. Table~\ref{tab:halving-budgets} compares these costs with exhaustive search. The accounting describes training resources simulated by the replay, excluding evaluation runtime and fresh post-selection retraining. The native reference is a score comparison; the stated cost reduction is relative to exhaustive profile search.

Figure~\ref{fig:halving} displays all nine saved first-cut budgets at $n=3$, using the stored cost per replication,
\begin{equation}
 C_{\rm fig}=\frac{C}{n}=K f_{\rm exhaustive},
\end{equation}
where $f_{\rm exhaustive}$ is the fraction of exhaustive-search training cost and $K=4,5,4$ for Acrobot, Walker, and Spot. One full run spans 244 recorded FPO blocks or 1,500 completed Spot updates. Native costs one equivalent; best-fixed and full-search references cost $K$. All candidates' cumulative search training is counted, not just the selected checkpoint's time to peak. Full-run equivalents do not imply equal FLOPs, wall time, or absolute training across tasks; the 27 ordered bootstrap draws are not 27 new training runs. Curves connect only SH budgets. The native and best-fixed points retain $\overline P(\alpha_{\rm native})$ and $\max_\alpha\overline P(\alpha)$, while SH and full-search points retain $\E[\overline P(\widehat\alpha)]$. Full-search selection and best-fixed performance are therefore distinct estimands, shown as independent reference markers even when they nearly coincide.

Scoring uses the fixed empirical mean peak defined above. Ranking outcomes by the peaks attained on the resampled seeds instead measures performance on those selection draws.

\clearpage
\section{V-GRPO: Held-Out Evaluation and Training Dynamics}
\label{app:vgrpo}
\label{sec:vgrpo}
\begin{figure}[ht]
\centering
\includegraphics[width=\linewidth]{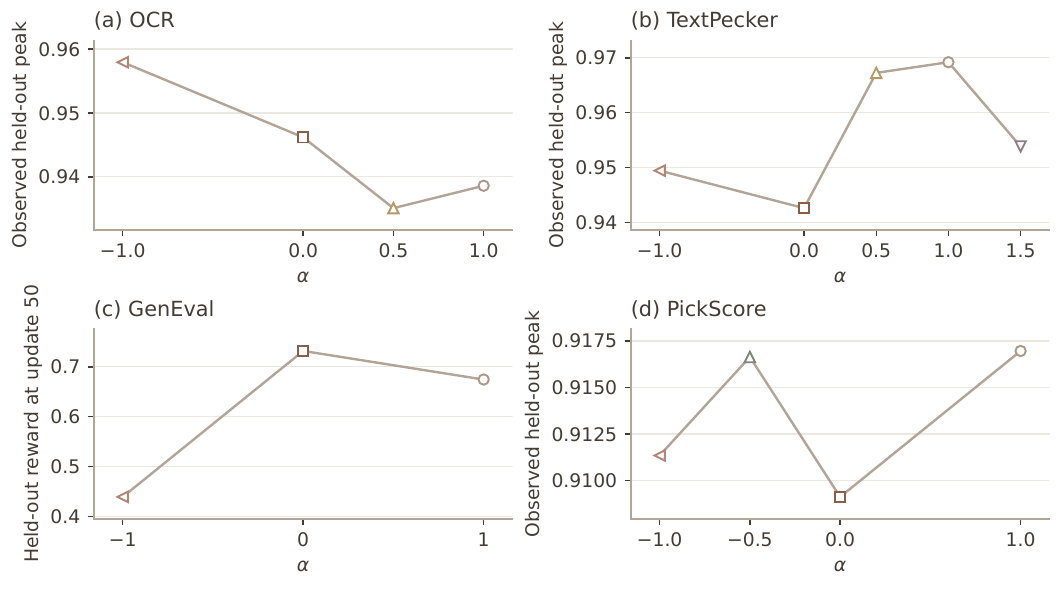}
\caption{V-GRPO image rewards versus timestep exponent $\alpha$: held-out observed reward from one training seed per arm. Preferred exponents differ by reward, extending the task-dependent weighting pattern beyond CIFAR; the displayed values are descriptive rather than replicated comparisons. OCR/TextPecker use peaks over updates 5--25, GenEval the update-50 value, and PickScore peaks over updates 100--300; trajectories are in Figure~\ref{fig:vgrpo-horizons}.}
\label{fig:vgrpo}
\end{figure}

Figure~\ref{fig:vgrpo} summarizes reward- and horizon-dependent weighting preferences in text-to-image adaptation.

OCR's early and middle behavior favors moderate exponents, while $\alpha=-1$ catches up late and attains the highest observed held-out peak. TextPecker favors $.5$--$1$ on held-out data. The matched short training runs favor velocity over clean-target weighting in training reward; this is a different statistic from the held-out comparison in Figure~\ref{fig:vgrpo}.

GenEval evaluates the official detector-backed compositional generation reward. Its held-out terminal scores favor the clean-target profile (Figure~\ref{fig:vgrpo}). PickScore provides another real-model example on a fixed 256-prompt bank, summarized in Figure~\ref{fig:vgrpo}. Velocity and $\alpha=-.5$ reach nearly equal observed peaks. At update 200, velocity has the highest of these four scores; by update 300, $-.5$ and $-1$ exceed it. Each comparison uses one training seed.

\begin{figure}[ht]
\centering
\includegraphics[width=\linewidth]{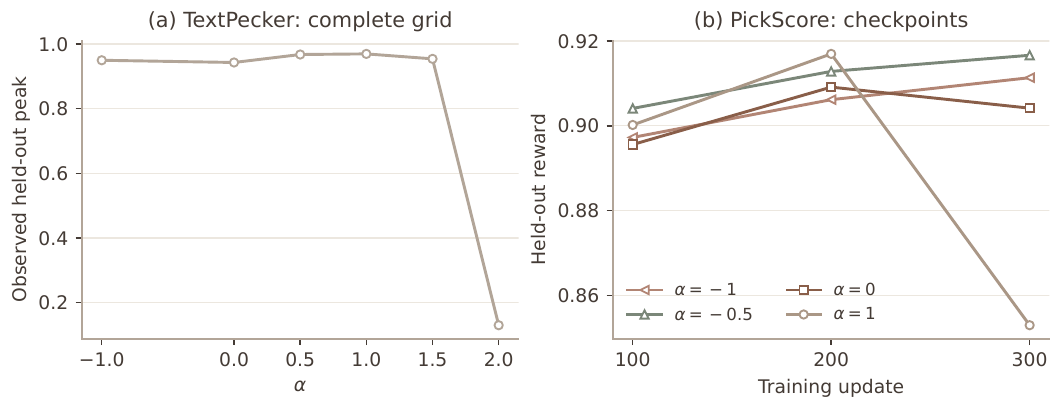}
\caption{One-seed V-GRPO held-out evaluations: TextPecker across tested exponents (left), and PickScore trajectories at updates 100, 200, and 300 (right). TextPecker's $\alpha=2$ arm collapses; PickScore's $-.5$ and velocity arms nearly tie in observed peak while negative exponents retain stronger update-300 scores. These checkpoints explain the peak summaries in Figure~\ref{fig:vgrpo} without using training maxima.}
\label{fig:vgrpo-horizons}
\end{figure}

\paragraph{Further method exploration.}
Our V-GRPO exploration showed substantial seed-to-seed variability, while the selected rewards were generally relatively saturated. Together with the high computational cost, these observations limited effective exploration of weighting methods. They motivate future work on more challenging rewards and better timestep-weighting search methods for V-GRPO.

\paragraph{Self-normalization and implicit allocation.}
V-GRPO's self-normalization rescales residual contributions according to the native branch's loss statistics. Because these statistics depend on time, the operation can induce a different effective timestep allocation. Its interaction with target parameterization is therefore an example of the same design dimension.

\clearpage
\section{Models, Datasets, and Experimental Settings}
\label{app:settings}
\paragraph{Peak evaluation metric.}
For CIFAR and robotics, the primary summary is the mean of per-run observed evaluation peaks,
\begin{equation}
 \overline P=\frac1n\sum_{i=1}^n\max_{k\in\mathcal K_i}\widehat R_{i,k},
 \label{eq:peak}
\end{equation}
where $\mathcal K_i$ contains the saved evaluation checkpoints for run $i$. Here $\widehat R_{i,k}$ is the reward on fixed CIFAR validation latent banks or the robot-control evaluation return. Each run's peak is taken before averaging across runs; this differs from the peak of the mean learning curve. Numerical tables report uncertainty and cohort sizes.

\subsection{CIFAR model, optimization, and evaluation}
\label{app:cifar}
\paragraph{Pretrained generator and data.}
The shared starting model is the publicly released unconditional CIFAR-10 flow-matching checkpoint \href{https://huggingface.co/FrankCCCCC/cfm-cifar10-32}{\texttt{FrankCCCCC/cfm-cifar10-32}}. Its model card identifies CIFAR-10 RGB images and flow-matching pretraining; it does not document the full pretraining recipe or augmentations. The released U-Net has channel widths $(128,256,256,256)$, two residual layers per block, attention at the second downsampling level and mirrored upsampling level, SiLU activations, and 32-group normalization; class embeddings are disabled. Each fresh run starts from the same pretrained state; reward/profile changes do not select a different generator. The prediction convention maps $\sigma=1-t$ and negates the checkpoint output to obtain data-minus-noise velocity. Rewards convert generated images from $[-1,1]$ to $[0,1]$, then apply their scorer-specific preprocessing in Appendix~\ref{app:rewards}. RL updates use generated rollouts; CIFAR-10 is the generator's pretraining dataset, with training images also used for the fixed Edge calibration.

The controlled model generates $3\times32\times32$ images with Euler sampling over 20 flow steps. Training uses batch size 64, advantage groups of 16, four Monte Carlo time/noise draws, and one inner optimization epoch per rollout. The velocity loss is normalized by 3,072 dimensions. Adam uses learning rate $10^{-5}$, betas $(.9,.999)$, epsilon $10^{-8}$, and zero weight decay; global gradient clipping is $.5$ and the PPO-style ratio clip is $.08$. Training is FP32. Three training seeds are used in each replicated main-image cell.

Algorithm~\ref{alg:cifar} uses contiguous groups of $G=16$ generated images, with $s_{\mathcal G}^2=(G-1)^{-1}\sum_{j\in\mathcal G}(R_j-\overline R_{\mathcal G})^2$ and $\varepsilon_A=10^{-6}$. The old model is a frozen evaluation-mode copy made at each outer update. Generated samples, old losses, advantages, and calibration are detached. Rollout generation and reward scoring use no gradient; the same endpoints, auxiliary times, and Gaussian draws enter both prediction losses. Equation~\eqref{eq:rl-mc} uses the same $d=3072$ normalization, four draws, and scalar $a(q,g)$ for both models. The log surrogate is clamped to $[-20,20]$ before exponentiation, followed by ratio clipping to $[.92,1.08]$ and the advantage-weighted minimum in Equation~\eqref{eq:rl-ppo}. One gradient-clipped Adam step follows per rollout; no explicit KL penalty is present. The auxiliary $t$ draws are separate from the fixed 20-step generation grid.

Validation uses four fixed latent banks of 128 images per training seed, evaluated every 20 updates. The fixed-scale and joint-grid studies run for at most 800 updates, with validation-based stopping for plateau or degradation after a minimum of 200. The schedules and designated full-budget controls run 800 updates and have 41 validation nodes including initialization; early-stopped controls retain their observed horizons (Appendix~\ref{app:schedules}). Stopping and checkpoint selection use the observed validation curve; the reported peak is therefore an observed selection metric on that bank. The full-budget Digit7 comparison uses its own fixed bank, separate from the shape--scale grid.

For image training, the twelve mean-one bin weights define a categorical sampler, followed by uniform sampling within the selected bin. The sampled velocity loss receives only the whole-loss scalar $a(q,g)$: the bin weight is already represented in the sampling probability. The calibration uses 512 fixed examples and $C_{x_0}=0.12348001494617646$. Diagnostic estimators and their training cohorts are specified separately in Appendix~\ref{app:cifar-gradients}.

\paragraph{Discrete power family.}
\label{app:discrete-alpha}
A stabilized continuous idealization of Equation~\eqref{eq:alpha} is $\widetilde w_\alpha(t)=[(1-t)^2+\epsilon_0]^{1-\alpha}/\E_{q_0}[((1-t)^2+\epsilon_0)^{1-\alpha}]$. The executed construction below instead normalizes the midpoint clean-target profile before adding its stabilizer.
Define $\operatorname{meanone}(z)_b=z_b/(B^{-1}\sum_j z_j)$, with $B=12$, midpoint $t_b=(b-\tfrac12)/B$, and $b=1,\ldots,B$. The implemented clean-target profile is evaluated at these midpoints:
\begin{align}
 x_b&=\operatorname{meanone}\bigl(((1-t_j)^2)_{j=1}^B\bigr)_b,\qquad v_b=1,\\
 w_{\alpha,b}&=\operatorname{meanone}\!\left(\exp\!\left[(1-\alpha)\log(x+10^{-12})+\alpha\log(v+10^{-12})\right]\right)_b,
 \label{eq:discrete-alpha}\\
 q_{\alpha,b}&=w_{\alpha,b}/B.
\end{align}
All vector operations are elementwise. The velocity term is constant across bins and cancels in the final normalization; the calibrated implementation omits it. Thus normalization of the midpoint clean-target profile precedes stabilization and exponentiation, followed by a second mean-one normalization. Equation~\eqref{eq:alpha} gives the common unstabilized power shape; Equation~\eqref{eq:discrete-alpha} specifies the discrete bin construction, rather than integrating the continuous powered profile within each bin.

\subsection{Robot-control environments and training}
\label{app:robotics}
\paragraph{Environments and policy inputs.}
FPO uses the DeepMind Control tasks implemented in MuJoCo Playground \citep{zakka2025playground}: \texttt{AcrobotSwingup}, \texttt{WalkerRun}, \texttt{FishSwim}, and \texttt{BallInCup} in the fixed-budget summaries; \texttt{CheetahRun} and \texttt{SwimmerSwimmer6} additionally appear in the gradient atlas. FPO++ uses Isaac Lab \citep{mittal2025isaaclab}: Cartpole maps to \texttt{Isaac-Cartpole-Direct-v0}; Spot, G1, and Go2 map to \texttt{Isaac-Velocity-Flat-Spot-v0}, \texttt{Isaac-Velocity-Flat-G1-v0}, and \texttt{Isaac-Velocity-Flat-Unitree-Go2-v0}. The task configurations specify 512 parallel environments for Cartpole and 2,048 for the locomotion tasks. Both policies condition action generation on observations; FPO normalizes observations using the policy's running statistics. Training data are online environment transitions. Robot policies are trained from scratch.

FPO uses noise coefficients $s\in\{.1,.2,\ldots,1\}$, equivalently $t\in\{0,.1,\ldots,.9\}$. Its weight table is mean-one on the executed support. The point $s=1$ has zero velocity-coordinate target factor $t^2$, leaving nine active bins for gradient analysis. A block contains 983,040 environment steps; 61 blocks correspond to 59,965,440 steps. The extended fixed-budget table uses eight seeds for Acrobot/Walker and five for Fish/Ball.

FPO++'s flow-time interval is $[.005,.995]$, with analytic mean-one normalization on that interval. Cartpole uses 150 updates, Spot 1,500, and evaluation uses zero-initialized flow sampling. Zero initialization is an inference-time sampling choice for an already trained policy. The underlying algorithms retain their native adaptive optimizers and trust-region constructions.

\subsection{Weighting-method configurations and comparison references}
\label{app:method-settings}
All six static-softmax searches use $\eta_{\rm norm}\in\{.25,.5,.75,1,1.25,1.5,2\}$, three seeds, $q$-sampling, and the common-bank $x_0$ scale anchor. The first five rewards use the fixed-validation and stopping protocol of Appendix~\ref{app:cifar}, with profiles fitted to each reward. The separate Digit7 method cohort uses the same sampling/scale convention, ratio clip $.08$, seeds 20260724--20260726, four fixed validation banks of 128 images, and an 800-update cap with validation-based stopping after update 200. Its $x_0$ and online controls share that cohort's reward and bank (Table~\ref{tab:online-full}). Figure~\ref{fig:static} instead compares the static result with fixed profiles from the seven-profile shape--scale sweep at $g=1$ (Figure~\ref{fig:shape-scale}). All three fixed bars use that one sweep, including true velocity at $\alpha=1$ and its best mean-peak profile at $\alpha=-.5$. The fixed and static cohorts use separate validation banks. The nine-profile full-800 landscape in Figure~\ref{fig:landscape} is a third cohort.

\paragraph{Online sampling and initialization.}
Each refresh uses one fit batch of 64 endpoints in groups of 16, with four time/noise draws and 20 flow-sampling steps. It refreshes every update, deploys the result in the next update, and recomputes $a(q,1)$ for the new profile. The five-reward comparison starts from uniform weighting. Digit7 instead starts from $q_{x_0}$; that initialization is not a persistent multiplicative base in subsequent refreshes. Concentration 300 operates on raw coherence, whereas the static $\eta_{\rm norm}$ operates on a standardized across-bin statistic.

\paragraph{CIFAR schedules.}
Mixed uses $(\alpha_{\rm early},\alpha_{\rm late},k_0,k_1)=(-3,-.5,160,260)$; CLIP uses $(-3,0,160,260)$; and the larger-span Digit7 schedule uses $(-2,1,200,400)$. The Digit7 $-2\rightarrow0$ schedule uses $(-2,0,200,400)$. Each recomputes the analytical bin probabilities and common-bank scalar as $\alpha_k$ changes, without resetting model parameters or optimizer state. The three seeds are 20260724--20260726, with 800 updates and validation every 20 updates, including initialization. The CLIP/Digit7 schedules use batch size 64, group size 16, four Monte Carlo time samples, 20 sampling steps, learning rate $10^{-5}$, ratio clip $.08$, and gradient clip $.5$.

\paragraph{Configuration selection.}
The fixed-profile study evaluates eleven exponents per reward in its main grid. Static concentrations are selected using mean per-seed validation peaks, with all three seeds retained in the selected result. EMA-online concentration is fixed at 300; full-refit concentration uses the raw-coherence grid in Appendix~\ref{app:online-analysis}. The six static sweeps contain 126 reward--concentration--seed cells in total. The schedules were designed from already observed learning curves as directed path interventions, rather than learned online controllers. The comparison describes the resulting trained configurations; total method-selection cost also includes the configurations evaluated to select them.

\subsection{Successive-halving configurations}
\label{app:search-settings}
\paragraph{CIFAR tournaments.}
The four-candidate experiment consists of four rewards (Auto-A35, CLIP, SimCLR, Digit7) and two complete tournaments per reward, using seeds 20260727 and 20260728. Each starts with $\alpha\in\{-2,-1,0,1\}$, retains two arms at update 200, retains one at update 400, and continues that winner to update 800. Ranking uses only the current rung's pooled fixed-validation mean; exact ties favor the smaller exponent. Surviving arms continue their existing model and optimizer state. All four arms within a tournament share the training seed and fixed validation bank, with fresh initial model/Adam states; the two seeds have different bank realizations.

\paragraph{Robot-control replay.}
The recorded replay uses FPO Acrobot and Walker fixed-arm runs through 244 blocks and the FPO++ Spot cohort through 1,500 updates. FPO uses seeds 901--903; repeated runs for the same arm and seed are averaged pointwise before peaks are taken. The candidate sets are Acrobot $\alpha_\epsilon\in\{0,.25,.5,1\}$ (there is no H=244 $.75$ arm), Walker $\alpha_\epsilon\in\{.25,.5,1,1.25,1.5\}$, and Spot $\alpha\in\{0,.5,1,1.5\}$. The FPO bank is the ordinary fixed-arm subset underlying Figure~\ref{fig:robotics-dynamic}, with a replay-specific within-seed aggregation; Spot is unchanged.

For $n\in\{1,2,3\}$, the replay enumerates all $3^n$ ordered seed draws with replacement, pairing the same draw across every candidate. At each cut target $X,2X,4X,\ldots<100\%$, it uses the last saved checkpoint at or before that budget, ranks candidates by the mean of their sampled per-seed peaks observed so far, and retains $\lceil K/2\rceil$. Ties are averaged over candidate orderings. Final survivors are ranked at the full budget. The first-cut grid is $X\in\{2,5,10,15,20,25,30,40,50\}\%$, plus a no-halving exhaustive-search reference.

\subsection{V-GRPO evaluation settings}
\label{app:vgrpo-settings}
The OCR and TextPecker evaluations use saved held-out checkpoints at updates $5,10,15,20,25$ with 1,018 scored samples per checkpoint and one training seed per arm. Effective batch size is held constant in the TextPecker comparison, across negative and larger exponents. The reported peak is the largest held-out score among these five checkpoints; no training-reward maximum is substituted.

The OCR experiment uses global batch 48, four Monte Carlo draws, and no self-normalization. GenEval uses the official detector-backed compositional reward; its reported held-out value is terminal at update 50. PickScore uses a fixed 256-prompt bank at updates 100, 200, and 300. Each displayed arm has one training seed.

\clearpage
\section{Supplementary Result Summaries}
\label{app:full-results}
This section provides analytical summaries and panel identities that complement the preceding figures.

\subsection{CIFAR fixed profiles and shape--scale grid}
\label{app:cifar-results}
The fixed-profile landscapes are shown in Figure~\ref{fig:image_landscape}.

The factorial grid uses $\alpha\in\{-1,-.5,0,.5,1,1.5,2\}$ and $g\in\{.5,.75,1,1.5,2\}$, six rewards, and three seeds: 210 configurations and 630 logical seed cells. Shared $g=1$ entries refer to the same runs. The interpretation of shape transfer is in Appendix~\ref{sec:scale}.
\begin{figure}[ht]
\centering\includegraphics[width=\linewidth]{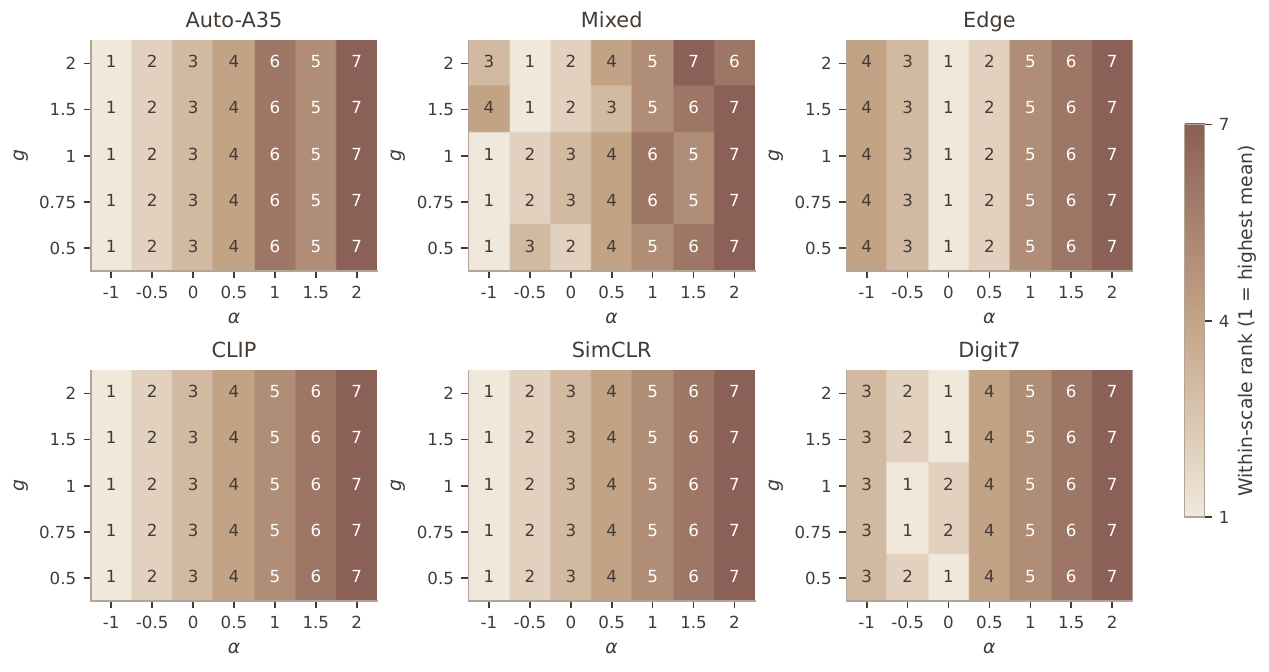}
\caption{Six CIFAR rewards: within-scale ranks of seven $\alpha$ profiles across five calibrated whole-loss scales, computed from mean per-run validation peaks ($n=3$). Most rewards retain similar shape rankings across scales, whereas Mixed and Digit7 change leading profiles, qualifying shape-first tuning. Colors share a rank scale rather than reward units; native reward surfaces are shown in Figure~\ref{fig:shape-scale}.}
\label{fig:scale-ranks}
\end{figure}

\begin{table}[ht]
\centering\small
\caption{Selected shape at each scale and maximum regret of retaining the $g=1$ winner. Winners maximize mean per-seed validation peak on the seven-profile grid; regret is in each reward's native units. These are descriptive grid selections, not an independent transfer experiment.}
\label{tab:scale}
\begin{tabular}{@{}lrrrrrr@{}}
\toprule
Reward & $g=.5$ & $.75$ & $1$ & $1.5$ & $2$ & Max regret \\
\midrule
Auto-A35 & $-1$ & $-1$ & $-1$ & $-1$ & $-1$ & 0.0000 \\
Mixed & $-1$ & $-1$ & $-1$ & $-0.5$ & $-0.5$ & 0.1268 \\
Edge & $0$ & $0$ & $0$ & $0$ & $0$ & 0.0000 \\
CLIP & $-1$ & $-1$ & $-1$ & $-1$ & $-1$ & 0.0000 \\
SimCLR & $-1$ & $-1$ & $-1$ & $-1$ & $-1$ & 0.0000 \\
Digit7 & $0$ & $-0.5$ & $-0.5$ & $0$ & $0$ & 0.0248 \\
\bottomrule
\end{tabular}

\end{table}

\clearpage
\subsection{Robot-control fixed-profile results}
\label{app:robotics-results}
Figure~\ref{fig:robotics} uses a 244-block FPO cohort with seeds 901--903 at every displayed exponent. For seed--exponent cells with repeated runs, evaluation returns are averaged pointwise within seed before taking that seed's peak, and the three seed peaks are then averaged. Its FPO++ panels use the 150-update Cartpole cohort ($n=8$ per exponent) and the 1,500-update Spot cohort ($n=3$--$5$ across displayed exponents). Equation~\eqref{eq:peak} defines the standard per-run peak summary; the repeat-averaged FPO cells use the seed-level trajectory just described as the reporting unit. Appendix~\ref{app:robotics} gives the environment and optimization settings.

Ball's zero-return seed for $\alpha_\epsilon\le0$ is included in the mean and uncertainty. Fish has a broad high-return region across the three evaluated profiles. G1 and Go2 favor the native FPO++ profile in their respective grids.

\clearpage
\subsection{Local response statistics}
\label{app:response-results}
Definitions are in Appendix~\ref{app:mechanisms}.
\begin{table}[ht]
\centering\scriptsize
\caption{Initial fine response at amplitude $4/255$ and mean reward advantage over updates 20--200. $M,C,O$ follow Appendix~\ref{app:mechanisms}; DD is double-decrease percentage. The last two columns compare the fixed $\{-2,-1\}$ average with the same-cohort $x_0$ control. Explicit and $q$-sampling cohorts remain separate.}
\label{tab:response-full}
\begin{tabular}{@{}lrrrrrr@{}}
\toprule Reward & DD (\%) & $M$ & $C$ & $O$ & Explicit $\Delta$ & $q$ cohort $\Delta$\\\midrule
Automobile &49.87&.005621&$-.005213$&.000733&$-.011499$&.108042\\
Auto-A35 &48.70&.004578&$-.004230$&.000638&.018274&.128433\\
CLIP &36.85&.010949&$-.003178$&.009414&.035520&.091558\\
SimCLR &40.76&.016113&$-.005121$&.015405&.027671&.052726\\
Digit7 &46.35&.008746&$-.001750$&.008322&.002082&---\\
Edge &6.25&.000209&.010453&.011644&$-.002609$&$-.003045$\\
Mixed &5.86&.000127&.000050&.000267&---&.055696\\
\bottomrule
\end{tabular}
\end{table}

\begin{table}[ht]
\centering\small
\caption{Multiscale response summaries. The Digit7 rows report mean advantage of the negative-profile pair over $x_0$; the remaining rows are cross-reward descriptive associations.}
\begin{tabular}{@{}lr@{}}\toprule
Statistic & Value\\\midrule
Digit7 $q$, updates 20--40 & $-.004286$\\
Digit7 $q$, updates 20--80 & $-.002453$\\
Digit7 $q$, updates 20--200 & $.014318$\\
Explicit radius 4, $H=200$: $(n,\rho,p)$ & $(6,.60,.24)$\\
Leave-one-reward $\rho$, explicit & $[.40,.90]$\\
Leave-one-reward $\rho$, $q$ cohort & $[-.30,.70]$\\
Explicit endpoint / slope association & $.71\;/\;.89$\\
Explicit double-decrease association & $-.37$\\\bottomrule
\end{tabular}\end{table}

\begin{table}[ht]
\centering\small
\caption{Descriptive cross-reward Spearman associations of standardized fine-damage magnitude with standardized early mean advantage, across six rewards within each cohort. Both quantities are divided by the initial unperturbed reward SD. Radius is in units of $1/255$.}
\label{tab:response-windows}
\begin{tabular}{@{}llrrrr@{}}
\toprule Cohort & Radius & $H=40$ & $80$ & $160$ & $200$\\\midrule
Explicit &2&$-.20$&$-.20$&$-.03$&$.60$\\
Explicit &4&$-.20$&$-.20$&$-.03$&$.60$\\
Explicit &8&$.03$&$.03$&$-.37$&$.14$\\
$q$ cohort &2&$.83$&$.14$&$.14$&$.26$\\
$q$ cohort &4&$.83$&$.14$&$.14$&$.26$\\
$q$ cohort &8&$.89$&$-.03$&$-.03$&$.09$\\
\bottomrule
\end{tabular}
\end{table}

\clearpage
\subsection{Gradient measurements, panel identities, and phase summaries}
\label{app:gradient-results}
Measurement definitions and interpretation are in Appendix~\ref{app:gradients}.
\begin{table}[htbp]
\centering\small
\caption{Profiles and full-trajectory peaks underlying Figure~\ref{fig:geometry}. Paired entries are higher/lower rows. CIFAR checkpoints are optimizer updates and Go2 checkpoints are training iterations.}
\label{tab:heatmap-profile-sources}
\begin{tabular}{@{}lllll@{}}
\toprule Task & $\alpha$ & Checkpoint & Training seeds & Peak reward/return\\\midrule
SimCLR & $-2/1$ & 200 & 20260724 / 20260724 & $.999091/.667570$\\
CLIP & $-2/1$ & 200 & 20260724 / 20260724 & $.895339/.250209$\\
Digit7 & $-1/1$ & 200 & 20260724 / 20260726 & $.621488/.285237$\\
Go2 & $1/0$ & 300 & 9200, 9201 (both rows) & $41.590/39.385$\\\bottomrule
\end{tabular}
\end{table}

At Go2 iteration 300, both seeds separately show broad positive coordination for $\alpha=1$ and local positive blocks with negative clean--noisy coupling for $\alpha=0$. The full-trajectory peaks are $41.59/39.77$ for seed 9200 and $41.59/39.00$ for seed 9201 ($\alpha=1/0$). The equal-seed panel therefore reflects a contrast present in both runs. At this checkpoint the corresponding mean current returns are $14.295/11.605$, distinct from the trajectory peaks.

\paragraph{Training evolution.}
Figure~\ref{fig:training-evolution} retains one profile and seed set within each column. CLIP $\alpha=-2$ and Digit7 $\alpha=-1$ both use seed 20260724 at updates 200, 400, and 800. Go2 $\alpha=1$ uses seeds 9200/9201 and G1 $\alpha=.5$ uses seeds 9900/9901, each at iterations 200, 700, and 1499. Each robot panel averages the two checkpoint matrices equally, without averaging across iterations. Go2's first stage is iteration 200, distinct from the iteration-300 profile comparison. G1's late weak structure is also visible within each seed: a majority of off-diagonal entries have absolute cosine below $.1$, so the near-zero pattern is not just an average of strong opposite-signed blocks. This is a descriptive scale for reading the existing colorbar, not a significance threshold. The complete task/profile phase atlas in Appendix~\ref{app:robotics-atlas} provides supplementary Cartpole and Walker evidence without repeating those panels in the main paper.

\begin{table}[htbp]
\centering\small
\caption{Digit7 at the independent early-prefix checkpoint. Each row is one selected training seed; entries report recorded raw training reward and mean off-diagonal cross-batch cosine. The better-reward branches have stronger agreement than the velocity branch at this checkpoint. These E-prefix states are not joined to the H-prefix trajectories in Figures~\ref{fig:geometry} and~\ref{fig:training-evolution}.}
\label{tab:digit7-early-geometry}
\begin{tabular}{rrrrr}
\toprule
$\alpha$ & Training seed & Update & Training reward & Cross-bin mean \\
\midrule
-2.0000 & 20260725 & 100 & 0.3472 & 0.1448 \\
-1.0000 & 20260724 & 100 & 0.3368 & 0.1520 \\
0.0000 & 20260724 & 100 & 0.3353 & 0.1787 \\
1.0000 & 20260726 & 100 & 0.2886 & 0.0778 \\
\bottomrule
\end{tabular}

\end{table}

\begin{table}[htbp]
\centering\small
\caption{Native-exponent late-phase cross-batch cosine by noise block. High noise is $\sigma\ge .6$ and low noise is $\sigma\le .3$; within-block means omit diagonals. Cross-block means include both symmetric rectangles.}
\label{tab:robotics-cross-blocks}
\begin{tabular}{@{}lrrr@{}}
\toprule Task & High--high & Low--low & High--low\\\midrule
Walker & $.537$ & $.375$ & $-.311$\\
Swimmer & $.400$ & $.723$ & $-.380$\\
Go2 & $.395$ & $.275$ & $-.182$\\\bottomrule
\end{tabular}
\end{table}

\begin{table}[htbp]
\centering\small
\caption{Checkpoint inventory and median off-diagonal within--cross cosine gap, by task. Every checkpoint gap is positive.}
\label{tab:robotics-within-cross}
\begin{tabular}{llrr}
\toprule
Method & Task & Checkpoints & Median within--cross gap \\
\midrule
FPO & Acrobot & 24 & 0.0724 \\
FPO & Ball in Cup & 23 & 0.1050 \\
FPO & Cheetah & 22 & 0.0668 \\
FPO & Fish & 24 & 0.1055 \\
FPO & Swimmer & 24 & 0.0461 \\
FPO & Walker & 24 & 0.0453 \\
FPO++ & Cartpole & 168 & 0.0102 \\
FPO++ & G1 & 144 & 0.2701 \\
FPO++ & Go2 & 176 & 0.1654 \\
FPO++ & Spot & 154 & 0.2403 \\
\bottomrule
\end{tabular}

\end{table}

\begin{table}[htbp]
\centering\small
\caption{Shared-fluctuation summaries over the 80 measured states and after omitting the two ratio-marked states. Correlation summaries first average entries within a state, then take the median across states. State counts are descriptive, not independent training replications.}
\label{tab:shared-fluctuation}
\begin{tabular}{@{}lrr@{}}\toprule
Quantity & All 80 & Omit two $\dagger$\\\midrule
$\operatorname{tr}(N)>\operatorname{tr}(K)$ & 75/80 & 73/78\\
Mean off-diagonal $C^{\rm fluct}>0$ & 80/80 & 78/78\\
Median $S=\operatorname{tr}(K)/\operatorname{tr}(N)$ & 0.17480 & 0.15766\\
Median mean off-diagonal $C^{\rm fluct}$ & 0.09156 & 0.09335\\
Median fraction of positive off-diagonal entries & 80.30\% & 80.30\%\\
States with negative off-diagonal tenth percentile & 68/80 & 66/78\\
Median middle-noise block correlation & 0.19253 & 0.19641\\
Median low-noise block correlation & 0.18213 & 0.18756\\
Median high-to-low-noise block correlation & 0.01309 & 0.01398\\
Paired control exceeds independent same-bin cosine & 75/80 & 73/78\\
Median paired-minus-independent same-bin cosine & 0.18600 & 0.18810\\
\bottomrule
\end{tabular}
\end{table}

\begin{table}[htbp]
\centering\small
\caption{Per-reward medians of the energy ratio $S$ and mean off-diagonal fluctuation correlation. These summarize checkpoint geometry, not a reward-quality ordering.}
\label{tab:shared-fluctuation-rewards}
\begin{tabular}{@{}lrrr@{}}\toprule
Reward & States & Median $S$ & Median mean $C^{\rm fluct}$\\\midrule
Auto-A35 & 20 & 0.09546 & 0.09064\\
CLIP & 20 & 0.19128 & 0.09596\\
SimCLR & 20 & 0.08243 & 0.08937\\
Digit7 & 20 & 0.40711 & 0.09098\\
\bottomrule
\end{tabular}
\end{table}

\clearpage
\subsection{Additional online weighting results}
\label{app:static-results}

\begin{table}[htbp]
\centering\small
\caption{Additional fixed-scale methods: mean per-seed validation peak $\pm$ sample SD ($n=3$). The shared fixed controls and selected static-softmax results are in Figure~\ref{fig:static}; Digit7's distinct matched method-cohort $x_0$ control is retained here. Online starts uniform except in the explicitly labeled Digit7 row.}
\label{tab:online-full}
\begin{tabular}{@{}llr@{}}
\toprule
Reward & Rule & Peak validation reward\\
\midrule
Auto-A35 & Static direct & $0.968577\pm0.039152$\\
Auto-A35 & Online coherence & $0.844915\pm0.259461$\\
Mixed & Static direct & $0.672966\pm0.279437$\\
Mixed & Online coherence & $0.542469\pm0.175938$\\
Edge & Static direct & $0.967028\pm0.002040$\\
Edge & Online coherence & $0.960761\pm0.008673$\\
CLIP & Static direct & $0.368911\pm0.027995$\\
CLIP & Online coherence & $0.328924\pm0.044492$\\
SimCLR & Static direct & $0.875200\pm0.020973$\\
SimCLR & Online coherence & $0.930397\pm0.027878$\\
Digit7 & $x_0$ & $0.674514\pm0.015923$\\
Digit7 & Online coherence ($x_0$ start) & $0.711589\pm0.137812$\\
\bottomrule
\end{tabular}

\end{table}

\begin{table}[ht]
\centering\small
\caption{Best concentration per reward for full-refit online weighting, selected by mean per-seed validation peak. Uncertainty is sample SD, $n=3$. Every run at these selected settings stopped by degradation; update ranges describe these selected settings only.}
\label{tab:full-refit-summary}
\begin{tabular}{@{}lrrr@{}}
\toprule Reward & $\eta_{\rm raw}$ & Peak validation reward & Updates\\\midrule
Auto-A35 & $.5$ & $.3024\pm.0518$ & 200--280\\
CLIP & $.25$ & $.2316\pm.0168$ & 200--440\\
Digit7 & $1$ & $.3161\pm.0428$ & 200--300\\\bottomrule
\end{tabular}
\end{table}
\clearpage
\subsection{CIFAR schedule results}
\label{app:dynamic-results}

\begin{table}[htbp]
\centering\small
\caption{Three-reward path interventions and controls: mean per-seed validation peak and step-800 reward $\pm$ sample SD, three seeds. A dash denotes the absence of a complete three-seed step-800 control, not a zero reward. Peaks use every trajectory's observed evaluations. The Digit7 static-method cohort in Figure~\ref{fig:static} is separate.}
\label{tab:dynamic}
\begin{tabular}{@{}llrr@{}}\toprule
Reward & Allocation & Observed peak & Step 800\\\midrule
CLIP & Fixed $-2$ & $0.852268\pm0.027500$ & ---\\
CLIP & Fixed $-3$ & $0.864639\pm0.012958$ & ---\\
CLIP & Fixed $0$ & $0.387094\pm0.023369$ & ---\\
CLIP & $-3\rightarrow0$ & $0.910677\pm0.011321$ & $0.907311\pm0.010997$\\
CLIP & Static softmax $\eta=1.25$ & $0.899151\pm0.006857$ & ---\\
Digit7 & Fixed $+1$ & $0.311454\pm0.042482$ & ---\\
Digit7 & Fixed $-2$ & $0.601804\pm0.010944$ & $0.594519\pm0.012210$\\
Digit7 & Fixed $0$ & $0.653819\pm0.024828$ & $0.649026\pm0.024737$\\
Digit7 & $-2\rightarrow+1$ & $0.689931\pm0.007325$ & $0.687373\pm0.010188$\\
Digit7 & $-2\rightarrow0$ & $0.641633\pm0.023065$ & $0.638969\pm0.025333$\\
Mixed & Fixed $-.5$ & $0.896055\pm0.139030$ & $0.722721\pm0.439106$\\
Mixed & Fixed $-3$ & $0.907879\pm0.039289$ & $0.907879\pm0.039289$\\
Mixed & $-3\rightarrow-.5$ & $0.987132\pm0.007565$ & $0.987132\pm0.007565$\\
\bottomrule\end{tabular}\end{table}

\clearpage
\subsection{Robot-control schedules: arms and cohort results}
\label{app:robotics-dynamic-results}
\paragraph{Additional controls and coverage.}
Separate Acrobot fixed $.5$ controls reach $214.492$, and Walker's separate-family fixed $.5$ reaches $815.786$ on five available curves. Thus the ordinary reference is not the strongest value across every batch. Acrobot A20's later cohort has mean $190.525$ versus its corresponding fixed $.5$ reference $202.226$; the pooled gain does not assert improvement in every cohort.

Three Walker nonfinite runs lack raw curves but have reported observed-prefix peaks: ordinary fixed $.5$: $800.6$; ordinary fixed $1$: $511.1$; separate-family fixed $.5$: $720.5$. Adding these reported peaks yields mixed-precision means $763.841$ ($n=11$), $699.838$ ($n=15$), and $799.905$ ($n=6$), respectively. These supplements preserve observed pre-failure performance without inventing full curves. Acrobot fixed $0$ also has a recorded nonfinite run and no supplied complete arm curves; no exact pooled value is inferred. 

\clearpage
\subsection{Four-candidate search and robotics replay results}
\label{app:halving-results}

\begin{table}[ht]\centering\small
\caption{Robotics replay in Figure~\ref{fig:halving}, bottom: $n=3$, first cut $20\%$. FPO uses the 244-block seed-901--903 cohort with repeated runs averaged pointwise within seed; Spot retains its 1,500-update cohort. Selected is the expected empirical mean-peak score, oracle is the best fixed-reference score, $\Delta$ is selected minus native, and gap is oracle minus selected. The lower block gives selection probabilities, the oracle exponent, and search/exhaustive fractions in native training budgets. Exponents are $\alpha_\epsilon$ for FPO and $\alpha$ for Spot.}
\label{tab:halving-choice}
\begin{tabular}{@{}lrrrrr@{}}
\toprule
Task & Native & Selected & Oracle & $\Delta$ & Gap\\
\midrule
Acrobot & 163.631157 & 185.187285 & 191.103217 & +21.556128 & 5.915933\\
Spot & 313.813333 & 233.430000 & 313.813333 & -80.383333 & 80.383333\\
Walker & 724.802788 & 742.842999 & 754.124350 & +18.040210 & 11.281352\\
\bottomrule
\end{tabular}
\par\medskip
\begin{tabular}{@{}llrr@{}}
\toprule
Task & Exponent: probability & Oracle exponent & Search / exhaustive\\
\midrule
Acrobot & $0:11/27;\ .25:15/27;\ .5:1/27$ & $0.25$ & 44.6721\%\\
Spot & $.5:1$ & $1$ & 40.0500\%\\
Walker & $.5:17/27;\ 1:3/27;\ 1.25:6/27;\ 1.5:1/27$ & $.5$ & 51.7213\%\\
\bottomrule
\end{tabular}
\end{table}

\begin{table}[ht]
\centering\small
\caption{Sensitivity to first-cut budget in the recorded replay, $n=3$ (244-block FPO cohort; unchanged 1,500-update Spot cohort). The range is the exact minimum/maximum across bootstrap selection draws, not a confidence interval over independently repeated training. Full search still selects from resampled seeds; its expected selected score therefore need not equal the fixed-reference oracle.}
\label{tab:halving-budgets}
\begin{tabular}{@{}llrrr@{}}
\toprule
Task & First cut & Selected score & Exact draw range & $C$ \\
\midrule
Acrobot & 10\% & 182.14 & [176.86, 191.10] & 4.1803 \\
Acrobot & 20\% & 185.19 & [176.86, 191.10] & 5.3607 \\
Acrobot & 40\% & 188.88 & [176.86, 191.10] & 7.7582 \\
Acrobot & Full search & 187.30 & [176.86, 191.10] & 12.0000 \\
Spot & 10\% & 197.73 & [197.73, 197.73] & 3.8060 \\
Spot & 20\% & 233.43 & [233.43, 233.43] & 4.8060 \\
Spot & 40\% & 313.81 & [313.81, 313.81] & 7.2060 \\
Spot & Full search & 313.81 & [313.81, 313.81] & 12.0000 \\
Walker & 10\% & 736.75 & [724.80, 754.12] & 5.3607 \\
Walker & 20\% & 742.84 & [716.29, 754.12] & 7.7582 \\
Walker & 40\% & 746.15 & [716.29, 754.12] & 10.7582 \\
Walker & Full search & 744.31 & [716.29, 754.12] & 15.0000 \\
\bottomrule
\end{tabular}

\end{table}

\clearpage
\section{Full Robotics Heatmap Atlas}
\label{app:robotics-atlas}
The full atlas contains all 44 recorded task--profile groups, each at early, middle, and late phases (132 panels). Measurement and phase aggregation are defined in Appendix~\ref{app:robotics-new-geometry}; complete phase statistics are in Appendix~\ref{app:gradient-results}. Captions retain the reduced-seed coverage after recorded failures.
\begin{figure}[htbp]
\centering
\includegraphics[width=.98\linewidth]{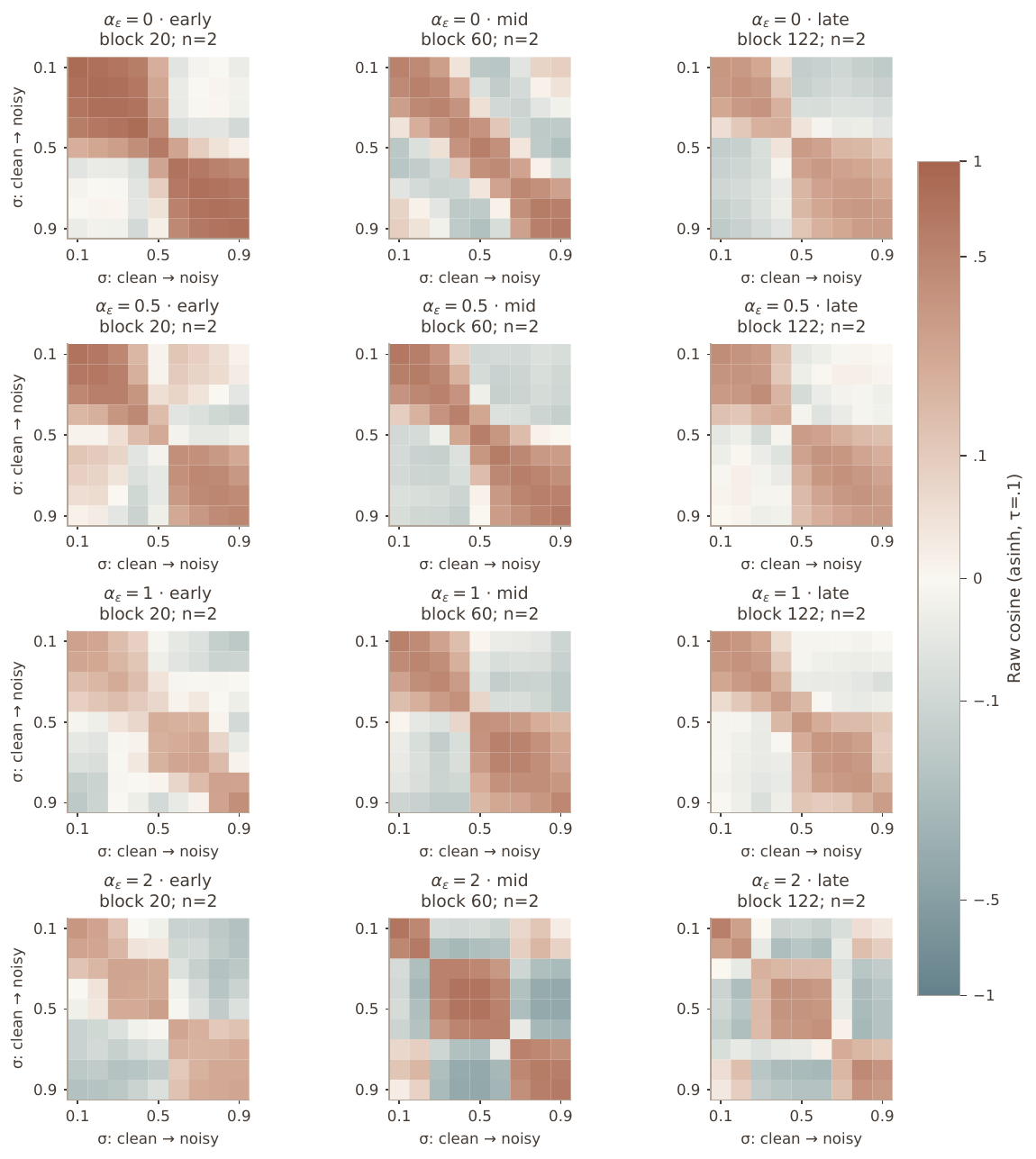}
\caption{Acrobot fixed-profile robot policies: phase-averaged cross-batch gradient cosines across clean-to-noisy bins (columns: early, middle, late; rows: tested exponents). Early cross-noise agreement is stronger at $\alpha_\epsilon=0$ than at 1, but the former weakens by the late phase. Colors show the common raw-cosine scale, and $n$ counts contributing seeds; Appendix~\ref{app:robotics-new-geometry} defines phase aggregation.}
\label{fig:robotics-cross-acrobotswingup-0}
\end{figure}
\clearpage
\begin{figure}[p]
\centering
\includegraphics[width=.98\linewidth]{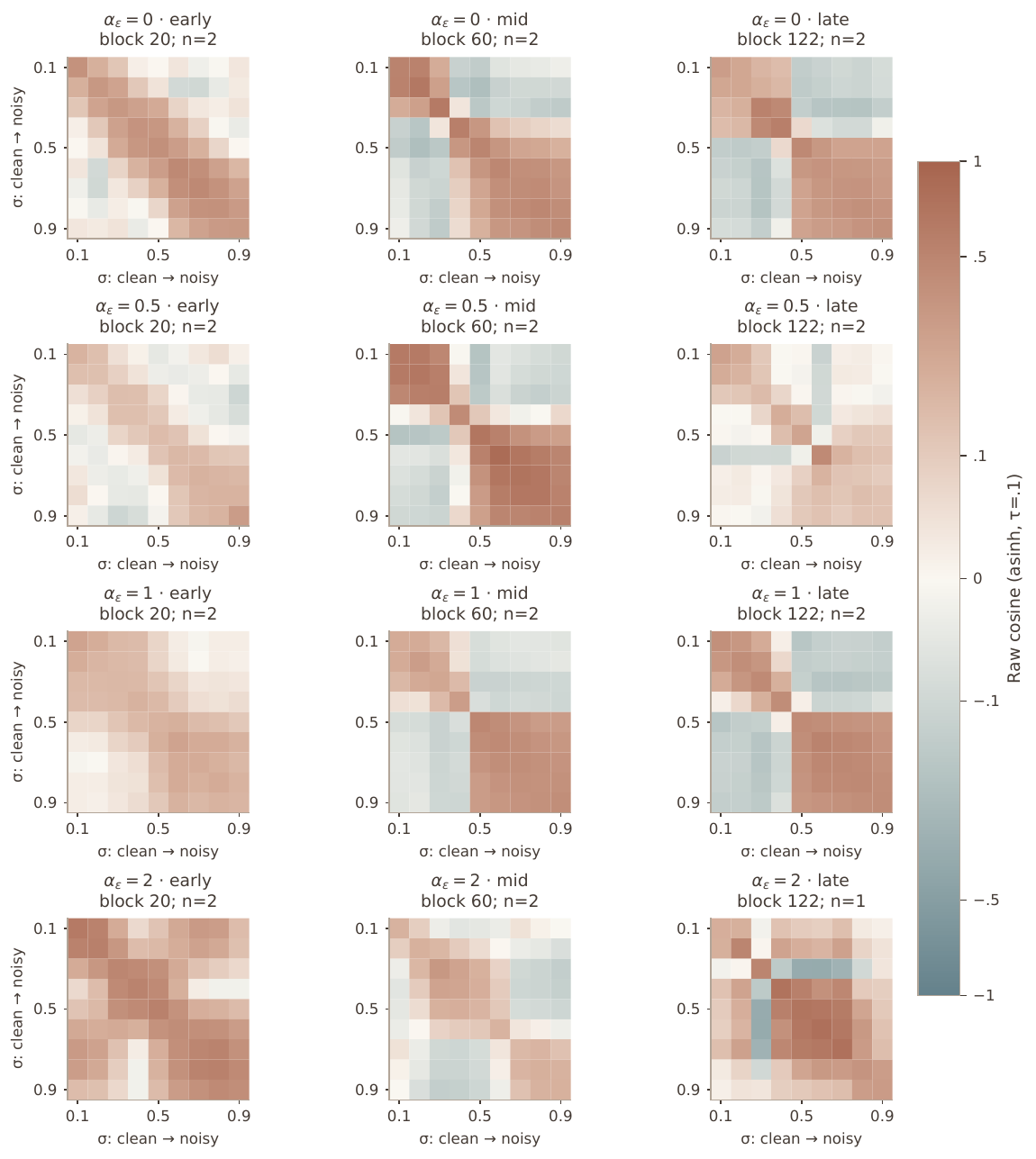}
\caption{Ball in Cup fixed-profile robot policies: phase-averaged cross-batch gradient cosines across clean-to-noisy bins (columns: early, middle, late; rows: tested exponents). The early $\alpha_\epsilon=2$ panels show more agreement than its middle phase, illustrating stage-dependent rather than fixed coordination. Colors show the common raw-cosine scale, and $n$ counts contributing seeds; Appendix~\ref{app:robotics-new-geometry} defines phase aggregation. At exponent 2, seed 901 failed at block 114; the late panel contains seed 902 only.}
\label{fig:robotics-cross-ballincup-0}
\end{figure}
\clearpage
\begin{figure}[p]
\centering
\includegraphics[width=.98\linewidth]{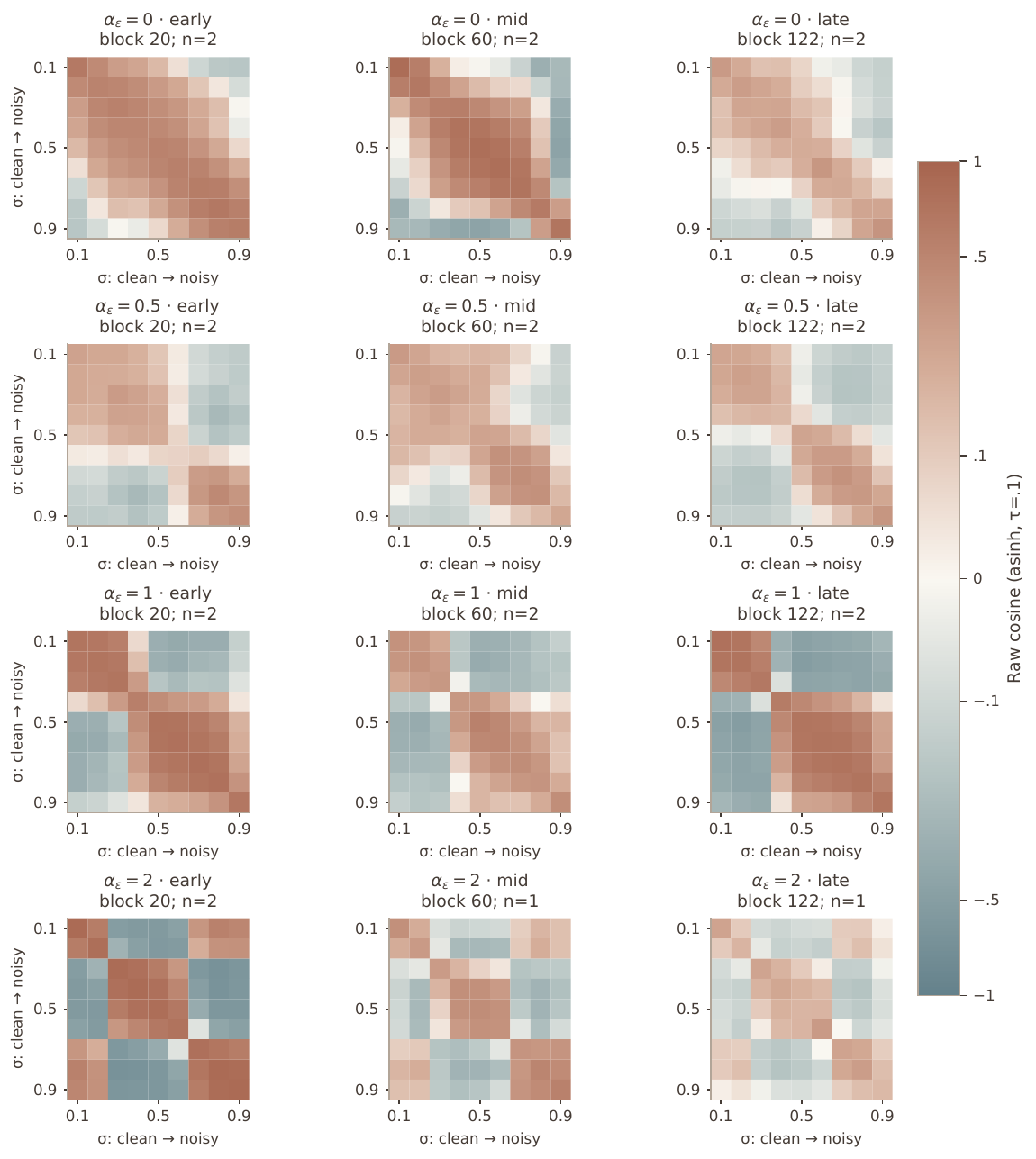}
\caption{Cheetah fixed-profile robot policies: phase-averaged cross-batch gradient cosines across clean-to-noisy bins (columns: early, middle, late; rows: tested exponents). Early $\alpha_\epsilon=0$ agreement weakens late, while the $2$ panels remain near zero on average. Colors show the common raw-cosine scale, and $n$ counts contributing seeds; Appendix~\ref{app:robotics-new-geometry} defines phase aggregation. At exponent 2, seed 902 failed at block 58; middle and late panels contain seed 901 only.}
\label{fig:robotics-cross-cheetahrun-0}
\end{figure}
\clearpage
\begin{figure}[p]
\centering
\includegraphics[width=.98\linewidth]{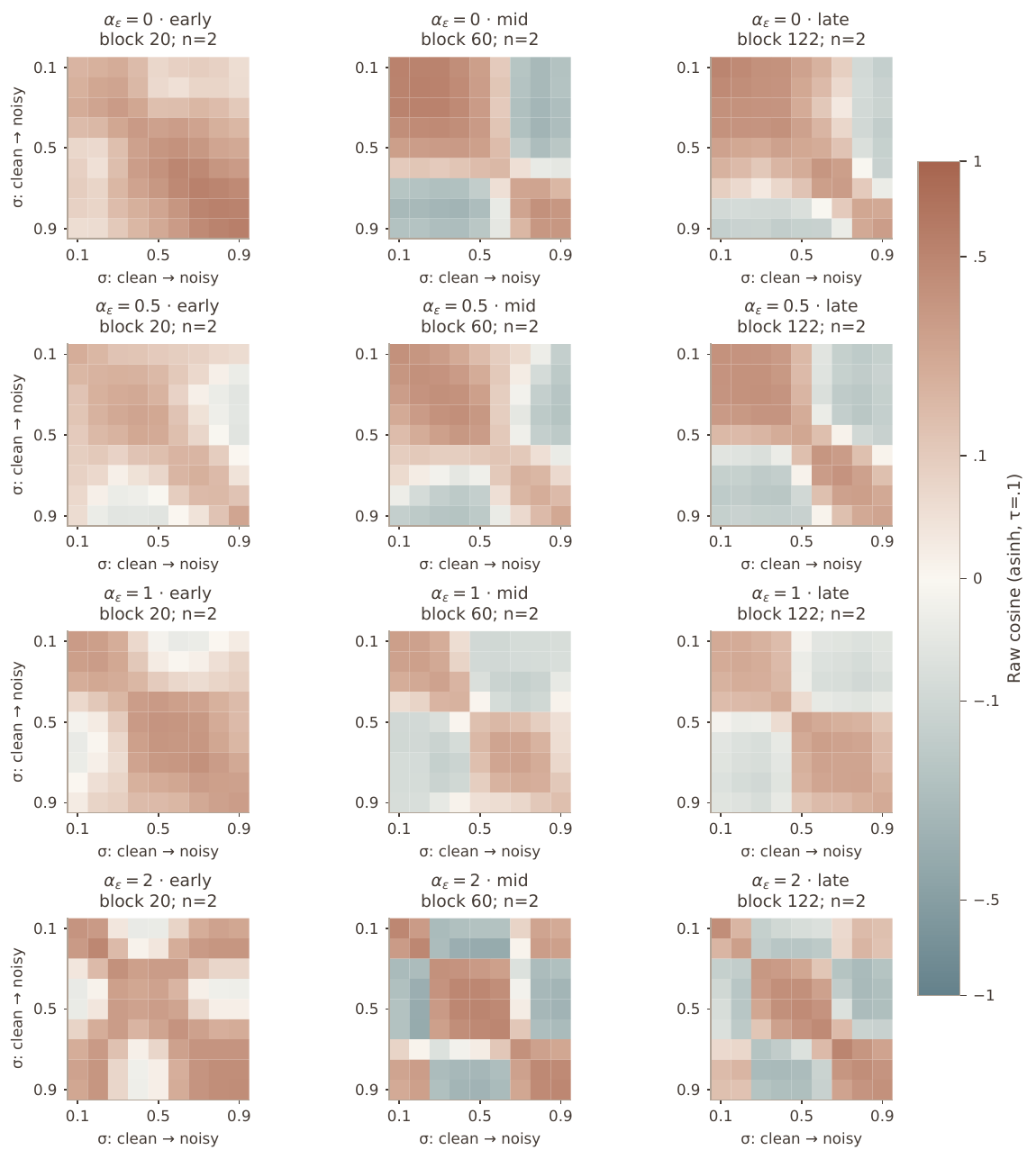}
\caption{Fish fixed-profile robot policies: phase-averaged cross-batch gradient cosines across clean-to-noisy bins (columns: early, middle, late; rows: tested exponents). Cross-noise agreement declines from early to late at every displayed exponent. Colors show the common raw-cosine scale, and $n$ counts contributing seeds; Appendix~\ref{app:robotics-new-geometry} defines phase aggregation.}
\label{fig:robotics-cross-fishswim-0}
\end{figure}
\clearpage
\begin{figure}[p]
\centering
\includegraphics[width=.98\linewidth]{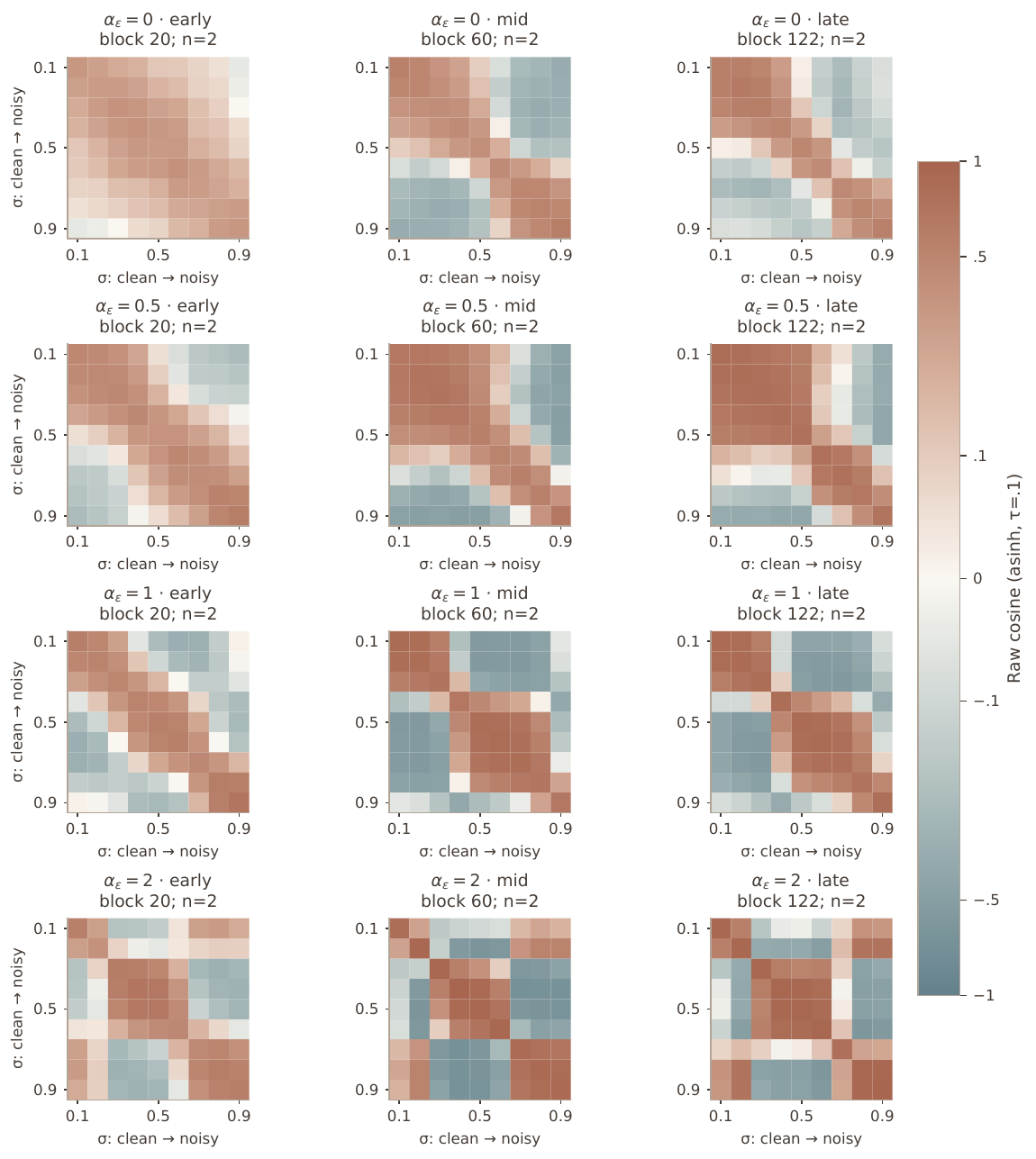}
\caption{Swimmer fixed-profile robot policies: phase-averaged cross-batch gradient cosines across clean-to-noisy bins (columns: early, middle, late; rows: tested exponents). The $\alpha_\epsilon=.5$ late panel retains more agreement than its early panel, unlike several other profiles. Colors show the common raw-cosine scale, and $n$ counts contributing seeds; Appendix~\ref{app:robotics-new-geometry} defines phase aggregation.}
\label{fig:robotics-cross-swimmerswimmer6-0}
\end{figure}
\clearpage
\begin{figure}[p]
\centering
\includegraphics[width=.98\linewidth]{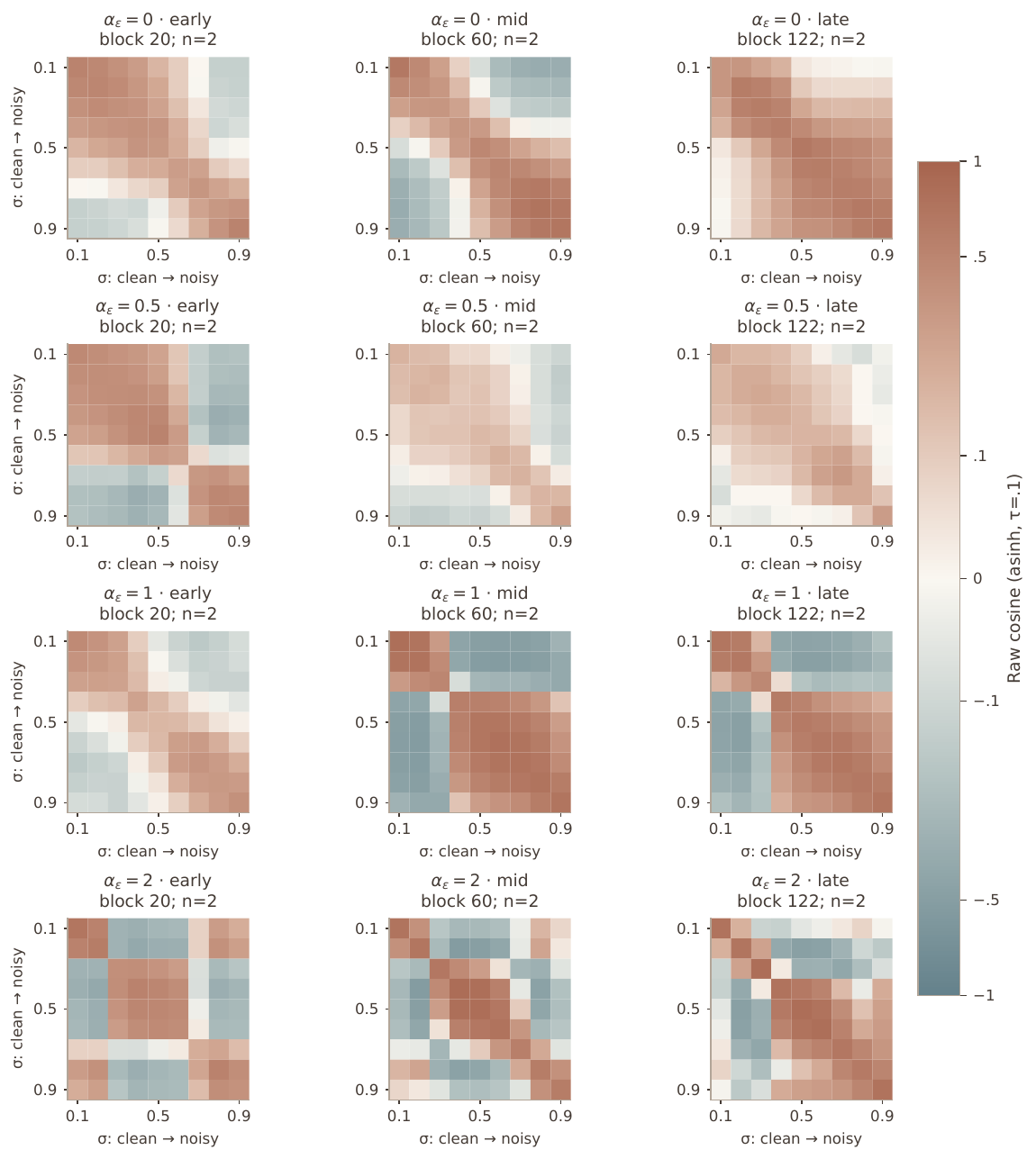}
\caption{Walker fixed-profile robot policies: phase-averaged cross-batch gradient cosines across clean-to-noisy bins (columns: early, middle, late; rows: tested exponents). The $\alpha_\epsilon=0$ late panel gains agreement, whereas the native $1$ panels also contain opposing noise blocks. Colors show the common raw-cosine scale, and $n$ counts contributing seeds; Appendix~\ref{app:robotics-new-geometry} defines phase aggregation.}
\label{fig:robotics-cross-walkerrun-0}
\end{figure}
\clearpage
\begin{figure}[p]
\centering
\includegraphics[width=.98\linewidth]{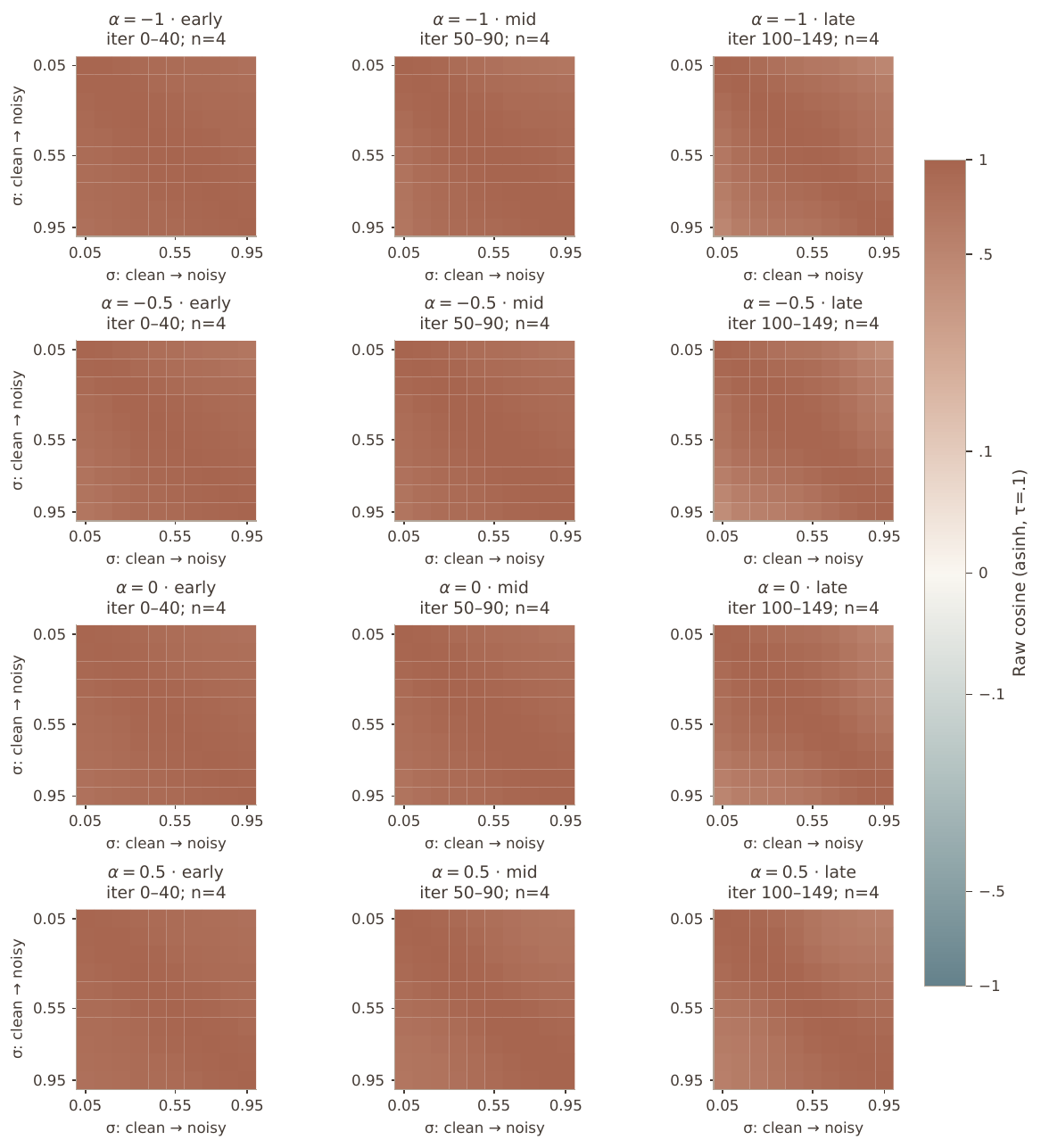}
\caption{Cartpole fixed-profile robot policies: phase-averaged cross-batch gradient cosines across clean-to-noisy bins (columns: early, middle, late; rows: tested exponents). Cross-noise agreement remains broadly high for every recorded profile, even as its magnitude changes by phase. Colors show the common raw-cosine scale, and $n$ counts contributing seeds; Appendix~\ref{app:robotics-new-geometry} defines phase aggregation.}
\label{fig:robotics-cross-cartpole-0}
\end{figure}
\clearpage
\begin{figure}[p]
\centering
\includegraphics[width=.98\linewidth]{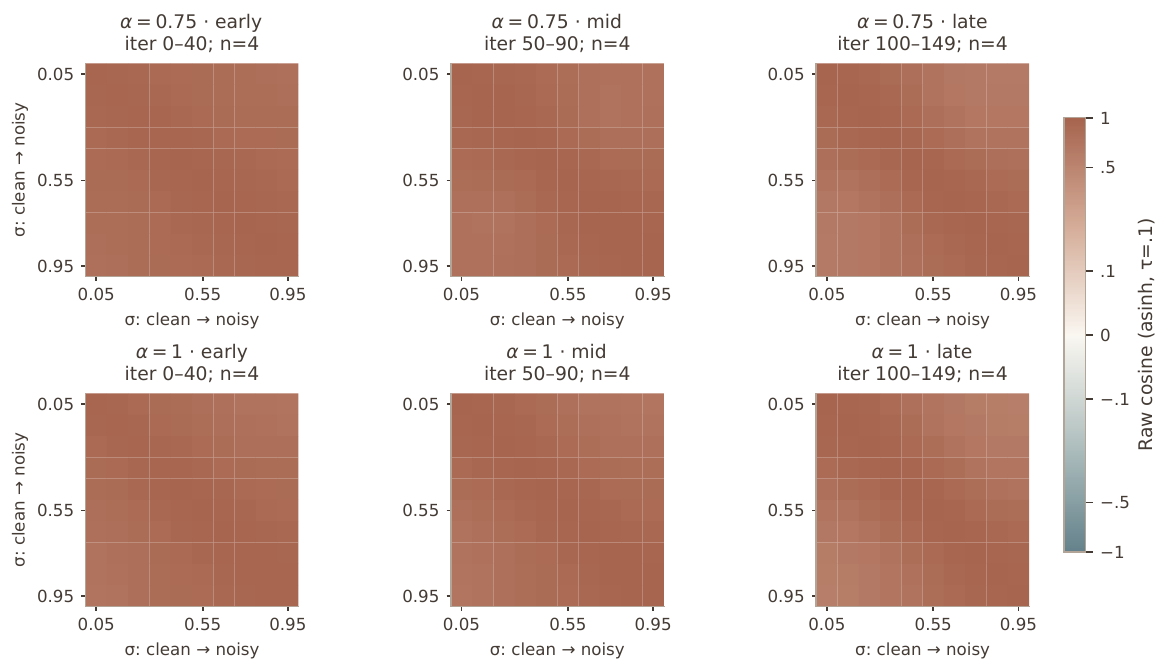}
\caption{Cartpole fixed-profile robot policies: phase-averaged cross-batch gradient cosines across clean-to-noisy bins (columns: early, middle, late; rows: tested exponents). Cross-noise agreement remains broadly high for every recorded profile, even as its magnitude changes by phase. Colors show the common raw-cosine scale, and $n$ counts contributing seeds; Appendix~\ref{app:robotics-new-geometry} defines phase aggregation.}
\label{fig:robotics-cross-cartpole-1}
\end{figure}
\clearpage
\begin{figure}[p]
\centering
\includegraphics[width=.98\linewidth]{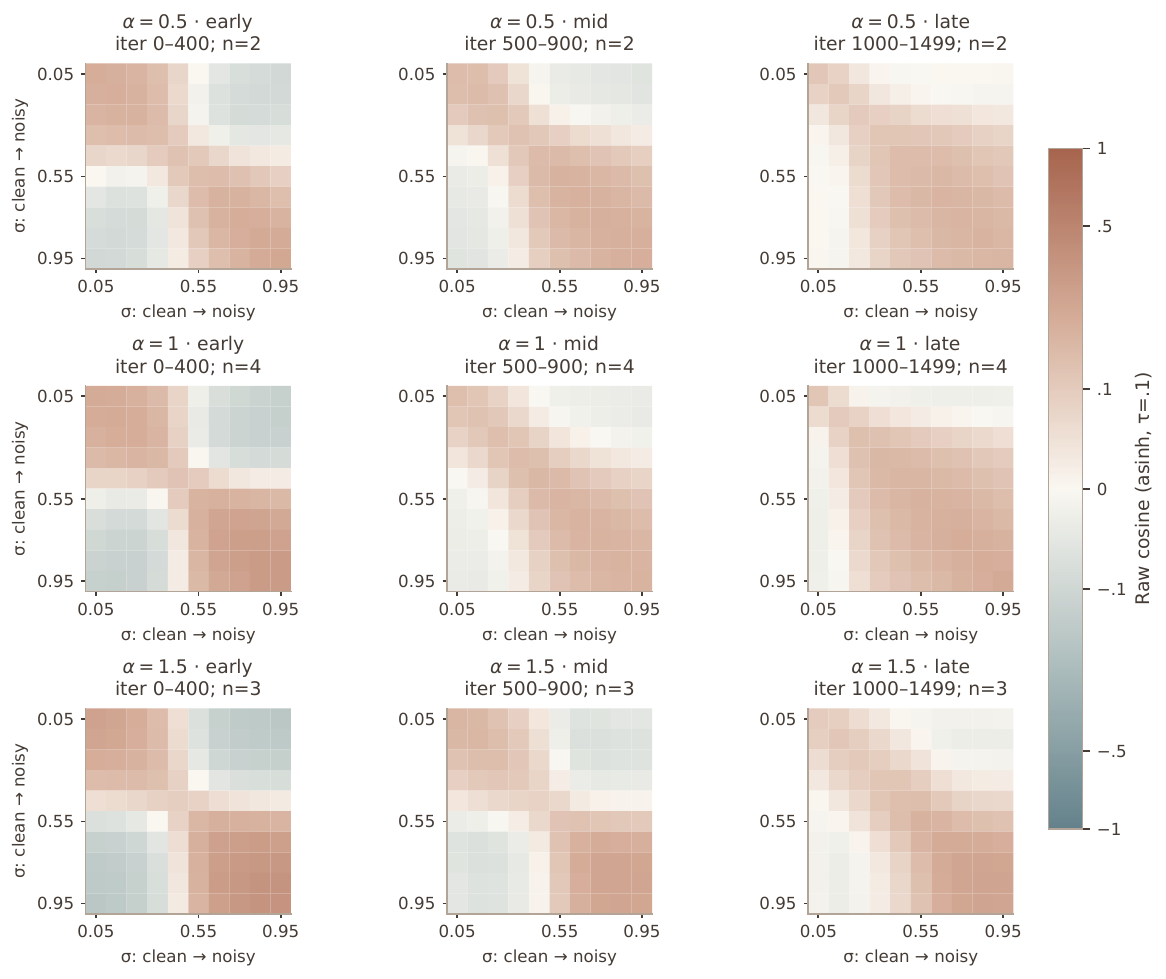}
\caption{G1 fixed-profile robot policies: phase-averaged cross-batch gradient cosines across clean-to-noisy bins (columns: early, middle, late; rows: tested exponents). Off-diagonal agreement stays weak across all recorded profiles and phases, contrasting with Cartpole. Colors show the common raw-cosine scale, and $n$ counts contributing seeds; Appendix~\ref{app:robotics-new-geometry} defines phase aggregation.}
\label{fig:robotics-cross-g1-0}
\end{figure}
\clearpage
\begin{figure}[p]
\centering
\includegraphics[width=.98\linewidth]{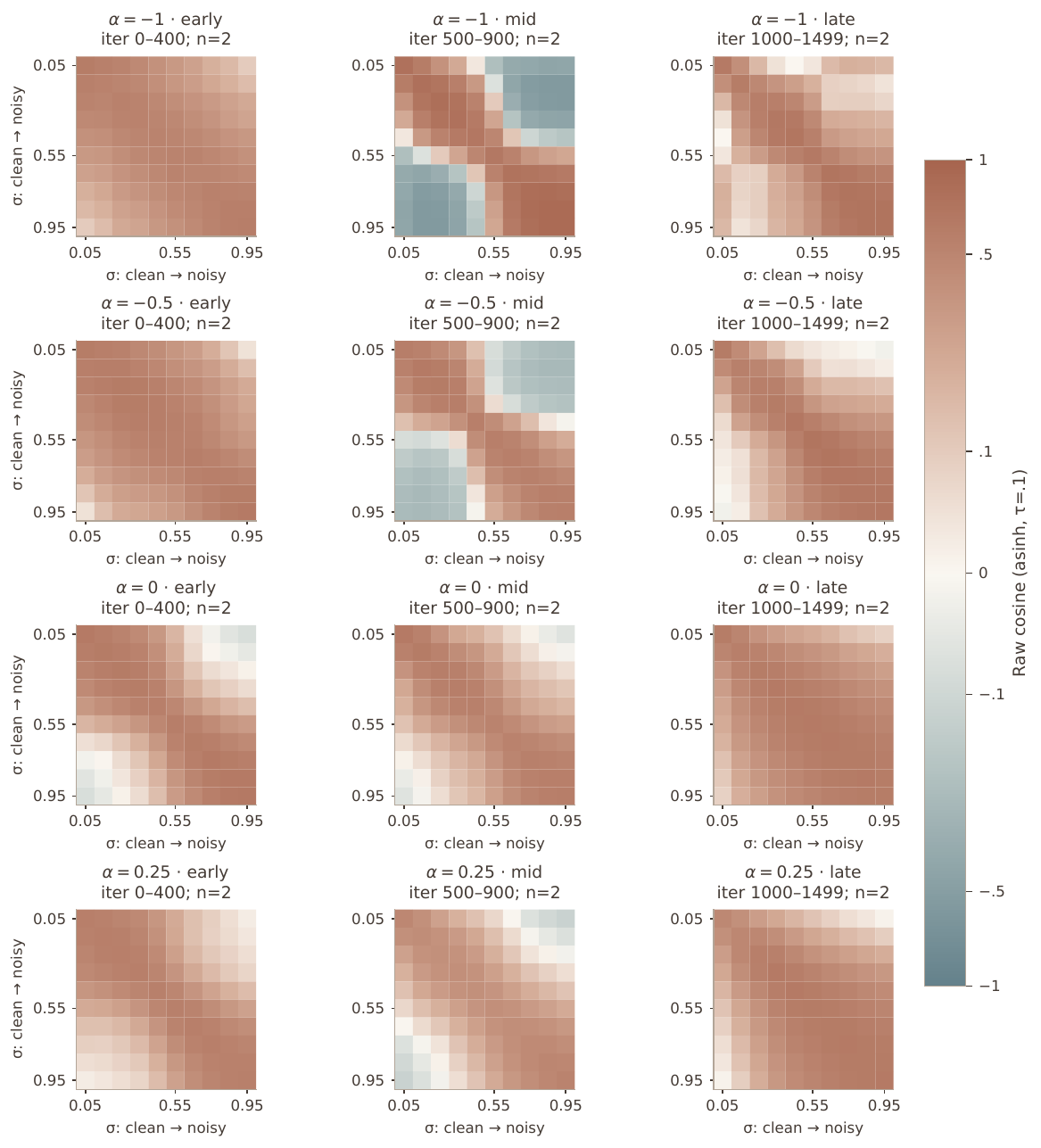}
\caption{Go2 fixed-profile robot policies: phase-averaged cross-batch gradient cosines across clean-to-noisy bins (columns: early, middle, late; rows: tested exponents). The $\alpha=0$ late phase retains more overall agreement than the negative-exponent profiles despite their stronger early agreement. Colors show the common raw-cosine scale, and $n$ counts contributing seeds; Appendix~\ref{app:robotics-new-geometry} defines phase aggregation.}
\label{fig:robotics-cross-go2-0}
\end{figure}
\clearpage
\begin{figure}[p]
\centering
\includegraphics[width=.98\linewidth]{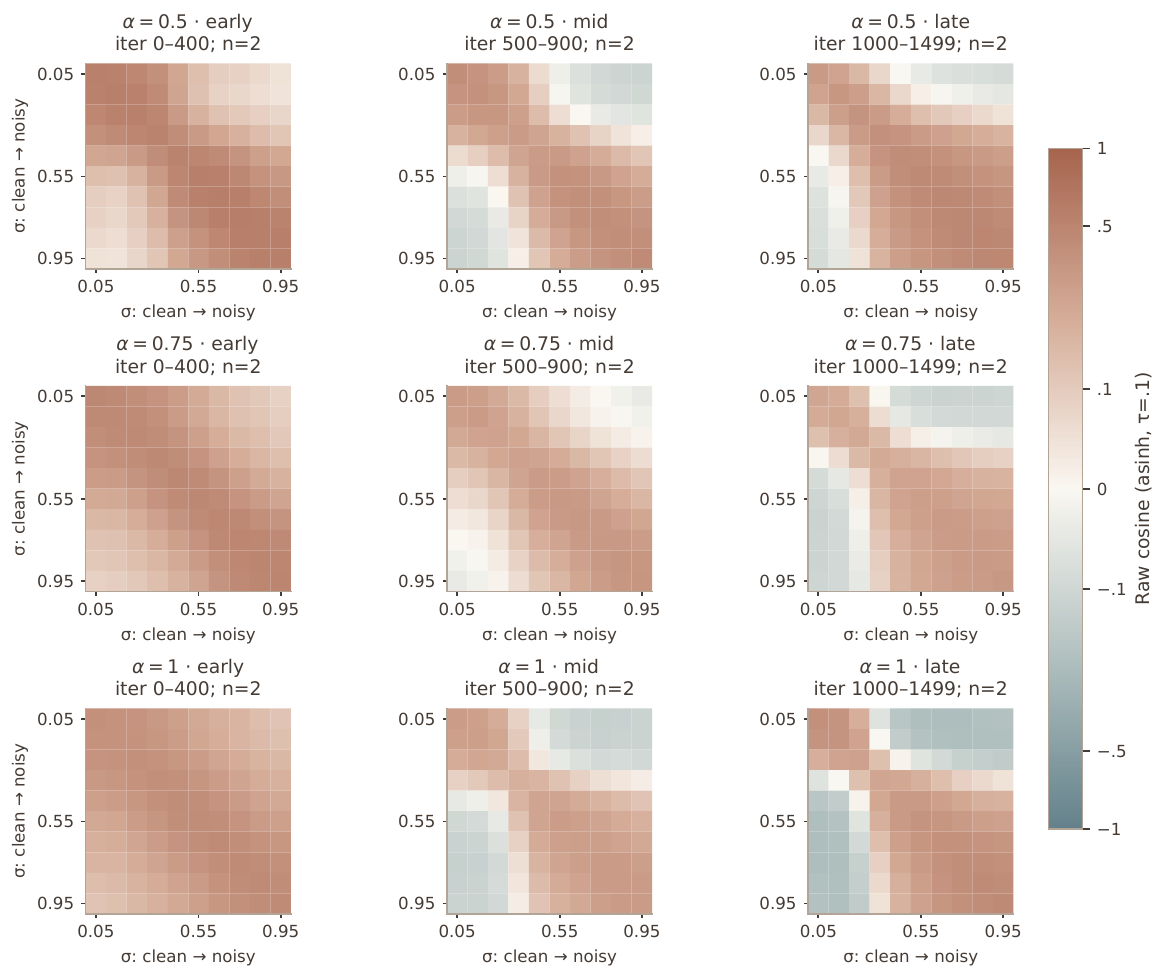}
\caption{Go2 fixed-profile robot policies: phase-averaged cross-batch gradient cosines across clean-to-noisy bins (columns: early, middle, late; rows: tested exponents). The native $\alpha=1$ profile loses much of its early agreement by the late phase; its higher peak return than $\alpha=0$ does not imply stronger late coordination. Colors show the common raw-cosine scale, and $n$ counts contributing seeds; Appendix~\ref{app:robotics-new-geometry} defines phase aggregation.}
\label{fig:robotics-cross-go2-1}
\end{figure}
\clearpage
\begin{figure}[p]
\centering
\includegraphics[width=.98\linewidth]{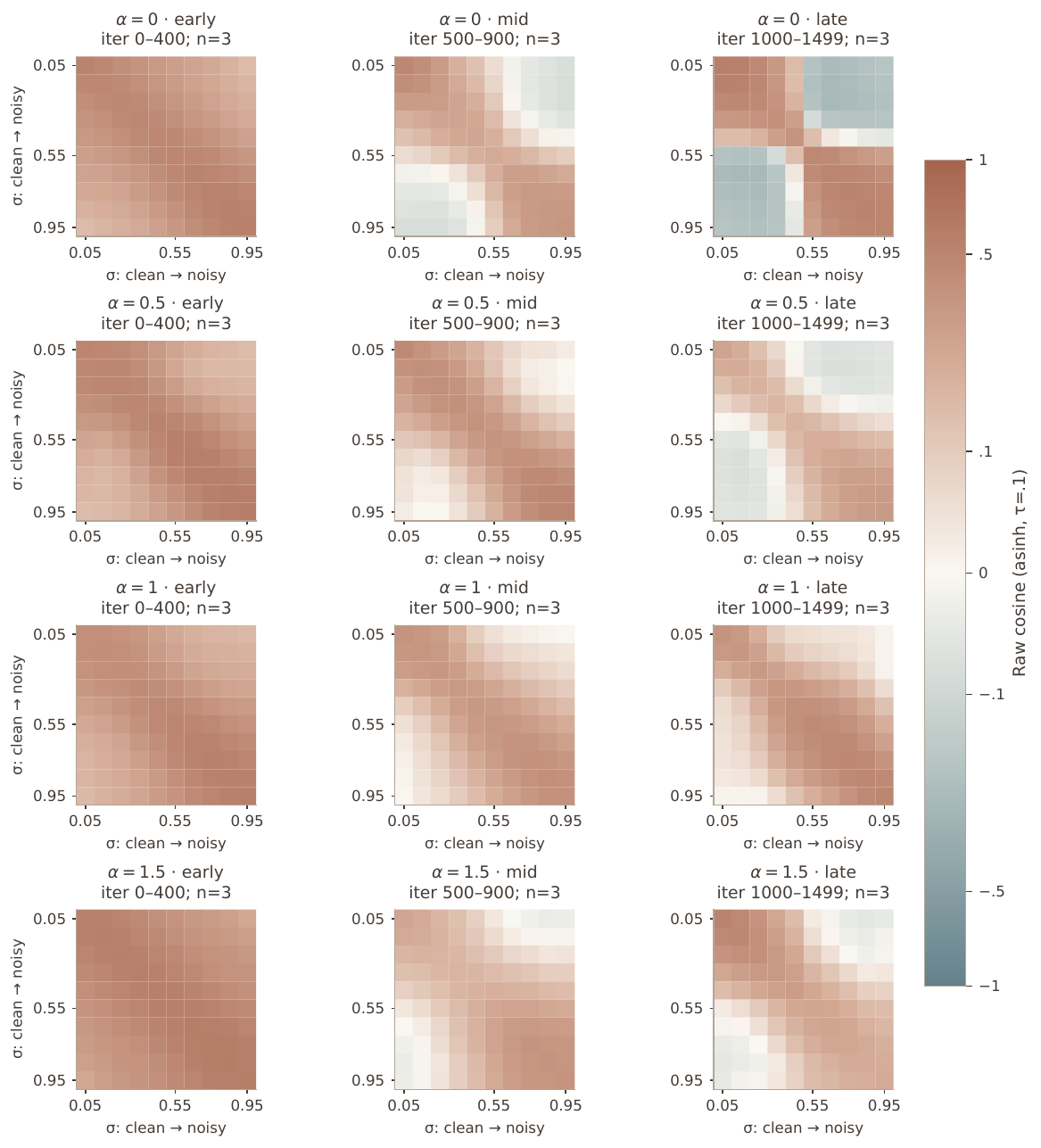}
\caption{Spot fixed-profile robot policies: phase-averaged cross-batch gradient cosines across clean-to-noisy bins (columns: early, middle, late; rows: tested exponents). The native $\alpha=1$ late phase retains more agreement than $\alpha=0$, illustrating a different within-task trajectory from Go2. Colors show the common raw-cosine scale, and $n$ counts contributing seeds; Appendix~\ref{app:robotics-new-geometry} defines phase aggregation.}
\label{fig:robotics-cross-spot-0}
\end{figure}

\clearpage

\end{document}